\RequirePackage{etex}
\documentclass{article}
\usepackage{microtype}
\usepackage{graphicx}
\usepackage{booktabs} 
\usepackage[numbers]{natbib}
\usepackage{tabularx} 
\usepackage[utf8]{inputenc} 
\usepackage[T1]{fontenc}    
\usepackage{url}            
\usepackage{booktabs}       
\usepackage{amsfonts}       
\usepackage{nicefrac}       
\usepackage{microtype}      
\usepackage{makecell}

\usepackage[export]{adjustbox}
\usepackage{etoolbox}
\usepackage{autonum}

\usepackage{amsthm}
\usepackage{graphicx}
\usepackage{xcolor}
\usepackage{comment}
\usepackage{subcaption}
\usepackage{multirow}
\usepackage{overpic}
\usepackage{bbm}
\usepackage{lipsum,tikz,wrapfig}
\usepackage[normalem]{ulem} 

\usepackage[ruled,vlined]{algorithm2e} 
\usepackage{soul}

\usepackage{amsmath,amsfonts,bm}

\newcommand{\set}[1]{{\{ #1\}}}

\def\eqref#1{equation~\ref{#1}}

\def\1{\bm{1}}

\def\eps{{\epsilon}}

\def\rx{{\textnormal{x}}}
\def\ry{{\textnormal{y}}}

\def\rvx{{\mathbf{x}}}

\def\ervx{{\textnormal{x}}}
\def\ervy{{\textnormal{y}}}

\def\vone{{\bm{1}}}

\def\vw{{\bm{w}}}
\def\vx{{\bm{x}}}
\def\vy{{\bm{y}}}

\def\mI{{\bm{I}}}

\def\mX{{\bm{X}}}

\DeclareMathAlphabet{\mathsfit}{\encodingdefault}{\sfdefault}{m}{sl}
\SetMathAlphabet{\mathsfit}{bold}{\encodingdefault}{\sfdefault}{bx}{n}

\def\gD{{\mathcal{D}}}

\def\gN{{\mathcal{N}}}

\def\gX{{\mathcal{X}}}

\def\sR{{\mathbb{R}}}

\newcommand{\norm}[1]{\left\lVert#1\right\rVert}

\newcommand{\ie}{\textit{i.e.}}
\newcommand{\eg}{\textit{e.g.}}
\usepackage{mathtools}
\usepackage{amsmath}
\usepackage{array}
\newcolumntype{L}{>{\centering\arraybackslash}m{3cm}}

\usepackage{csquotes}

\definecolor{shadecolor}{rgb}{0.95,0.95,1}
\usepackage{framed}
\usepackage{color}

\usepackage{makecell}
\usepackage{float}
\usepackage{array}
\newcolumntype{P}[1]{>{\centering\arraybackslash}p{#1}}

\usepackage[colorinlistoftodos]{todonotes}
\usepackage{tocloft}
\usepackage{titletoc}
\usepackage{diagbox}
\usepackage[bookmarks=false]{hyperref}
\usepackage[capitalize,noabbrev]{cleveref}

\newcommand\bmt[1]{{#1}}

\newcommand\tuan[1]{{#1}}


\usepackage{siunitx}
\newtheorem*{remark}{Remark}
\makeatother

\newcommand{\lift}{\textsc{LIFT}}
\newcommand{\gpt}{GPT}
\newcommand{\gptt}{GPT-3}
\newcommand{\gptj}{GPT-J}

\newcommand{\qr}{PR}
\newcommand{\knn}{KNN}
\newcommand{\krr}{KR}
\newcommand{\ann}{MLP}
\newcommand{\xg}{XG}
\newcommand{\gbt}{GBT}
\newcommand{\randomf}{RF}
\newcommand{\gp}{GP}

\newcommand{\openml}{\citep{OpenML2013} }
\newcommand{\mcc}{majority class classifier}

\newcommand{\linear}{Linear}
\newcommand{\quadr}{Quadratic}
\newcommand{\cosi}{Cosine}
\newcommand{\expo}{Exponential}
\newcommand{\pw}{Piecewise}
\newcommand{\lone}{L1norm}
\newcommand{\prompt}[1]{\textcolor{purple}{\textbf{#1}}}

\makeatletter 
\let\c@table\c@figure
\let\c@lstlisting\c@figure
\makeatother

\usepackage[final]{neurips_2022}

\usepackage[utf8]{inputenc} 
\usepackage[T1]{fontenc}    
\usepackage{url}            
\usepackage{booktabs}       
\usepackage{amsfonts}       
\usepackage{nicefrac}       
\usepackage{microtype}      
\usepackage{xcolor}         
\usepackage{bigstrut}

\makeatletter
\newcommand\footnoteref[1]{\protected@xdef\@thefnmark{\ref{#1}}\@footnotemark}
\makeatother

\newcommand\blfootnote[1]{
  \begin{NoHyper}
  \renewcommand\thefootnote{}\footnote{#1}
  \addtocounter{footnote}{-1}
  \end{NoHyper}
}

\title{\lift{}: Language-Interfaced Fine-Tuning \\ for Non-Language Machine Learning Tasks}

\author{
Tuan Dinh\footnotemark[1],\ \ Yuchen Zeng\footnotemark[1],\ \  Ruisu Zhang
,\ \ \normalsize \textbf{Ziqian Lin}
,\ \ \normalsize \textbf{Michael Gira},\\ 
\textbf{\ \ Shashank Rajput
,\ \ Jy-yong Sohn
,\ \ Dimitris Papailiopoulos
,\ \ Kangwook Lee
} \\ \\
\normalsize 
University of Wisconsin-Madison, USA \\ 
}

\begin{document}

\maketitle
\vspace{-.2in}
\begin{abstract}
Fine-tuning pretrained language models (LMs) without making any architectural changes has become a norm for learning various language downstream tasks.
However, for \emph{non}-language downstream tasks, a common practice is to employ task-specific designs for input, output layers, and loss functions.
For instance, it is possible to fine-tune an LM into an MNIST classifier by replacing the word embedding layer with an image patch embedding layer, the word token output layer with a 10-way output layer, and the word prediction loss with a 10-way classification loss, respectively.
A natural question arises: 
Can LM fine-tuning solve non-language downstream tasks \emph{without} changing the model architecture or loss function?
To answer this, we propose \textbf{Language-Interfaced Fine-Tuning (\lift{})} and study its efficacy and limitations by conducting an extensive empirical study on a suite of non-language classification and regression tasks.
\lift{} does not make {\it any} changes to the model architecture or loss function, and it solely relies on the natural language interface, enabling ``no-code machine learning with LMs.''
We find that \lift{} performs comparably well across a wide range of low-dimensional classification and regression tasks, matching the performances of the best baselines in many cases, especially for the classification tasks. 
We also report experimental results on the fundamental properties of \lift{}, including inductive bias, robustness, and sample complexity.
We also analyze the effect of pretraining on \lift{} and a few properties/techniques specific to \lift{}, \textit{e.g.,} context-aware learning via appropriate prompting, calibrated predictions, data generation, and two-stage fine-tuning.
Our code is available at \url{https://github.com/UW-Madison-Lee-Lab/LanguageInterfacedFineTuning}.
\blfootnote{$^*$Equal contribution. Emails: Tuan Dinh (\texttt{tuan.dinh@wisc.edu}), Yuchen Zeng (\texttt{yzeng58@wisc.edu})}
\end{abstract}

\vspace{-.2in}
\section{Introduction}\label{sec:introduction}
\vspace{-.12in}
Deep neural networks have been highly successful across a multitude of domains, from computer vision~\citep{forsyth2011computer,dosovitskiy2020image} and natural language processing~\citep{chowdhary2020natural,bommasani2021opportunities}, to game playing~\citep{wang2016does,arulkumaran2019alphastar}.
Most advances in deep learning have come with a variety of domain-specific designs for network architectures, such as convolutional filters~\citep{lo1995artificial,he2016deep,he2017mask} for vision tasks, or recurrent modules~\citep{rumelhart1985learning,hochreiter1997long} and attention mechanisms~\citep{vaswani2017attention,so2019evolved} in the context of natural language processing.
A domain-and-modality agnostic model that can be adapted to solve tasks across different modalities and domains has become a desideratum~\citep{kaiser2017one}, motivating great efforts in transfer learning~\citep{weiss2016survey} and multi-modal learning~\citep{ramachandram2017deep}.
Recently, transformer-based language models (LMs)~\citep{so2019evolved,collobert2008gpt2,lu2021pretrained,brown2020gpt3} exhibited impressive versatility across different domains and modalities.
They have shown great performances for various language-based tasks~\citep{li2021quantifying} such as question answering~\citep{radford2019language,su-etal-2019-generalizing}, or commonsense reasoning~\citep{zhou2019common}.
They have also been applied to non-language modalities~\citep{lu2021pretrained}.
For instance, GPT-2~\citep{collobert2008gpt2} pretrained on language data can be efficiently fine-tuned to perform image classification and numerical computation~\citep{lu2021pretrained}.

\begin{figure}
    \vspace{-.3in}
    \centering
    \includegraphics[width=1\textwidth]{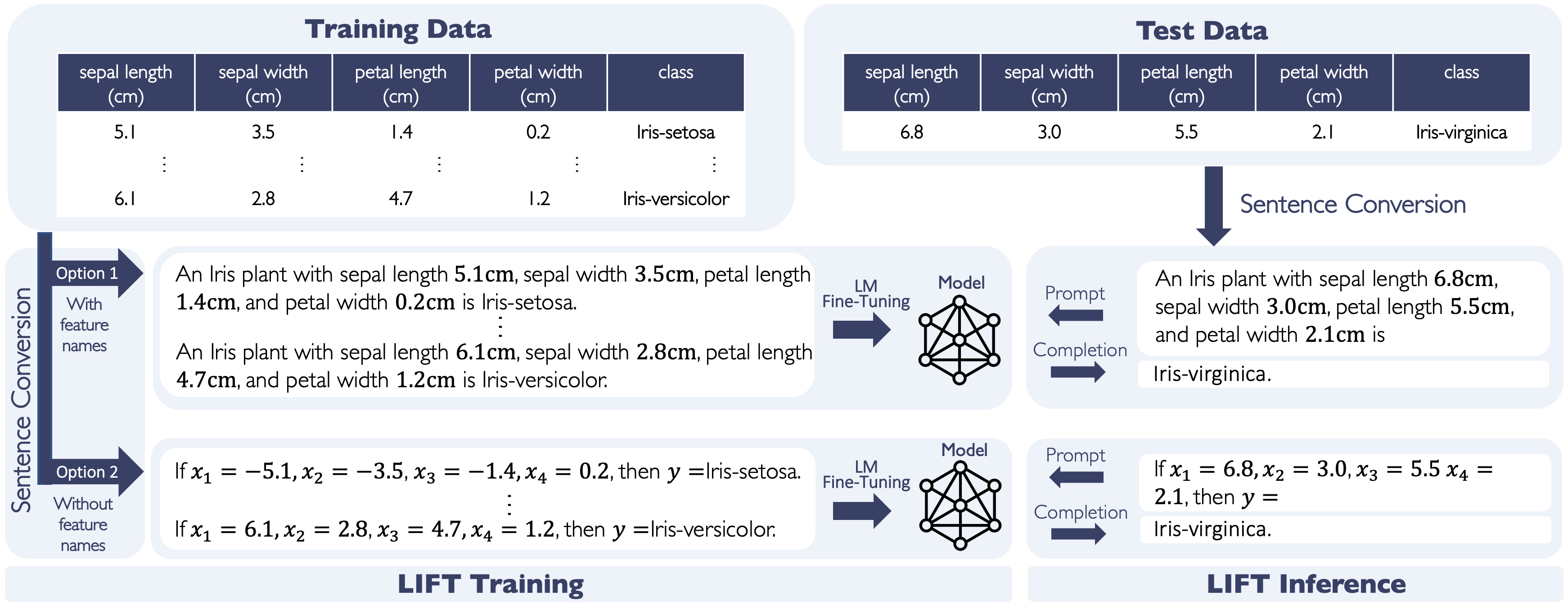}
    \vspace{-4mm}
    \caption{\textbf{A high-level illustration of the Language-Interfaced Fine-Tuning (\lift{}) framework.}
    \lift{} has a two-phase procedure: (1) converting the dataset into sentences and (2) fine-tuning the pretrained language model (\textit{e.g.}, \gpt{}) on the obtained sentences. 
    This figure visualizes how LIFT can be applied to the Iris classification task.
    We first convert the Iris dataset into plain English sentences (left).
    Since feature names and the task description are available for this task, one could incorporate them as part of the prompt (as option 1 in the figure).
    (In Sec.~\ref{sec:feature_names}, we show that adding such contextual information to prompts helps LIFT achieve higher predictive accuracy.)
    One may also choose to use a simpler prompt with a generic naming convention ($x_1, x_2, \ldots, x_d$) for $p$ features (as option 2 in the figure).
    After the sentence conversion step, \lift{} fine-tunes a pretrained LM with the sentence set without making any changes to model architecture or loss.
    At inference time, we convert the test samples to a sentence form using the same prompt, excluding the label part. 
    \lift{} performs surprisingly well in various non-language  regression/classification tasks, and we summarize our main findings in Table~\ref{tab:summary_basic_findings}.
    Note that to obtain a model for a given task, all we need here is to design proper sentence templates for \lift{} and no changes to architecture or loss functions are needed.
     }
    \label{fig:prompts}
    \vspace{-.1in}
\end{figure}

When downstream tasks are language-based tasks, adapting pretrained LMs can be achieved without modifying the models' architecture.
Typically, this adaptation is enabled via simple fine-tuning~\citep{hu2021lora,houlsby2019parameter,rebuffi2017learning,wang2020k} or in-context few-shot learning methods~\citep{lu2021fantastically,min2021recent}.
However, not altering the architecture may pose a limitation for transferring to non-language tasks.
As their input and output formats are not in some language form, adapting LMs to these domains may seem to require architectural changes.
Indeed, it has been a common practice to re-design the input/output layers and loss functions to accommodate a different predictive task.
For instance, to adapt GPT-2~\citep{radford2019language} to other modalities, the frozen pretrained transformer~\citep{lu2021pretrained} adds new input/output layers to handle different types of input/output.
To make such changes, one must have a good understanding of the underlying principles of LMs and an ability to make proper modifications at the code level. 

A natural question that arises is whether such changes are necessary. In other words,\\
\noindent\fbox{\begin{minipage}{\dimexpr\textwidth-2\fboxsep-2\fboxrule\relax}
\centering
Does language model fine-tuning work for non-language tasks \\ \textbf{without} changing the architecture or loss function at all?
\end{minipage}}

To answer this, we consider a simple fine-tuning procedure for LMs, referred to as \textbf{Language-Interfaced Fine-Tuning (\lift{})}.
This procedure can be used to learn predictors for any classification or regression task.
\lift{} runs in two phases: (1) converting labeled samples into sentences, and (2) fine-tuning pretrained LMs on the sentence dataset without altering the architecture or loss function.

Fig.~\ref{fig:prompts} illustrates how we fine-tune \gpt{} with \lift{} to solve the Iris classification task~\cite{iris}. 
\lift{} first converts each labeled sample into a sentence with two options.
The first option is to incorporate feature names and the task description into the sentence template. 
In this example, we could convert a training sample \prompt{r} into ``An Iris plant with sepal length \prompt{r.sepal\_length}, sepal width \prompt{r.sepal\_width}, petal length \prompt{r.petal\_length}, and petal width \prompt{r.petal\_width} is \prompt{r.class}.''
Here, we use the dot notation, \textit{i.e.,} \prompt{r.$\star$} denotes the string conversion of the corresponding attribute of sample \prompt{r}.
One may also adopt a simpler (and more generic) sentence template, such as 
``If x1=\prompt{r.x1}, x2=\prompt{r.x2}, \ldots, xp=\prompt{r.xp}, then y=\prompt{r.y},'' if there are $p$ features.  
We then fine-tune LMs without changing either  architecture or loss function. 
Then, we perform inference as follows.
\lift{} first converts test samples into sentences using the same template while leaving the prediction part empty. 
It then feeds the converted sentences as prompts to the fine-tuned model.
The output tokens are parsed to provide the final predictions.

\begin{figure}[t]
    \centering
    \vspace{-.5in}
    \includegraphics[width=0.85\textwidth]{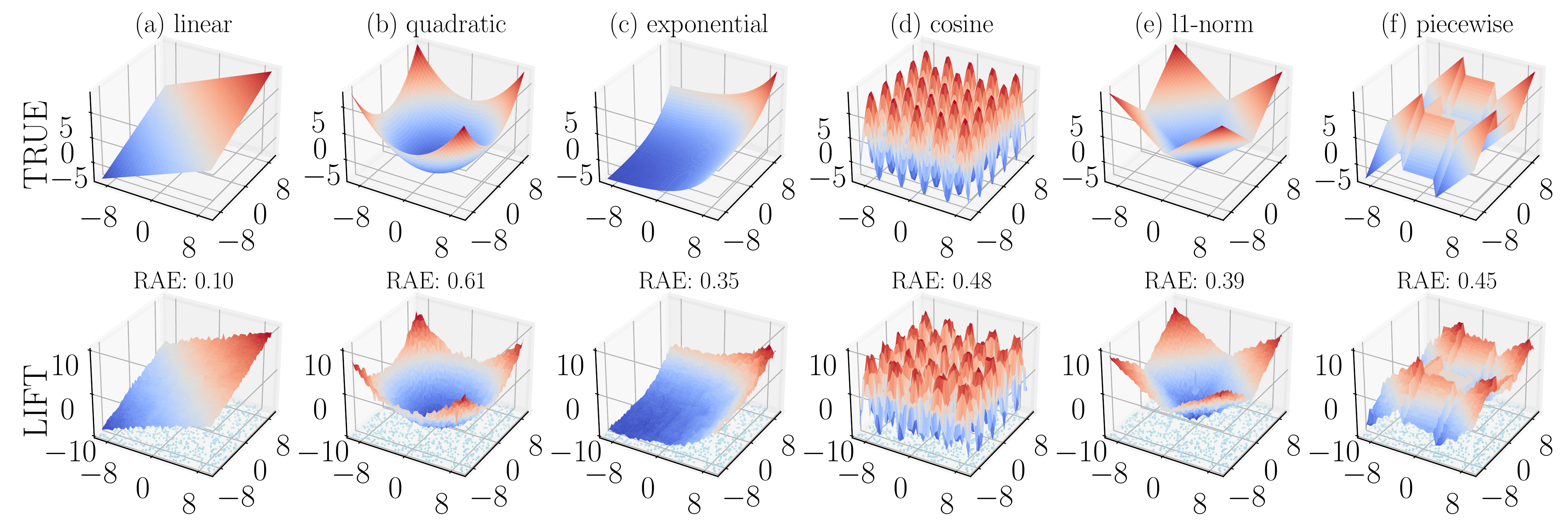}
    \caption{\textbf{Approximating various functions with \lift{}  using \gptj{}. }
    We visualize the target functions (first row) and the predictor functions learned by \lift{} on \gptj{} (second row). 
    Blue dots show the $1000$ training samples.
    One can observe that \lift{} well approximates the target functions.
    }
    \label{fig:reg_2d}
    \vspace{-.15in}
\end{figure}

Our work empirically shows that \lift{} can provide high-accuracy solutions for a variety of non-language tasks. 
Fig.~\ref{fig:reg_2d} shows examples of real functions learned by \gptj{} models~\citep{gpt-j} fine-tuned using \lift{} given 1000 samples.
Recall that \lift{} does not require any changes in the architecture or loss function,
Thus, our findings show that such changes to architecture/loss function might \emph{not} be necessary, even when the target predictive task is not a language task.
Thus, \lift{} can be almost perceived as a ``no-code machine learning'' framework as the data-to-sentence conversion is extremely straightforward even without extensive programming skills and machine learning backgrounds.

Motivated by these intriguing properties, we investigate the efficacy and limitations of \lift{} on non-language tasks by conducting an extensive empirical study on a suite of classification and regression tasks.
\textit{First}, we observe that \lift{} performs well across a wide range of low-dimensional classification and regression tasks.
In most cases, it nearly matches (or slightly outperforms) the best baselines' performance.
To further understand \lift{}, we conduct experiments testing the fundamental learning properties, \textit{e.g.,} its 
inductive bias, 
sample efficiency, 
ability to extrapolate, 
worst- and average-case noise robustness, 
and how the pretraining of LMs affects \lift{}.
\textit{Third}, we study a few unique properties specific to \lift{}, \textit{e.g.,} 
context-aware learning with task-specific prompting, 
prediction calibration, 
and the additional use of \lift{} for data generation.
\textit{Lastly}, to improve upon the basic fine-tuning, we employ a few techniques: 
two-stage fine-tuning with synthetic pretext tasks
and 
data augmentation. Both techniques improve the performance of \lift{}.
We finally provide discussions on limitations and future investigations of \lift{}.

\textbf{Scope of the study.~} 
Our work proposes the use of natural language interface for learning with LMs via \lift{}.
We emphasize that our goal is \emph{not} to achieve the state-of-the-art performance, but to investigate thoroughly: (i) what \lift{} can and cannot do, (ii) properties of models fine-tuned via \lift{}, and (iii) whether we can improve \lift{} with advanced techniques.

\section{Methodology and Experimental Setup}\label{sec:methodology}
\textbf{\lift{} training.~} To fine-tune a pretrained LM with \lift{} on a target supervised learning task, we apply two steps: (1) convert each sample into a sentence with a fixed template, and (2) fine-tune LMs with sentence datasets. 
We use the default cross-entropy loss for token prediction in LMs.
Our generic template (without feature names and task description) for sample \prompt{r} is
\vspace{-1mm}
\[
\underbrace{\text{When we have x1=}\prompt{r.x1}\text{, x2=}\prompt{r.x2}\text{, \ldots, xp=}\prompt{r.xp}\text{, what should be y?}}_{\text{question}}\underbrace{\#\#\#}_{\text{q/a separator}} \underbrace{y=\prompt{r.y}}_{\text{answer}}\underbrace{\text{@@@}}_{\text{end of answer}},
\]
if \prompt{r} has $p$ attributes.  
Here, we use the separator convention recommended by OpenAI~\citep{openaimanual} -- ``\#\#\#'' for question/answer separation, and ``\texttt{@@@}'' for end of generation.
When attributes names and task descriptions are available, one can use a more informative prompt (shown in Fig.~\ref{fig:prompts}) with all actual prompts are provided in Sec.~\ref{sec:feature_names}. 
We report learning curves of \lift{} on several tasks in Appendix~\ref{app:training_curves}.

\textbf{\lift{} inference.~} 
For inference, we use the same prompt template except for the answer and end-of-answer parts.
Once the fine-tuned LM completes the test prompt, we simply parse the output tokens.
For classification, we simply compare the generated text with the string representation of the class names.
For regression, we convert the generated string into a number. 
For instance, with the output being ``$y$=10.35\texttt{@@@}\texttt{extratokens}'', we split the output sentence by the stop separator ``\texttt{@@@}'' into two parts.
Taking the first part ``$y$=10.35'', we parse the value ``10.35'' as our final prediction.

The generated output might be \textit{invalid}.
For classification tasks, the output string may not match any actual class, which we declare as misclassification. Note that one may obtain better accuracy by returning its closest class using string metrics.
For regression tasks, we consider output invalid if the string-to-number parsing fails.
In these cases, we adjust the generation randomness by increasing the decoding temperature~\cite{ippolito2019comparison,holtzman2019curious,openai_temperature} from $0$ (deterministic mode) to $0.75$ (random mode). 
We repeat the inference up to five times, then return the average value of the training set if all attempts fail.
Note that invalid output occurs very rarely (less than or around 1\ in most tested cases). 

For evaluation metrics, we use accuracy for classification tasks, and RMSE, RAE errors for regression tasks, where $\text{RAE}= \sum_{i=1}^n |\hat{y}_i - 
    y_i|/\sum_{i=1}^n | \frac{1}{n}\sum_{j=1}^n y_j - y_i|$ and 
    $\text{RMSE}  = \sqrt{\sum_{i=1}^n (\hat{y}_i -y_i)^2/n}$ 
    on each dataset $\gD = \set{(\vx_i, y_i)}_{i=1}^n$ with $n$ samples, features $\vx \in \gX \subset \sR^p$, and outcome $y$.

\textbf{Pretrained LMs.~} 
We apply \lift{} on two pretrained LMs: \gptj{}~\citep{gpt-j} and \gptt{}~\citep{brown2020gpt3}. 
To fine-tune \gptj{}, we use LoRA~\citep{hu2021lora}, a parameter-efficient method that constrains weight matrix updates to be low-rank.
For experiments on \gptj{}, we used \texttt{p3.8xlarge} and \texttt{p3.2xlarge} instances from AWS and RTX3090 GPUs in the local server.
Since \gptt{} is 
not fully publicly available, we use the API provided by OpenAI to perform black-box \gptt{}
fine-tuning.
More details are in Appendix~\ref{app:exp_model_baseline}.

\textbf{Datasets.~}
We design and select a wide range of datasets 
to better understand the behavior of \lift{}.
For \textit{classification}, we use 
three types of non-language data:  low-dimensional synthetic datasets, real tabular datasets in OpenML~\citep{OpenML2013}, and vision datasets (MNIST~\citep{lecun1998mnist}, Fashion-MNIST~\citep{xiao2017fashion} and their permuted variants~\citep{goodfellow2013empirical}).
For \textit{regression}, we use both synthetic and real datasets. For synthetic datasets, we defined samples $(\vx_i, y_i)$ of input-output pair as $\ervy \sim f(\ervx) + \gN(0, \sigma^2)$,
where 
$\sigma^2 \geq 0$ is the noise level. 
Unless otherwise stated, we sample the feature $\ervx$ uniformly from a hypercube $[L,U]^p$, where $L$ and $U$ are minimum/maximum feature values, and $p$ is the number of features.
Following the suggestion by~\citep{xu2020neural}, we consider various functions $f$ for regression tasks: (i) linear function, (ii) quadratic function, (iii) exponential function, (iv) cosine function, (v) $\ell$1-norm function, and (vi) piece-wise linear function. 
Their 2D visualizations are provided in the first row of Fig.~\ref{fig:reg_2d}.
We also use four real datasets: Medical Insurance (Insurance)~\citep{medins},
Combined Cycle Power Plant (CCPP)~\citep{cccp}, 
Servo~\citep{servo}, 
and Student Performance (Student)~\citep{cortez2008using}.
More details are included in Appendix~\ref{app:setup_datatset}.

\textbf{Baselines.~} 
We consider standard learning algorithms~\citep{shalev2014understanding,Borisov2021-uo}.
For classification, we use logistic regression (\textit{LogReg}), decision tree (\textit{DT}), k-nearest neighbor (\textit{\knn{}}), support vector machine with Gaussian kernel (\textit{SVM}), a four-layer ReLU neural network (\textit{\ann{}}) with 200 neurons per hidden layer, random forest (\textit{\randomf{}}), and XGBoost (\textit{\xg{}}). We also use the majority class classifier (\textit{MCC}) that outputs the most dominant class.
For regression, we use polynomial regression (\textit{\qr{}}), kernel ridge regression (\textit{\krr{}}) with radial basis function kernel, $k$-nearest neighbors (\textit{\knn{}}), a three-layer ReLU neural network (\textit{\ann{}}) with 50 hidden neurons per each layer, Gradient Boosting Trees (\textit{\gbt{}}), random forest (\textit{\randomf{}}), and Gaussian process (\textit{\gp{}}).
For hyperparameter selection, we apply the grid search on a set of parameters' values and use cross-validation on the training set (see details in Appendix~\ref{app:setup_model}).

\vspace{-1mm}
\section{Basic Findings of \lift{}}
\label{sec:lift_basics}
\vspace{-1mm}
Table~\ref{tab:summary_basic_findings} summarizes our main findings.
We also study 
sample complexity (Sec.~\ref{sec:sample_complexity}),
comparison with in-context learning (Sec.~\ref{sec:ft_vs_icl}),
models' decision boundaries (Sec.~\ref{sec:inductive_bias}),
and the effect of LMs' pretraining on \lift{} (Sec.~\ref{sec:dependency_lms}).
Appendix~\ref{app:additional_findings} provides additional results, including the effect of input and output layers (\ref{sec:fpt}), model size (\ref{sec:different_gpt}), and \lift{} for Ridge regression (\ref{app:ridge}).

\vspace{-2mm}
\subsection{How Well Does \lift{} Perform on Standard ML Tasks?}
\label{sec:performance}

\vspace{-1mm}
\textbf{Classification.~}
Table~\ref{tab:classification_accuracy} compares classification accuracies between algorithms on a wide range of tasks. 
We observe that \lift{} \bmt{achieves comparable performance to most baselines.}
In most cases, \lift{}/\gpt{} ranks highly in the top three best-performing methods.
We find that \lift{} can learn non-linear relationships between features and the responses:
\lift{}/\gptt{} achieves $81.17\%$ accuracy on the circle dataset, while logistic regression failed to perform better than the MCC ($50\%)$.
As the difficulty of tasks varies, which can be estimated by the average performance of baselines, \lift{} also suffers from performance degradation. 
\lift{} can perform comparably well even when the number of features is as large as hundreds, though the limited number of tokens as inputs to LMs restricts the number of features \lift{} can input.
However, when the number of classes is large (say 100s), both \lift{}/\gpt{} models have lower accuracies than many baselines, though they manage to be better than MCC.
For instance, on the 100-class Margin dataset, the accuracy gap between \lift{}/\gptj{} and the best algorithm (RBF-SVM)  is nearly 30\%.
Note that \lift{} can directly use raw data while most baselines require feature scaling and normalization for good performance.
More results are provided in Appendix~\ref{app:acc_full}, including comparisons with methods leveraging larger models. 

\begin{table}[h!]
\vspace{-.25in}
    \caption{\textbf{Summary of the main findings.}
    }
    \centering
    \small{
    \begin{tabular}{p{2cm}p{11cm}}
    \toprule
        \multicolumn{1}{>{\centering\arraybackslash}p{2cm}}{\textbf{Topic}} &  
       \multicolumn{1}{>{\centering\arraybackslash}p{11cm}}{\textbf{Findings}} \\ \midrule 
        \centering{Overall performance} & 
        On various classification tasks, \lift{} achieves accuracies comparable to strong baselines (Table~\ref{tab:classification_accuracy}). 
        For regression, \lift{}
        well approximates different types of low-dimensional functions (Fig.~\ref{fig:reg_2d}) but does not perform well for high-dimensional cases (Table~\ref{table:reg_all}). \\ \midrule
        \centering{Robustness} & 
        For regression, \lift{} is robust to outliers in training data (Fig.~\ref{fig:app_outliers}). For classification, \lift{} is comparable to baselines under label corruption on training data  (Fig.~\ref{fig:app_label_corruption}) but more vulnerable to feature corruption on test data (Table.~\ref{tab:adv_rob}).
        \\ \midrule
        \centering{Context-aware learning} &
        We can improve \lift{} on classification tasks by designing prompts to specify feature names and the target task. The improvement is significant when the description of the feature names and the target task can be interpreted with common knowledge (Table~\ref{tab:clf_feature_name}).  \\ \midrule
        \centering{Two-stage training} & Warming up \lift{} with pretext tasks using synthetic data improves the prediction performance, especially in the low-data regime (Fig.~\ref{fig:app_pretext}). \\
        \midrule
        \centering{Data augmentation} & For classification tasks, training with augmented data significantly improves the tolerance of \lift{} against perturbed test data (Table~\ref{tab:data_augmentation}).
        \\
        \bottomrule
    \end{tabular} }
    \label{tab:summary_basic_findings}
    \vspace{-3mm}
\end{table}

\begin{table}[t]
\centering
    \caption{
    \textbf{Accuracies ($\uparrow$) on classification datasets.}
    We evaluate \lift{}/\gpt{}s on 2D synthetic data, tabular data in OpenML~\cite{OpenML2013}, and image data, varying number of features ($p$) and data classes ($c$). 
    Overall, \lift{}/\gpt{}s perform comparably well across tasks, adapting to non-linear datasets (circles, two circles) beyond the capacity of logistic regression.
    For OpenML datasets, they achieve competitive performances with the best methods, \textit{e.g.}, XGBoost). 
    The performance degrades when more classes are given, \textit{e.g.}, $c$=100.
    They achieve competitive accuracies on both MNIST and Fashion MNIST.
    Note that MNIST's classes are not fully balanced; thus, MCC achieves 11.35\% instead of 10\%. Table~\ref{tab:classification_accuracy_full} provides the full comparison with all baselines (KNN, MLP, Random Forest). 
    }
    \begin{adjustbox}{width=\textwidth,center}
    \begin{tabular}{lccccccc|cc}
    \toprule[0.05cm]
    \multicolumn{2}{c}{Dataset (ID)} & 
\textbf{$p$ / $c$} & \textbf{MCC} & \textbf{LogReg}   & \textbf{DT}    & \textbf{RBF-SVM}    & \textbf{XG} & \textbf{\lift{}/\gptj{}}  & \textbf{\lift{}/\gptt{}} \\[0.05cm] \midrule 
    \multicolumn{10}{c}{\textbf{Synthetic Data}} \\ \midrule
     & circles (3)  & 2 / 2     & 50.00    & 48.58\relscale{0.8}{$\pm$1.94} & 77.42\relscale{0.8}{$\pm$0.24} & \textbf{83.08\relscale{0.8}{$\pm$0.59}} &  81.42\relscale{0.8}{$\pm$0.31} & 79.95\relscale{0.8}{$\pm$1.53} & 81.17\relscale{0.8}{$\pm$0.42} \bigstrut[t]\\
          & two circles (6)  & 2 / 2     & 50.00    & 49.83\relscale{0.8}{$\pm$4.18} &  75.50\relscale{0.8}{$\pm$0.20} &  {80.00\relscale{0.8}{$\pm$0.54}} &  79.25\relscale{0.8}{$\pm$0.35} & 75.92\relscale{0.8}{$\pm$1.65} & \textbf{81.42\relscale{0.8}{$\pm$0.82}} \\
          & blobs (2)  & 2 / 4     & 25.00    & \textbf{96.75\relscale{0.8}{$\pm$0.00}} & 96.08\relscale{0.8}{$\pm$0.82} &  \textbf{96.75\relscale{0.8}{$\pm$0.00}} & 96.17\relscale{0.8}{$\pm$0.12} & 96.17\relscale{0.8}{$\pm$0.59} & 96.67\relscale{0.8}{$\pm$0.24} \\
          & moons (4)  & 2 / 4     & 50.00    & 88.58\relscale{0.8}{$\pm$0.12} &  99.25\relscale{0.8}{$\pm$0.41} &  \textbf{100.00\relscale{0.8}{$\pm$0.00}} &  99.83\relscale{0.8}{$\pm$0.12} & 99.58\relscale{0.8}{$\pm$0.42} & \textbf{100.00\relscale{0.8}{$\pm$0.00}} \\
          & 9Clusters (1)  & 2 / 9     & 11.25 & 100.00\relscale{0.8}{$\pm$0.00} & 100.00\relscale{0.8}{$\pm$0.00} &  \textbf{100.00\relscale{0.8}{$\pm$0.00}} & \textbf{100.00\relscale{0.8}{$\pm$0.00}} & 99.75\relscale{0.8}{$\pm$0.00} & \textbf{100.00\relscale{0.8}{$\pm$0.00}} \bigstrut[b]\\
    \midrule
   \multicolumn{10}{c}{\textbf{Tabular Data (OpenML)}} \\ \midrule
          & Customers (1511)  & 8 / 2     & 68.18 & \textbf{87.12\relscale{0.8}{$\pm$0.54}} &  85.98\relscale{0.8}{$\pm$0.53} &  {86.36\relscale{0.8}{$\pm$0.00}} & 85.23\relscale{0.8}{$\pm$0.00} & 85.23\relscale{0.8}{$\pm$1.61} & 84.85\relscale{0.8}{$\pm$1.42} \\
          & Pollution (882)   & 15 / 2    & 50.00 & 58.33\relscale{0.8}{$\pm$11.79} &  \textbf{77.78\relscale{0.8}{$\pm$3.93}} &  58.33\relscale{0.8}{$\pm$6.81} &  63.89\relscale{0.8}{$\pm$7.86} & 63.89\relscale{0.8}{$\pm$3.93} & 63.89\relscale{0.8}{$\pm$7.86} \\
          & Spambase (44)   & 57 / 2    & 60.59 & 93.27\relscale{0.8}{$\pm$0.00} &  90.7\relscale{0.8}{$\pm$0.14}  & 93.70\relscale{0.8}{$\pm$0.00} &  \textbf{95.87\relscale{0.8}{$\pm$0.00}} & 94.03\relscale{0.8}{$\pm$0.54}  & 94.90\relscale{0.8}{$\pm$0.36} \\
          & Hill-Valley (1479) & 100 / 2   & 49.79 & {77.78\relscale{0.8}{$\pm$0.00}} &  56.38\relscale{0.8}{$\pm$0.89} & 68.72\relscale{0.8}{$\pm$0.00} &  59.26\relscale{0.8}{$\pm$0.00} & \textbf{100.00\relscale{0.8}{$\pm$0.20}} & \textbf{99.73\relscale{0.8}{$\pm$0.19}} \\
          & IRIS (61) & 4 / 3 & 33.33 & 96.67\relscale{0.8}{$\pm$0.00} & 97.77\relscale{0.8}{$\pm$3.85}  &  \textbf{100.00\relscale{0.8}{$\pm$0.00}}  & \textbf{100.00\relscale{0.8}{$\pm$0.00}}  & 96.67\relscale{0.8}{$\pm$0.00} &   97.0\relscale{0.8}{$\pm$0.00}  \\
          & TAE (48)   & 5 / 3     & 35.48 & 45.16\relscale{0.8}{$\pm$4.56} & {65.59\relscale{0.8}{$\pm$5.49}}  & 53.76\relscale{0.8}{$\pm$6.63} &  \textbf{66.67\relscale{0.8}{$\pm$8.05}} & 61.29\relscale{0.8}{$\pm$6.97} & 65.59\relscale{0.8}{$\pm$6.63} \\
          & CMC (23)   & 9 / 3    & 42.71 & 49.49\relscale{0.8}{$\pm$0.83} &  {56.72\relscale{0.8}{$\pm$0.32}} &  56.50\relscale{0.8}{$\pm$0.97} &  52.43\relscale{0.8}{$\pm$0.42} & 49.83\relscale{0.8}{$\pm$0.28} & \textbf{57.74\relscale{0.8}{$\pm$0.89}} \\
          & Wine (187)  & 13 / 3    & 38.89 & \textbf{100.00\relscale{0.8}{$\pm$0.00}}  & 93.52\relscale{0.8}{$\pm$2.62} &  \textbf{100.00\relscale{0.8}{$\pm$0.00}} &  97.22\relscale{0.8}{$\pm$0.00} & 93.52\relscale{0.8}{$\pm$1.31} & 92.59\relscale{0.8}{$\pm$1.31} \\
          & Vehicle (54)   & 18 / 4    & 25.88 & {80.39\relscale{0.8}{$\pm$1.00}} &  63.92\relscale{0.8}{$\pm$2.37} &  \textbf{81.18\relscale{0.8}{$\pm$0.48}} &  73.14\relscale{0.8}{$\pm$0.28} &
            64.31\relscale{0.8}{$\pm$2.37} & 70.20\relscale{0.8}{$\pm$2.73} \\
          & LED (40496)& 7 / 10     & 11.00  & {68.67\relscale{0.8}{$\pm$0.94}} &  66.33\relscale{0.8}{$\pm$2.87} &  68.00\relscale{0.8}{$\pm$0.82} &  66.00\relscale{0.8}{$\pm$0.82} & 65.33\relscale{0.8}{$\pm$0.47} & \textbf{69.33\relscale{0.8}{$\pm$2.05}} \\
          & OPT (28)   & 64 / 10     & 10.14  & 96.53\relscale{0.8}{$\pm$0.22} & 89.8\relscale{0.8}{$\pm$1.09} &  {97.95\relscale{0.8}{$\pm$0.00}} &  97.48\relscale{0.8}{$\pm$0.17} & 98.22\relscale{0.8}{$\pm$0.11} & \textbf{98.99\relscale{0.8}{$\pm$0.30}} \\
          & Mfeat (12)   & 216 / 10    & 10.00    & 97.67\relscale{0.8}{$\pm$0.12} &  87.67\relscale{0.8}{$\pm$1.05} &  \textbf{98.83\relscale{0.8}{$\pm$0.24}} &  96.75\relscale{0.8}{$\pm$0.00} & {94.17\relscale{0.8}{$\pm$1.75} } & 93.08\relscale{0.8}{$\pm$0.24} \\
          & Margin (1491) & 64 / 100   & 0.94  & {81.35\relscale{0.8}{$\pm$0.15}} &  43.86\relscale{0.8}{$\pm$1.21}  & \textbf{81.98\relscale{0.8}{$\pm$0.30}} & 70.21\relscale{0.8}{$\pm$0.29} & 50.23\relscale{0.8}{$\pm$1.33} & 59.37\relscale{0.8}{$\pm$0.92} \\
          & Texture (1493) & 64 / 100    & 0.94  & {81.67\relscale{0.8}{$\pm$0.97}} &  46.88\relscale{0.8}{$\pm$1.93} & \textbf{83.44\relscale{0.8}{$\pm$0.89}} &  70.73\relscale{0.8}{$\pm$1.41} & 50.32\relscale{0.8}{$\pm$2.18} & 67.50\relscale{0.8}{$\pm$1.42} \bigstrut[b]\\
    \midrule
    \multicolumn{10}{c}{\textbf{Image Data}} \\ \midrule
    & MNIST & \multirow{3}[2]{*}{784 / 10} & 11.35 & 91.95\relscale{0.8}{$\pm$0.69} &  87.42\relscale{0.8}{$\pm$0.64} &  {97.70\relscale{0.8}{$\pm$0.97}} & 97.69\relscale{0.8}{$\pm$0.04} & 97.01\relscale{0.8}{$\pm$1.15} & \textbf{98.15\relscale{0.8}{$\pm$0.67} } \bigstrut[t]\\
          &  Permuted MNIST  &       & 11.35 & 92.58\relscale{0.8}{$\pm$0.04} &  87.87\relscale{0.8}{$\pm$0.69} &  \textbf{98.06\relscale{0.8}{$\pm$0.31}} & {97.62\relscale{0.8}{$\pm$0.09}} & 95.80\relscale{0.8}{$\pm$ 0.07}& 96.25\relscale{0.8}{$\pm$0.35}  \\
           &  Fashion MNIST  &       & 10.00 & 85.59\relscale{0.8}{$\pm$0.09} & 80.52\relscale{0.8}{$\pm$0.40}  &  \textbf{90.59\relscale{0.8}{$\pm$0.02}} & 90.19\relscale{0.8}{$\pm$0.04} & 85.10 \relscale{0.8}{$\pm$ 0.19}& {90.18 \relscale{0.8}{$\pm$0.12}}  \\
          &  Permuted F-MNIST  &       & 10.00    & 84.95\relscale{0.8}{$\pm$0.84} & 79.91\relscale{0.8}{$\pm$0.93} &  {88.04\relscale{0.8}{$\pm$1.69}} &  \textbf{89.93\relscale{0.8}{$\pm$0.14}} & 82.25\relscale{0.8}{$\pm$0.27}  & 88.92\relscale{0.8}{$\pm$0.71}  \bigstrut[b]\\
    \bottomrule[0.05cm]
    \end{tabular}
    \end{adjustbox}
\label{tab:classification_accuracy}
\vspace{-5mm}
\end{table}

\textbf{Regression.~}
Tables~\ref{table:reg_all} and \ref{table:reg_real} present our \textit{function approximation} comparison.
For the low-dimensional cases, \lift{} is comparable to baselines. Still, it fails to beat the strongest baselines, such as \gp{}, as \gpt{} models measure the error by comparing tokens instead of measuring how close the prediction values are to true values.
We conjecture that we can improve our performance by level encoding, {\it i.e.}, representing numerical values as binary values.
We also investigate the \textit{interpolation and extrapolation} of \lift{} and defer the details to Sec.~\ref{app:acc_full}.
All methods fail to extrapolate and interpolate well for all functions, and the interpolation performance of \lift{} is only good in the linear regression case.  
Interestingly, \lift{} tends to output seen values (from training data) for extrapolation. 

\vspace{-1mm}
\subsection{How Many Samples Does \lift{} Need?}
\vspace{-2mm}
\setlength\intextsep{0pt}
\label{sec:sample_complexity}

We investigate whether \lift{} is sample efficient.
Fig.~\ref{fig:sample_complexity} in Appendix shows the sample complexity evaluation on classification and regression tasks.
We find that \gpt{} models can be quickly fine-tuned to learn new tasks with \lift{}.
For \textit{classification}, as the number of classes increases (left to right columns in Fig.~\ref{fig:sample_complexity}a), \lift{} does need more samples for adaptation, probably because the data input and output spaces are more complex to learn.
For \textit{regression} tasks, we find that $1000$ samples are sufficient for \lift{} to have a small RMSE, similar to other baselines. 
There exist some functions (\textit{e.g.}, cosine and piecewise) where \lift{} has lower sample complexity than popular baselines. 

\subsection{Language-Interfaced Learning: \lift{} versus In-Context Learning (ICL)}
\vspace{-2mm}
\label{sec:ft_vs_icl}
\begin{table}[t] 
\vspace{-2mm}
   \caption{
\textbf{Comparison of accuracies ($\uparrow$) between ICL and fine-tuning with \lift{} on OpenML datasets.}
   ``\lift{}/Full-Data'' and ``\lift{}/Subset'' represent \lift{} on the full dataset and and its subset used correspondingly in the ICL setting (number of prompts). Here, the size of subset is chosen to satisfy the LMs' context length. 
   Overall, \lift{}/\gpt{}s on full data achieve the best performances.
    However, when using the same number of samples, \lift{} and ICL are more comparable in most cases.
   Note that both methods may be worse than MCC due to the limited training data in some cases.
   }
  \centering
 \begin{adjustbox}{width=\textwidth,center}
    \begin{tabular}{cc|c|cc|c|cc|cc}
\toprule[0.05cm]
\multirow{2}[2]{*}{Dataset (ID)} & \multirow{2}[2]{*}{\#Prompts} & \multirow{2}[2]{*}{\textbf{MCC}} & \multicolumn{3}{c|}{\textbf{\gptj{}}} & \multicolumn{3}{c}{\textbf{\gptt{}}} &  \bigstrut[t]\\
          &       &       & In-Context & \lift{}/Subset  & \lift{}/Full-data & In-Context & \lift{}/Subset & \lift{}/Full-data &  \bigstrut[b]\\
    \midrule 
    Breast (13)    & 35    & 70.69    & {56.90\relscale{1}{$\pm$19.51}} & \textbf{58.62\relscale{1}{$\pm$2.44}} & 64.94\relscale{1}{$\pm$11.97} & 62.07\relscale{1}{$\pm$1.41} & \textbf{70.69\relscale{1}{$\pm$0.00}} & 71.26\relscale{1}{$\pm$1.62} &  \\
    TAE (48)    & 50    & 35.48 &  \textbf{34.33\relscale{1}{$\pm$1.47}}     & 32.26\relscale{1}{$\pm$9.50} & 61.29\relscale{1}{$\pm$4.56} & \textbf{37.64\relscale{1}{$\pm$4.02}} & 33.33\relscale{1}{$\pm$1.52} & 65.59\relscale{1}{$\pm$6.63} &  \\
    Vehicle (54)    & 14    & 25.88    & \textbf{25.49\relscale{1}{$\pm$0.55}} & 26.04\relscale{1}{$\pm$1.69} & 64.31\relscale{1}{$\pm$2.37} & \textbf{28.82\relscale{1}{$\pm$2.10}} & 23.73\relscale{1}{$\pm$2.27} & 70.20\relscale{1}{$\pm$2.73} &  \\
    Hamster (893)   & 43    &  53.33     & 48.89\relscale{1}{$\pm$3.14} & \textbf{60.00\relscale{1}{$\pm$10.88}} & 55.55\relscale{1}{$\pm$16.63} & \textbf{57.78\relscale{1}{$\pm$6.29}} & 53.33\relscale{1}{$\pm$0.00} & 53.33\relscale{1}{$\pm$0.00} &  \\
    Customers (1511)  & 29    & 68.18 & 56.06\relscale{1}{$\pm$17.14} & \textbf{59.85\relscale{1}{$\pm$2.84}} & 85.23\relscale{1}{$\pm$1.61} & 60.61\relscale{1}{$\pm$1.42} & \textbf{63.26\relscale{1}{$\pm$6.96}} & 84.85\relscale{1}{$\pm$1.42} &  \\    
    LED (40496) & 33    & 68.67    & 10.00\relscale{1}{$\pm$0.82} & \textbf{13.04\relscale{1}{$\pm$3.27}} & 65.33\relscale{1}{$\pm$0.47} & 8.00\relscale{1}{$\pm$1.63} & \textbf{11.33\relscale{1}{$\pm$2.62}} & 69.33\relscale{1}{$\pm$2.05} &  \\
    \bottomrule[0.05cm]
    \end{tabular}
\end{adjustbox}
\vspace{-4mm}
  \label{tab:classification_in_context}
\end{table}
Beyond fine-tuning (with \lift{}), our language-interfaced learning framework can be used for other learning methods for LMs, including in-context learning (ICL)~\citep{reynolds2021prompt,shin2020autoprompt,brown2020gpt3} that performs inference on new tasks without fine-tuning by conditioning on a few training examples.
Table~\ref{tab:classification_in_context} compares the classification performances between (a) ICL, (b) \lift{} trained on a subset with $n$ samples, and (c) \lift{} trained on the full dataset. 
Note that the number of training samples ($n$) used for ICL depends on the context length of given LMs. 
As we can see, \lift{} using the full dataset always achieves the best performances.
However, \lift{}/Subset and ICL are more comparable in most cases when they use the same number of training samples, which are sufficiently small for ICL methods to fit in LMs.

\begin{remark}
One can replace fine-tuning with ICL in our language-interfaced procedure when the target tasks require fewer training samples.
\end{remark}

\vspace{-2mm}

\begin{wrapfigure}[18]{r}{0.53\textwidth}
    \vspace{-0.5mm}
    \centering
    \includegraphics[width=0.47\textwidth]{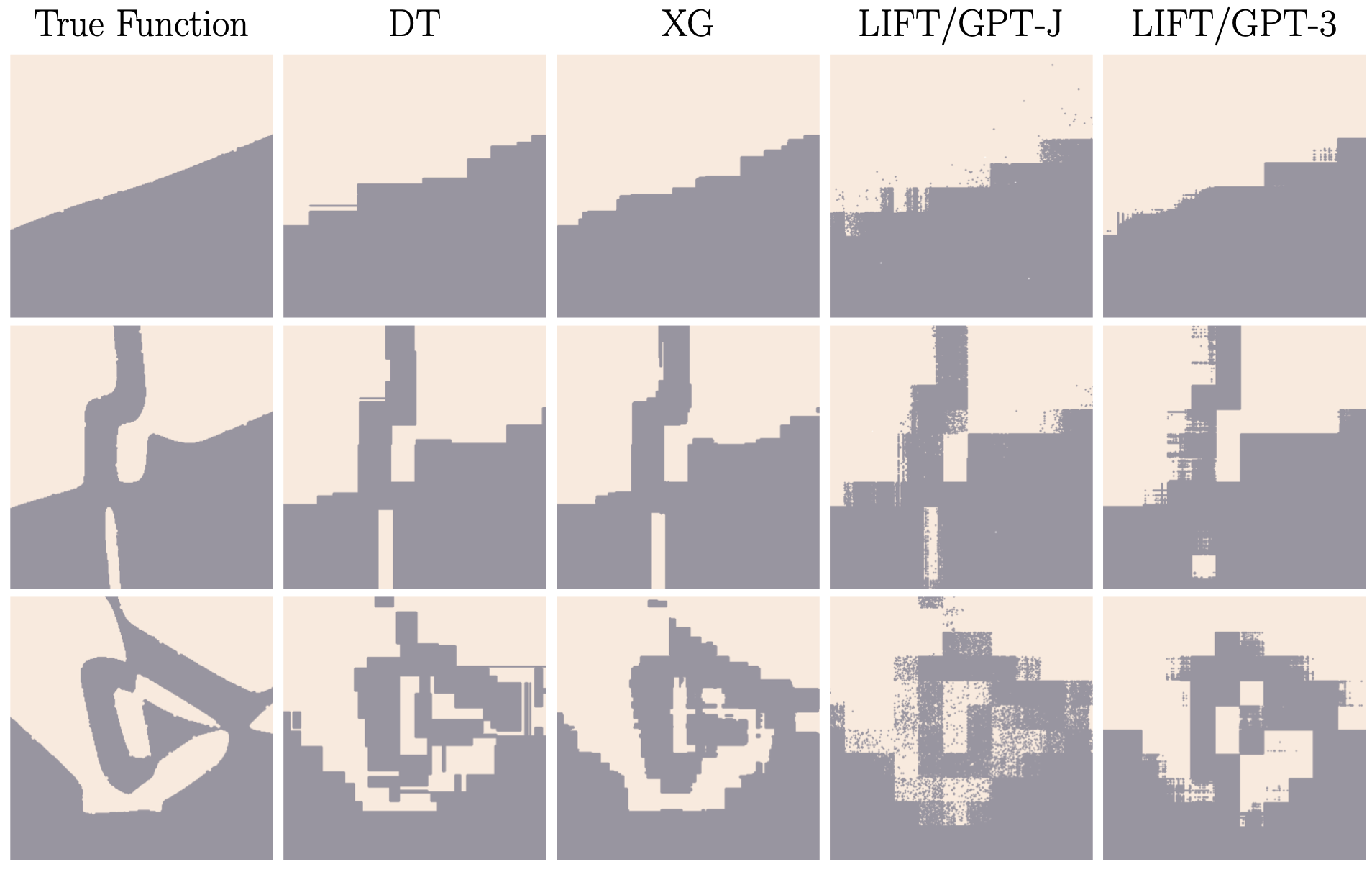}
    \vspace{-2mm}
    \caption{\textbf{Decision boundary visualization.}
    We use three snapshots of a trained network to construct datasets having labels as their predictions (the first column). 
    Top to bottom: snapshots with more training epochs, corresponding to more complex boundaries. 
    \lift{}/\gpt{}s adapt well on different boundaries. }
    \label{fig:clf_decision_boundary_reduced}
    \vspace{-0.1in}
\end{wrapfigure}
\vspace{-1mm}
\subsection{Can We Understand the Inductive Biases of Language Models via \lift{}?}
\vspace{-1mm}
\label{sec:inductive_bias}
To better understand \lift{}/\gpt{}s' inductive biases, we investigate their classification decision boundaries varying the boundaries' complexity, as shown in Fig.~\ref{fig:clf_decision_boundary_reduced}.
We first train a binary-class neural network and use its snapshots at different training epochs to construct datasets having decision boundaries at different complexity levels (first column in Fig.~\ref{fig:clf_decision_boundary_reduced}). 
We observe that \lift{}/\gpt{} models adapt well to three boundaries and capture their rough shapes.
Furthermore, their boundary shapes are axis-parallel, similar to the boundary of tree-based classifiers. They also show a lot of fractals similar to the observations on some convolution neural networks~\citep{somepalli2022can}.
See Appendix~\ref{app:inductive_bias} for results of 3-class and 5-class datasets and quantitative measurements.

\vspace{-1mm}

\subsection{How Robust is \lift{}?}\label{sec:robustness}
We investigate the robustness of \lift{} against the outlier samples
in training data and the feature corruption on test data.
Appendix~\ref{app:robustness} provides additional experimental results, including the robustness for the case of label corruption on training data and class-imbalanced data.

\textbf{Robustness to outliers in training data.~}\label{sec:robutness:outlier}
We consider regression tasks where we have outliers whose outcome $y$ is not consistent with the majority of samples in terms of fitting $(\vx, y)$. 
Fig.~\ref{fig:app_outliers}a compares RAE values of methods with and without outliers (2\% outliers in the training set).
\lift{}/\gpt{} models are among the most robust ones: their performances are almost unaffected, while baselines suffer huge performance drops.
Furthermore, we evaluate models under various percentages of outliers (1\%, 2\%, 5\%, 10\%, 20\%), as shown in Fig.~\ref{fig:app_outliers}b.
Compared to the robust baselines (median-3NN and median-5NN)~\citep{huber2011robust}, \lift{}/\gptt{} is comparably robust, while \lift{}/\gptj{} is more vulnerable when more outliers are present.  

\begin{wraptable}[]{t}{0.35\textwidth} 
\centering
\setlength\tabcolsep{2pt}
\scriptsize
\caption{
\textbf{Accuracies ($\uparrow$) 
under the perturbation on the input feature of MNIST data.} 
See the full results in Table~\ref{tab:adv_rob} in Appendix~\ref{app:robustness}. 
}
\resizebox{0.35\textwidth}{!}{
\begin{tabular}{c|ccc}
\toprule
Source   & 
\multicolumn{3}{c}{\textbf{PGD attack on LeNet-5}} \\ [.05in] 
Target   & 
LeNet-5 & MLP 
& \lift{}/\gptt{}\\
\midrule
$\varepsilon=0$  
& 99.22 & 98.09
& 98.15
\\
$\varepsilon = 0.01$  
& 97.27  & 97.77
&  44.88  
 \\
$\varepsilon=0.1$ 
& 26.80  & 93.99
&   33.66  
\\
$\varepsilon=0.3$  
& 0.00  &   36.62
& 20.31   
\\
\bottomrule
\end{tabular}}
\label{tab:adv_rob_rebuttal}
\end{wraptable}

\textbf{Robustness to feature corruption on test data.~}
Given a clean test data $(\vx,y)$ having feature $\vx$ and label $y$, we explore whether adding small perturbation $\boldsymbol{\delta}$ on the feature changes the performance; we measure the accuracy of \lift{} on perturbed data $(\vx+\boldsymbol{\delta}, y)$.
We apply transfer attack~\citep{kurakin2016adversarial} since we do not have full access to the \gptt{} model, and finding adversarial examples in the discrete input space is complex~\citep{zhang2020adversarial}.
Table~\ref{tab:adv_rob_rebuttal} reports robustness results on MNIST classification under PGD attacks transferred from LeNet-5. 
The perturbation radius is set to $\varepsilon \in [0, 0.01, 0.1, 0.3]$ where MNIST pixel value is within [0,1].
We compare three networks: LeNet-5, MLP (2 hidden layers with 300 and 100 neurons), and \lift{}/\gptt{}. 
When $\eps \in \{0.01, 0.1\}$, \lift{}/\gptt{} tolerates random noise (as in Table~\ref{tab:adv_rob}) but cannot tolerate transferred adversarial attack, 
implying that the adversarial attack on LeNet-5 
is transferred to \lift{}/\gptt{}.

\vspace{-1mm}
\subsection{Does LIFT Need Large-Scale Models Pretrained on Natural Language Data?}
\label{sec:dependency_lms}
\vspace{-2mm}
We investigate the requirement of pretrained LMs for which \lift{}  performs well. 
We compare variants of \lift{} under different types of LMs: \gpt{}s pretrained on natural language data (our models), a large LM without pretraining (Rand-\gptj{}), and LMs pre-trained on non-human language data, including CodeParrot~\citep{codeparrot} and CodeGen-2B-mono~\citep{nijkamp2022conversational} trained mainly on programming language data, and a \gptj{} fine-tuned on Gibberish data~\citep{gibberish}.
See Appendix~\ref{app:dependency_lms} for the detailed setting.

\textbf{Does LIFT only need a large pretrained model?~} 
To answer this question, we compare performances of \lift{} when \gpt{}s are pretrained (\lift{}/\gpt{}s) and when \gptj{} have weights being randomly initialized (\lift{}/Rand-\gptj{}).
More specifically, for \lift{}/Rand-\gptj{},  we randomly initialized a GPT-J model and fine-tuned the whole model (instead of LoRA).
As shown in Table~\ref{tab:lm_comparison}, accuracies of \lift{}/Rand-\gptj{} are much lower than those of our models (\lift{}/\gpt{}s), across all datasets.
These results indicate that \lift{} benefits from pretraining, not just from the large-scale design of LMs.

\textbf{Does LIFT need a model trained on natural language data?~}
As shown in Table~\ref{tab:lm_comparison}, \lift{}/\gpt{}s perform much better than \lift{}/CodeGen and \lift{}/CodeParrot for all datasets.
This implies that \lift{} may perform better with LMs pretrained on natural language data.
When the pretrained \gptj{} is fine-tuned on gibberish data~\citep{gibberish}, the accuracies drop for a few tasks and are lower than \lift{}/\gpt{}s overall.
However, \lift{}/Gibberish still achieves comparably good performance and its small performance gaps to \lift{}/\gptj{} can be attributed to the relatively light impact of fine-tuning on large pretrained LMs.
Thus pretraining on natural language data is necessary for \lift{}.

\begin{table}[t]
  \centering
  \vspace{-.5in}
  \caption{\textbf{Accuracies ($\uparrow$) of LIFT with different LMs. }
  We compare variants of \lift{} with different LMs: \lift{}/\gpt{}s using \gpt{}s pretrained on natural language data (our models), \lift{}/Rand-\gptj{} using a randomly initialized \gptj{},  \lift{}/CodeGen and \lift{}/CodeParrot using LMs pretrained on programming language data, and \lift{}/Gibberish using \gptj{} fine-tuned on gibberish data.
  }
 \resizebox{\textwidth}{!}{
    \begin{tabular}{c|cccccccccc}
 \toprule[0.05cm]
 Dataset (ID)    &  \textbf{MCC}   &  \textbf{\lift{}/GPT-3}      &  \textbf{\lift{}/GPT-J}      &  \textbf{\lift{}/Rand-\gptj{}} & \textbf{\lift{}/Gibberish}  &  \textbf{\lift{}/CodeGen}    &  \textbf{\lift{}/CodeParrot}   \\ \midrule
 Blobs (2)        &  25.00 &  96.67$\pm$ 0.24 &  96.17$\pm$ 0.59 & 25.65$\pm$ 1.58&  96.42$\pm$ 0.24 &  93.67$\pm$ 0.72 &  93.39$\pm$ 1.82  \\
 Two Circles (6)  &  50.00 &  81.42$\pm$ 0.82 &  75.92$\pm$ 1.65 &  49.88$\pm$ 5.01&68.67$\pm$ 1.50 &  53.02$\pm$ 0.66 &  50.08$\pm$ 2.47 \\
 Iris (61)        &  33.33 &  97.0$\pm$ 0.00  &  96.67$\pm$ 0.00 &  27.78$\pm$ 20.79&94.44$\pm$ 1.57 &  43.31$\pm$ 6.67 &  60.00$\pm$ 8.82    \\        
 Customers (1511) &  68.18 &  84.85$\pm$ 1.42 &  85.23$\pm$ 1.61 & 52.47$\pm$ 7.15 &67.43$\pm$ 1.42 &  45.96$\pm$ 8.96 &  43.11$\pm$ 3.34  \\
 Wine (187)       &  38.89 &  92.59$\pm$ 1.31 &  93.52$\pm$ 1.31 & 22.22$\pm$ 15.71 &84.26$\pm$ 3.46 &  77.78$\pm$ 0.00 &  33.88$\pm$ 3.87   \\
 LED (40496)      &  11.0  &  69.33$\pm$ 2.05 &  65.33$\pm$ 0.47 & 11.68$\pm$ 4.44 &72.67$\pm$ 1.25 &  11.00$\pm$ 4.00 &  23.46$\pm$ 13.85  \\  
    \bottomrule[0.05cm]
    \end{tabular}
    }
\vspace{-4mm}
\label{tab:lm_comparison}
\end{table}

\section{Evaluation of \lift{}-Specific Learning Properties}
\label{sec:lift_specific}
\vspace{-1mm}
In this section, we study the behavior of \lift{} in a more fine-grained manner. 
\vspace{-2mm}
\subsection{Does \lift{} Benefit from Incorporating Feature Names?}\label{sec:feature_names}
\vspace{-1mm}

Unlike standard machine learning algorithms, \lift{} can be provided context information by incorporating the feature names and task descriptions in the prompts. 
Intuitively, this incorporation may improve the sample complexity of \lift{} as the prior knowledge already learned in the pretraining phase may help \lift{} predict better.
We design seven prompt templates to assess how incorporating feature names affects the performance of \lift{} (see more details in Appendix~\ref{app:context}).
We empirically verify this intuition and show our results in Table~\ref{tab:clf_feature_name} for several classification tasks using pretrained \gptt{} models.
We first observe that all \lift{} models outperform MCC with significant accuracy gaps, indicating that they are all properly trained.
Second, we observe that correctly incorporating feature names helps boost the performances of \lift{} for datasets except for \texttt{CMC}.
Third, if we use similar prompts with shuffled feature names (\tuan{\texttt{Shuffled-Names I, II}}), then the performance of \lift{} drops by a significant margin. 
These results imply that the aforementioned performance improvements are indeed due to proper prompting with correct feature/value association.

\begin{table}[t]
\vspace{-.1in}
   \caption{\textbf{The effect of using feature names on \lift{}}. We compare classification accuracy ($\uparrow$) of \lift{}/\gptt{} when feature names provided in the target dataset \textit{are and are not} incorporated into the prompts.
  We provide four versions of \lift{} when feature names are correctly incorporated (Correct-Names columns) and when feature names are randomly shuffled (Shuffled-Names columns).
  We evaluate models on three OpenML datasets, including \texttt{CMC (23)}, \texttt{TAE (48)}, \texttt{Vehicle (54)}, and \texttt{German}.
  We also compare our models with two baselines: the \mcc{} (MCC) and XGBoost.
  As a result, all \lift{} models achieve better performance than MCC.
  Among the evaluated models, \lift{}s with correct feature names achieve the best accuracies on both \texttt{TAE}, \texttt{Vehicle}, and \texttt{German} datasets while achieving the comparable accuracies to the best model on the \texttt{CMC} dataset.
  \textit{*Two designs of the prompt format result in the same template for the} \texttt{Vehicle} \textit{dataset.}
  } 
  \centering
\resizebox{\textwidth}{!}{
    \begin{tabular}{c|c|cccccc}
    \toprule[0.05cm]
    \multirow{2}{*}{Dataset (ID)} & \multirow{2}{*}{\textbf{MCC}} & \multicolumn{6}{c}{{\textbf{\lift{}}}} \bigstrut[t]\\
    [0.1cm]
          &  &  W/o Names I & W/o Names II & Shuffled-Names I & Shuffled-Names II & Correct-Names I &   Correct-Names II  \bigstrut[b]\\[0.05cm]
    \midrule
    {CMC (23)} &  {42.71}  &  \textbf{57.74$\pm$0.89} & 57.40$\pm$1.37 & 56.27$\pm$2.06 & 57.06$\pm$4.24 & 57.40$\pm$1.09 & 56.27$\pm$2.22 \bigstrut[t]\\
    
    {TAE (48)}  & {35.48}  &  65.59$\pm$6.63 & 66.67$\pm$5.48 & 60.22$\pm$6.72 & 64.52$\pm$8.53 & \textbf{69.89$\pm$9.31} & \textbf{69.89$\pm$6.72} \\
    
    {Vehicle (54)} &  {25.88} & 70.20$\pm$2.73 & 71.96$\pm$3.09 & 70.20$\pm$5.34 & 69.22$\pm$2.72 &   \multicolumn{2}{c}{\textbf{75.29$\pm$2.04*}}\\
    
    German & 70.00 &  $71.33 \pm 5.20$& $67.83 \pm 2.72$ &  $73.00\pm1.87$&  $71.67\pm0.94$& $72.33\pm1.70$&  \textbf{74.17$\pm$ 1.25}\\ 
    \bottomrule[0.05cm]
    \end{tabular}
}
\label{tab:clf_feature_name}
\vspace{-5mm}
\end{table}

\vspace{-1mm}
\subsection{\bmt{Is \lift{} Well Calibrated?}}
\label{sec:Bayesian}
\vspace{-1mm}

We investigate whether \lift{} is \textit{calibrated}, \ie, the prediction reflects the confidence, by exploring how \lift{} performs under various noise levels, as shown in 
Fig.~\ref{fig:bayesian} in Appendix.
We conduct experiments on six synthetic regression datasets, each consisting of 1,000 noisy training samples shown as blue markers in the first row. 
To be specific, we generate (1) the input $x$ following the guideline in Sec.~\ref{app:setup_datatset} for regression tasks and (2) the noisy outcome $y$ where the standard deviation of noise $\sigma(x)=(x+10)/10$ increases along the $x$-axis (from $x=-10$ to $x=10$), and study how different noise level affects the predictive behavior of \lift{}.
In the inference phase, we set the decoding temperature $T=1$ for \lift{} to make random predictions. 
For visualization purposes, we generate an additional 103 samples uniformly in $[-10,10]$ for each task and plot the standard deviation of 20 \lift{}/\gptj{} predictions on each sample in the bottom row of Fig.~\ref{fig:bayesian}.  
Note that the bottom row of Fig.~\ref{fig:bayesian} shows that the standard deviation of \lift{}/\gptj{}'s prediction nearly matches that of noisy training samples (observations) across different tasks. 
These results imply that \lift{}/\gptj{} is calibrated.
Similarly, Fig.~\ref{fig:bayesian_gpt3} of Sec.~\ref{app:bayesian} shows that \lift{}/\gptt{} is calibrated.

\begin{wrapfigure}[15]{r}{0.5\textwidth}
    \vspace{-0.15in}
    \centering
    \begin{minipage}{0.5\textwidth}
    \centering\captionsetup[subfigure]{justification=centering}
    \includegraphics[width=0.9\textwidth]{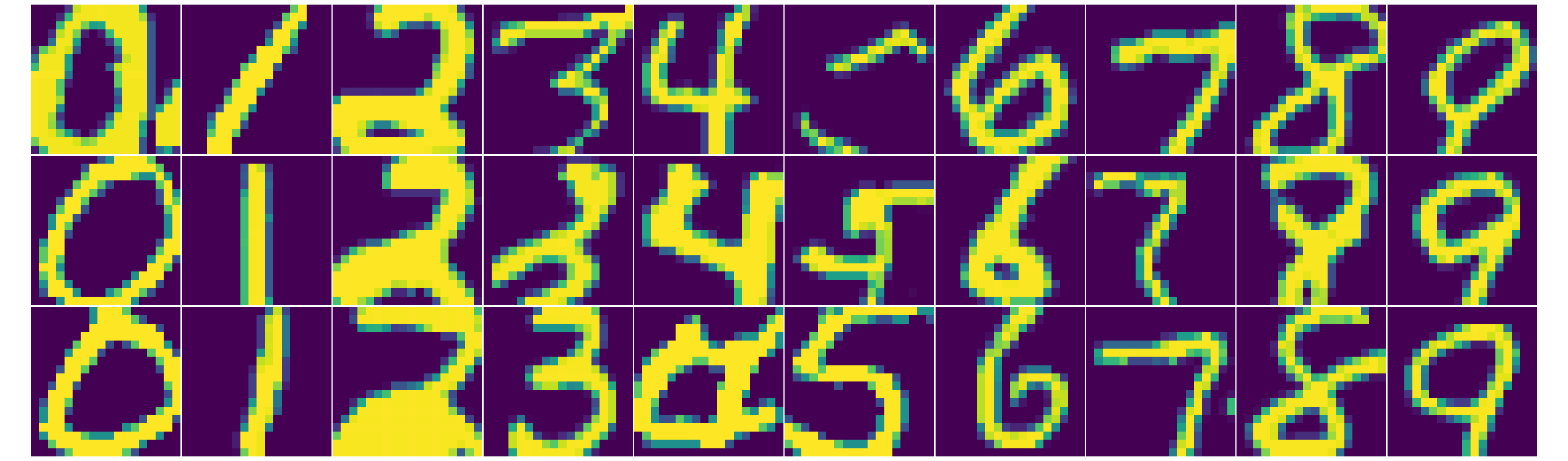}
    \subcaption{Given only the digit number.}
    \label{fig:a}\par\vfill
    \includegraphics[width=0.9\textwidth]{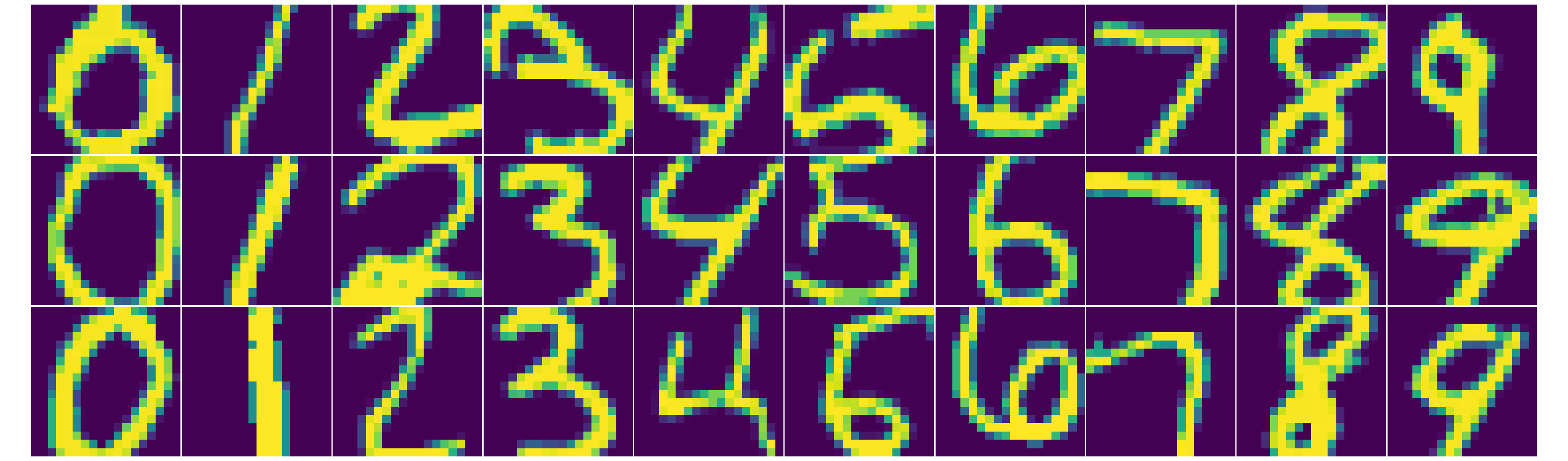}
    \subcaption{Given the digit number and a half of image pixels.}
    \label{fig:b}
    \end{minipage}
\vspace{-2mm}
\caption{
\textbf{Generating MNIST images using \lift{}/\gptj{}.}
We observe that \lift{}/\gptj{} can generate images of comparably high quality. The temperature is set to 1.
}
\label{fig:lift_generated_mnist}
\end{wrapfigure}

\vspace{-1mm}
\subsection{Can We Use \lift{} for Data Generation?}\label{sec:generator}
\vspace{-1mm}
Generative models have been widely used in computer vision~\citep{goodfellow2020generative,kingma2019introduction,croitoru2022diffusion}.
Beyond classification and regression tasks, we study whether \lift{} can be used for generative tasks, \ie, learning the underlying data distribution and generating realistic data samples. 
In particular, we consider two image generation tasks on MNIST dataset: (a) generating an image given a digit number, and (b) completing an image given a digit number and its pixels on the top half of the image.
Fig.~\ref{fig:lift_generated_mnist}a and Fig.~\ref{fig:lift_generated_mnist}b show our generated images for the two tasks respectively.
We observe that the generated images have the correct digit shape and reasonably high quality in most cases, especially for the image completion (Fig.~\ref{fig:lift_generated_mnist}b).
See Appendix~\ref{app:generator} for more details.

\section{Improving \lift{} with Existing Techniques}
\label{sec:lift_improve}
\vspace{-1mm}

We improve \lift{} with advanced techniques: two-stage fine-tuning and data augmentation.

\vspace{-1mm}
\subsection{Two-Stage Fine-Tuning for \lift{} with Synthetic Pretext Tasks}
\vspace{-1mm}
\label{sec:improve_two_stage}
\begin{wrapfigure}[12]{r}{0.5\textwidth}
\vspace{-1mm}
    \centering
    \includegraphics[width=0.5\textwidth]{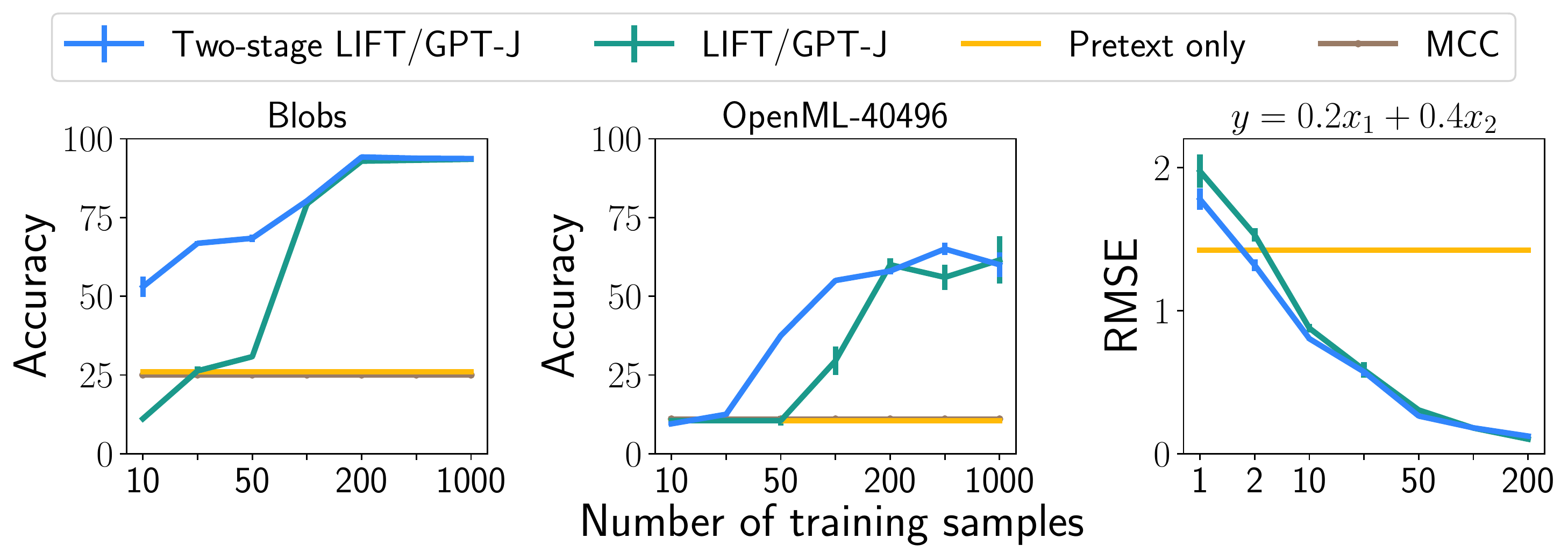}
    \vspace{-.25in}
    \caption{\textbf{Two-stage fine-tuning.}
    The two-stage method (blue) applies \lift{} first on synthetic pretext data before the real datasets, outperforming fine-tuning (green) when training data is small.
    The full experiment results are presented in Fig.~\ref{fig:app_pretext}.
    }
    \label{fig:pretext}
\end{wrapfigure}
In Sec.~\ref{sec:sample_complexity}, we observe that LMs need a sufficient number of samples to start adapting.
We suspect that LMs' adaptation to non-language tasks contains two phases: (1) learn the task description, \textit{i.e.,} input space, label space, and sentence templates~\citep{reynolds2021prompt,min2022rethinking}, and (2) learn the target task. 
Thus, we consider utilizing synthetic data to describe the task for LMs in the first phase, thus reducing the sample complexity.
This results in a new two-stage training procedure for \lift{}.\footnote{The recent work~\citep{chen2022improving} demonstrates the usefulness of the intermediate fine-tuning method for LMs. However, they focus on self-supervised objectives for fine-tuning pretrained LMs for few-shot in-context learning.}
In particular, for any given dataset, we first generate two pretext tasks with simple synthetic Gaussian datasets (discussed in \ref{app:setup_datatset}) sharing the same number of features and the label space (for classification tasks) or the range of responses' values (for the regression tasks) to the actual data.
We apply \lift{} on pretext tasks for a few (2 or 3) epochs, then continue \lift{} with the target (given) dataset.
For \gptt{}, it is unclear how to keep the order of samples not shuffled with the current black-box API during the fine-tuning stage.
Hence, we only provide the experimental results of \gptj{}.
Fig.~\ref{fig:pretext} shows that two-stage  fine-tuning improves \lift{} over the original fine-tuning when the number of training samples is small on both classification and regression tasks. 

\vspace{-1mm}
\subsection{Data Augmentation}
\label{sec:improve_augmentation}

\begin{wraptable}[17]{t}{0.6\textwidth}
\vspace{-0.25in}
\caption{
\textbf{Accuracies ($\uparrow$) of \lift{} with/without data augmentation (DA), as well as baselines (LeNet-5, MLP) on MNIST.} Each row represents different ways of training, and each column means different test data. 
Data augmentation (DA) means that we are using a noisy version of MNIST training data by adding Gaussian noise. Given an MNIST image having range [0,1], the noise is added in the $L_{\infty}$ ball with radius $\eps$.
One can confirm that the data augmentation significantly improves the tolerance of \lift{}/\gptj{} against perturbed test data in both Gaussian and signed constant noise.
For each column, we boldfaced the highest value among baselines and the highest value among \lift{}/\gptj{}.
}
\centering
\setlength\tabcolsep{4pt}
\footnotesize
\begin{adjustbox}{width=0.6\textwidth,center}
\begin{tabular}{l|c|cc|cc}
\toprule[0.04cm]
& \textbf{Clean} & \multicolumn{2}{c|}{\textbf{Gaussian noise}} & \multicolumn{2}{c}{\textbf{Signed const. noise}} \\
& $\eps=0$ & $\eps=0.01$ & $\eps=0.1$ & $\eps=0.01$ & $\eps=0.1$ \\
\midrule
LeNet-5 & \textbf{99.22} & \textbf{99.25} & \textbf{99.20} & \textbf{99.26} & \textbf{99.06} \\
MLP & 98.09 & 98.05 & 97.70 & 98.08 & 97.39 \\
\midrule
\lift{}/\gptj{} 
& \textbf{96.88} 
& \textbf{95.27} 
& 56.14 
& 55.83 
& 27.73 
\\
\lift{}/\gptj{}, DA (Gaussian, $\eps=0.05$) 
& 93.80 
& 94.39 
& 93.40 
& 93.46 
& 61.24 
\\
\lift{}/\gptj{}, DA (Gaussian, $\eps=0.1$) 
& 93.78 
& 94.31 
& \textbf{94.98} 
& \textbf{94.12} 
& \textbf{75.25} 
\\
\bottomrule[0.04cm]
\end{tabular}
\end{adjustbox}
\label{tab:data_augmentation}
\end{wraptable}
\vspace{-1mm}
Data augmentation~\citep{shorten2019survey} is a simple tool for improving the generalization performance for various classification problems. 
Here, we investigate whether data augmentation benefits \lift{}. Table~\ref{tab:data_augmentation} shows the effect of adding random noise in the training data on the performance of \lift{}/\gptj{} for the MNIST classification problem. 
Here, we test each model on three settings: (1) clean data, (2) Gaussian noise, and (3) signed constant noise. 
We allow each noise can perturb up to the magnitude of $\eps \in [0,1]$ at each dimension (\textit{i.e.}, each pixel) when the black/white pixel of MNIST is represented in the $[0,1]$ range. 
We defer the generation procedure of random Gaussian noise to Sec.~\ref{sec:feat_corrupt} in Appendix.

One can observe that \lift/\gptj{} without any data augmentation (DA) is vulnerable to random noise, unlike existing baselines (LeNet-5 and MLP). However, when we apply data augmentation, \textit{i.e.}, train \lift/\gptj{} with noisy training data, the accuracy improves significantly for the perturbed (either adding Gaussian noise or Signed constant noise) test data. This shows the effectiveness of simple data augmentation in \lift{}. Exploring the effect of other data augmentation schemes, \textit{e.g.}, mixup~\cite{zhang2017mixup} and its variants~\cite{yun2019cutmix,kim2020puzzle,sohn2022genlabel}, is remained an interesting future work.

\section{Related Works}
\textbf{Fine-tuning for adapting LMs to non-language tasks.~}
Fine-tuning~\citep{li2017learning} pretrained LMs is the standard practice for learning downstream tasks, which may involve simple architecture modifications, such as adding linear layers~\citep{donahue2016adversarial,chen2020simple} or freezing layers~\citep{lu2021pretrained,perez2018film,chen2018self}.
The recent progress focuses on \emph{parameter-efficient} techniques for reducing trainable parameters, including adapter-based fine-tuning~\citep{houlsby2019parameter,rebuffi2017learning,wang2020k}
that trains additional small residual blocks between layers,
freezing-based fine-tuning~\citep{gheini2021cross,lu2021pretrained,dinh2022improved} that freezes most of the pretrained parameters, and distillation-based fine-tuning~\citep{chen2019distilling}. 
Our \lift{}/\gptj{} is fine-tuned with LoRA~\citep{hu2021lora}, a parameter-efficient method approximating the weight updates using low-rank matrices.

To directly adopt existing fine-tuning methods of LMs for non-language tasks, it is common practice to modify the input/output layers and the loss functions, which may cause undesired behaviors like catastrophic forgetting~\citep{li2017learning,chen2020recall}. 
\tuan{Our work is highly motivated by} Frozen Pretrained Transformer (FPT)~\citep{lu2021pretrained} that directly fine-tunes GPT-2~\citep{radford2019language} pretrained on language tasks for other modalities by freezing most pretrained parameters and adding only input and output layers for the modality adaptation.
Unlike FPT, our method requires no such changes in the architecture and objective function.
Several works also extend the existing LMs to handle different input data types, such as images~\citep{lu2019vilbert,tsimpoukelli2021multimodal}, audio~\citep{chuang2019speechbert}, tabular data~\citep{liu2021tapex}, and knowledge base~\citep{agarwal2020knowledge} by updating the pretraining phase with these data and their corresponding tasks or using general-purpose architecture~\citep{jaegle2021perceiver}. 
Our work is based on \gpt{} language models trained \emph{only} on textual data.

\textbf{Analyzing the adaptability of LMs.~} 
Similar to ours, recent works~\citep{abnar2021exploring,bigbench,lego,bigbench,hao2022language} attempt to understand and quantify the adaptability~\citep{li2021quantifying} and capacity of large LMs, such as Big-Bench~\citep{bigbench} with a new benchmark of more than 200 tasks on a diverse set of topics.

\textbf{General-purpose models.~}
A primary goal of our work is to push the limit of the existing generalist language models (\textit{e.g.,} \gptt{}~\citep{brown2020gpt3}) to other modalities and domains, supporting the idea of building a domain-and-modality agnostic generalist model~\citep{brown2020gpt3,rae2021scaling,chowdhary2020natural,li2020unicoder,alayrac2022flamingo,radford2021learning,jia2021scaling,reed2022generalist}.
Note that \lift{} can be applied to any generalist model with LM-like architectures, such as GATO~\cite{reed2022generalist}.
Furthermore, our work shares the general goal with automated machine learning (AutoML)~\citep{feurer-neurips15a,feurer-arxiv20a} in improving the usability of machine learning, though \lift{} uses only a single pretrained LMs for all tasks while
AutoML automates the standard machine learning pipeline from a set of existing algorithms.

\section{Discussion and Conclusion}\label{sec:dis_con}
\vspace{-1mm}
We propose the use of \textbf{language-interfaced} framework, via  \textbf{Language-Interfaced Fine-Tuning (\lift{})}, for using LMs to solve non-language downstream tasks without changing the models' architecture or loss function.
\lift{} first converts labeled samples into sentences and then fine-tunes pretrained LMs on the sentence dataset using the standard fine-tuning method and loss function.
Via an extensive empirical study, we show that \lift{}/\gpt{} performs relatively well on low-dimensional classification and regression non-language tasks.
Furthermore, \lift{}/\gpt{}s are robust in several practical settings, and can properly calibrate the predictions and generate realistic data samples.
\lift{} can be improved using in-context feature names, two-stage fine-tuning, and data augmentation.
Moreover, our work is arguably one of the \textit{first to thoroughly study the efficacy of language-interfaced learning framework} with pretrained language models on standard regression and classification tasks, paving the way for enabling ``no-code machine learning with language models.''

\textbf{Limitations and open questions.~}
Despite promising performances on various tasks and settings, we observe some limitations of \lift{} to basic learning tasks.
\lift{}/\gpt{} do not perform well if the features have high dimensions (for regression) or when the number of classes is large (for classification).
In addition, the context length of \lift{} is restricted to the context length of LMs and \lift{}/\gpt{} is memory-inefficient. 
One can combine \lift{} with memory-efficient LMs such as LinTransformer~\citep{wang2020linformer} to address this issue. 
Besides, our works open some interesting questions for future works. First, do LMs and \lift{}/\gpt{} have behaviors similar to ensemble methods or decision tree since 
they have similar decision boundaries? Secondly, are LMs universal models that can adapt well to any modalities and domains? Lastly, can \lift{}/\gpt{}s adapt better for regression tasks using more sophisticated encoding schemes for numeric features?

\textbf{Social impacts.~} Future research should also investigate potential fairness issues of applying \lift{}. Based on large language models, \lift{} might have embedded bias targeting certain social groups. Especially when feature names are included in the training prompts, the models may be more sensitive to social biases and thus might make unfair and harmful predictions. 
\bmt{We leave measuring the embedded bias in LIFT as one of the interesting future directions. }

\newpage
\section*{Acknowledgements}
This work was supported by NSF Award DMS-2023239, the Understanding and Reducing Inequalities Initiative of the University of Wisconsin-Madison, and the Office of the Vice Chancellor for Research and Graduate Education with funding from the Wisconsin Alumni Research Foundation.
\bibliography{refs}
\bibliographystyle{unsrt}

\section*{Checklist}

\begin{enumerate}

\item For all authors...
\begin{enumerate}
  \item Do the main claims made in the abstract and introduction accurately reflect the paper's contributions and scope?
    \answerYes{The contribution and scope of this paper has been summarized in the abstract and the last part of Introduction.}
  \item Did you describe the limitations of your work?
    \answerYes{We wrote the limitations in Sec.7.}
  \item Did you discuss any potential negative societal impacts of your work?
    \answerYes{This has been discussed in Sec.7.}
  \item Have you read the ethics review guidelines and ensured that your paper conforms to them?
    \answerYes{Our paper conforms to the ethics guideline.}
\end{enumerate}

\item If you are including theoretical results...
\begin{enumerate}
  \item Did you state the full set of assumptions of all theoretical results?
    \answerNA{}
    \item Did you include complete proofs of all theoretical results?
    \answerNA{}
\end{enumerate}

\item If you ran experiments...
\begin{enumerate}
  \item Did you include the code, data, and instructions needed to reproduce the main experimental results (either in the supplemental material or as a URL)?
    \answerYes{The code, data, and instructions are provided in the shared anonymous GitHub repository.}
  \item Did you specify all the training details (e.g., data splits, hyperparameters, how they were chosen)?
    \answerYes{All the training details are discussed in the paper and they can be found at the anonymous GitHub repository.}
        \item Did you report error bars (e.g., with respect to the random seed after running experiments multiple times)?
    \answerYes{All the experiments are run multiple times and at the tables mean and standard deviation of those runs are presented.}
        \item Did you include the total amount of compute and the type of resources used (e.g., type of GPUs, internal cluster, or cloud provider)?
    \answerYes{This has been described in Sec.3.}
\end{enumerate}

\item If you are using existing assets (e.g., code, data, models) or curating/releasing new assets...
\begin{enumerate}
  \item If your work uses existing assets, did you cite the creators?
    \answerYes{Our work uses public datasets/models, and has been cited properly, in the paper and GitHub repo.}
  \item Did you mention the license of the assets?
    \answerNo{}
  \item Did you include any new assets either in the supplemental material or as a URL?
    \answerNA{}
  \item Did you discuss whether and how consent was obtained from people whose data you're using/curating?
    \answerNA{}
  \item Did you discuss whether the data you are using/curating contains personally identifiable information or offensive content?
    \answerNA{}
\end{enumerate}

\item If you used crowdsourcing or conducted research with human subjects...
\begin{enumerate}
  \item Did you include the full text of instructions given to participants and screenshots, if applicable?
    \answerNA{}
  \item Did you describe any potential participant risks, with links to Institutional Review Board (IRB) approvals, if applicable?
    \answerNA{}
  \item Did you include the estimated hourly wage paid to participants and the total amount spent on participant compensation?
    \answerNA{}
\end{enumerate}

\end{enumerate}

\appendix
\newpage

{\large \textbf{Appendix}} \par 
Due to the space limit, many interesting findings, discussions, and further details are included in Appendix. 
We start by discussing the detailed motivation of our work in (\ref{sec:motivation}) and introducing related work in (\ref{sec:related_works}).
We then describe our datasets in (\ref{app:setup_datatset}) and the implementation of all models in (\ref{app:setup_model}).
For the experimental results, we provide detailed and extended results for all findings presented in the main paper in (\ref{app:full_findings}), with more visualizations, score tables, and additional evaluations on more functions, datasets, and properties.
In (\ref{app:additional_findings}), we present additional interesting findings and visualizations which have not been discussed so far in the main paper. 
Lastly, we provide a more profound discussion in (\ref{app:discussion}).

\startcontents[sections]
\printcontents[sections]{ }{1}{}
\newpage

\section{Further Motivations of \lift{}}\label{sec:motivation}
While we mainly emphasized the ``no-code-ML'' property of \lift{}, indeed, it has a lot of potentials to be more useful and powerful than many of the current ML models. Particularly, \lift{} can bring an entirely novel approach to enable (i) explainability and (ii) updatability via information retrieval. 

\subsection{Explainability} 
Most ML models cannot interpret their predictions. While specific algorithms are developed to enable such models' explainability, their efficacy is still in question.
On the other hand, LMs have been shown to be able to explain their predictions ~\citep{wei2022chain,kojima2022large}.
This ability has been used in many fields such as human moral judgements~\citep{jin2022make} and mathematical reasoning~\citep{cobbe2021training}.
Therefore, \lift{}, based on a large pretrained LM, can be made to explain its prediction using its reasoning capabilities. 

Consider the German-credit dataset~\citep{Dua:2019}, which aims to predict whether the bank should approve or decline loan applications. After asking \gptt{} if one should approve/decline a loan application via \lift{}, one can also ask \gptt{} to explain its own prediction result. This can be implemented by making two consecutive inference calls, as shown in Table~\ref{tab:motivation}. Here, we provide five different responses generated with different random seeds.

\begin{table}[!htbp]
    \vspace{-0.4in}
    \caption{
    \textbf{An example illustrating the reasoning capability of large pretrained language models.}
    We convert a data sample from German-credit dataset~\citep{Dua:2019} into text and use \gptt{} to predict whether the bank should approve or decline the individual's loan application. 
    After receiving the first response of ``accept'', we add ``The bank will accept this loan application. Let's think logically. This is because'' to the original prompt and use it as the new input of \gptt{}.
    The five different responses generated by \gptt{} with different random seeds demonstrate the reasoning ability of language models. 
    Note that these explanations are generated without seeing any reasoning examples, \ie, these are
    solely based on the zero-shot reasoning capability of language models. 
    }
    \centering
    \resizebox{\textwidth}{!}{
    \begin{tabular}{p{1.5cm}|p{12cm}}
    \toprule
         &  \textbf{Prompt or Generated Text}\\ \midrule
        Raw input &  Checking Account: less than 200 DM; Loan Duration: 48 months; Credit History: no credits/paid back duly; Loan Purpose: education; Number of applying for credits: 18424; Savings Account: less than 100 DM; Employment Time: 1~4 years; Installment Rate: 1\%; Sex: Female; Other Debtors: No; Property: building society savings agreements/life insurance; Age: 32; Housing: self-owned house; Job: highly qualified employee/ self-employed; Foreigner: no\\ \midrule
        First prompt (soliciting prediction) & A 32-year-old female is applying for a loan of 18424 credits for 48 months for other purposes. She has a checking account with less than 200 Deutsche Mark and a savings account with less than 100 Deutsche Mark. She had no credits or paid back all credits duly. The installment rate of this applicant in percentage is 1\%. She is employed for more than one year but less than four years and is recognized as a highly qualified employee or is self-employed. She has lived in a self-owned house. She owns some building society savings agreements or life insurance. Given such information, will the bank accept or decline this loan application? \\ \midrule
        First response & accept \\ \midrule
        Second prompt (soliciting explanation) & A 32-year-old female is applying for a loan of 18424 credits for 48 months for other purposes. She has a checking account with less than 200 Deutsche Mark and a savings account with less than 100 Deutsche Mark. She had no credits or paid back all credits duly. The installment rate of this applicant in percentage is 1\%. She is employed for more than one year but less than four years and is recognized as a highly qualified employee or is self-employed. She has lived in a self-owned house. She owns some building society savings agreements or life insurance. Given such information, will the bank accept or decline this loan application? \\ &  \textbf{The bank will accept this loan application. Let’s think logically. This is because} \\ \midrule
        Second response (random seed 1) & the applicant is employed and is qualified in a high level. Since she is employed, she is able to make all the payments. Furthermore, she has a self-owned house and owns some building society savings agreements or life insurance. She has less than 2000 Deutsche Mark in checking and savings account. She had no credits or paid back all credits duly. She is able to pay the loans. All of these are positive factors to this loan. The bank will accept this loan. \\ \midrule
        Second response (random seed 2) & the applicant has a high income. She has lived in a self-owned house. She has some savings agreements or life insurance. She is a highly qualified employee. She has no negative credits. All of these facts show that she is not in danger of not being able to pay back the loan. Therefore, the bank will accept this loan application. \\ \midrule
        Second response (random seed 3) & the applicant has a checking account with less than 200 Deutsche Mark and savings account with less than 100 Deutsche Mark. Hence she is a low-risk customer. Furthermore, she has no credits or paid back all credits duly. And the installment rate of this applicant is 1\%. She is employed for less than four years and is recognized as a highly qualified employee. She has lived in a self-owned house. She owns all the building society savings agreements or life insurance. Hence she is considered a good customer. So the bank will accept her loan application \\ \midrule
        Second response (random seed 4) & the applicant is employed with a high salary and is economically stable. She is not likely to file for bankruptcy. She can pay back the loan. All in all, we can see that the bank will accept this loan application. \\ \midrule
        Second response (random seed 5) & the bank will accept all applications with a very low risk of default. There is a very low probability that she will be unemployed in the near future. She can afford to pay off this loan. Furthermore, this applicant has a high annual income. \\ 
        \bottomrule
    \end{tabular}
    }
    \label{tab:motivation}
\end{table}

Note that these explanations are generated without seeing \emph{any} reasoning examples, \ie, these are solely based on the zero-shot reasoning capability of language models. 
While these explanations are not even close to perfect, they are still reasonable.
Improving the few-/zero-shot reasoning capabilities of language models is a fast-growing research field, so incorporating new techniques can further improve the explainability of \lift{}. 

\subsection{Updatability via Information Retrieval}
Another drawback of current ML models is that it is difficult to ``update'' models when a distribution shift occurs. Handling these distribution shifts in a compute-efficient manner has recently become one of the most active research areas in the field. 
Recently, augmenting LMs~\citep{izacard2020leveraging,singh2021end,yogatama2021adaptive,borgeaud2022improving} with a retrieval mechanism has shown to be an efficient way of updating LMs. With such a retrieval mechanism equipped, LMs can be efficiently updated as one can simply update its associated database, or even connect LMs to the Internet~\citep{borgeaud2022improving}. 

While we only used LMs not equipped with a retrieval mechanism in this paper, it is straightforward to apply our framework to other LMs that can retrieve information from databases or the Internet, \ie, \lift{} can support a compute-efficient update mechanism.

\section{Detailed Related Works}\label{sec:related_works}
This section provides detailed related work.

\paragraph{Pretraining and adapting language models (LMs). }
Our work uses modern large LMs, which promoted significant advances in the field of natural language processing (NLP)~\citep{bommasani2021opportunities}.
Most popular LMs use transformer architectures~\citep{vaswani2017attention,so2019evolved} as the backbone, from early models like BERT~\citep{devlin2018bert} built on Transformer encoders to GPT variants~\citep{radford2018improving,radford2019language} built on Transformer decoders.
Multiple modern large LMs have been proposed, including RoBERTa~\citep{liu2019roberta}, ALBERT~\citep{lan2019albert}, XLNet~\citep{yang2019xlnet}, and the latest models with billions or trillions of parameters, such as \gptt{}~\citep{brown2020language}, Switch-Transformers~\citep{fedus2021switch}, and  PALM~\citep{chowdhery2022palm}.
 
LMs are trained to encode a large amount of linguistic knowledge from multiple sources in their contextual representations~\citep{li2022survey}, which are helpful and can be easily adapted to various other tasks.
Thus, starting with BERT~\citep{devlin2018bert}, it has become a standard practice to pretrain and then fine-tunes a large LM for plenty of downstream tasks in lieu of training a model from scratch for a specific task~\citep{devlin2018bert,brown2020gpt3, bommasani2021opportunities,liu2019roberta, li2022pre}.
This technique dramatically impacts a wide range of NLP tasks, such as language modeling~\citep{brown2020gpt3,chowdhery2022palm}, question answering systems ~\citep{mccann2018natural,su2019generalizing}, text summarization~\citep{aksenov2020abstractive}, neural machine translation~\citep{jauregi-unanue-piccardi-2020-pretrained}, and reasoning~\citep{liu2021gpt,suadaa2021towards}.

However, the excellent performance of fine-tuned LMs without architecture changes has been mainly limited to NLP tasks so far. 
This work, instead, investigates whether we can leverage LM fine-tuning for non-language tasks across different modalities.
\tuan{Our work is highly motivated by} Frozen Pretrained Transformer (FPT)~\citep{lu2021pretrained}, which directly adapts GPT-2~\citep{radford2019language} pretrained on language tasks and textual data to other modalities.
FPT freezes most pretrained parameters except the layer normalization layers and adds input and output layers for fine-tuning.
The authors empirically show that GPT-2 can be efficiently fine-tuned for different modalities and domains, including vision and numeric computation.
Nevertheless, FPT requires changes in the architecture and objective function to adapt to different data representations, while our method \lift{} does not.
Furthermore, we mainly focus on basic machine learning tasks, such as function approximation or tabular data classification.

Several works have attempted to extend the existing LMs to handle different types of input data, such as images~\citep{lu2019vilbert,tsimpoukelli2021multimodal}, audio~\citep{chuang2019speechbert}, tabular data~\citep{liu2021tapex}, and knowledge bases ~\citep{agarwal2020knowledge} by updating the pretraining phase with these data and their corresponding tasks.
For instance, XGPT~\citep{xia2021xgpt} takes images as the input and uses the image captioning task as the primary task in the pretraining stage for GPT to generate images' captions.
Similarly, multiple works utilize pretrained LMs for generating text descriptions of images or videos in image captioning (VisualGPT~\citep{chen2021visualgpt}) and video captioning (VideoBERT~\citep{sun2019videobert}, Unified VLP~\citep{zhou2020unified}, UniVL~\citep{luo2020univl}).
SpeechBERT~\citep{chuang2019speechbert} also integrates LMs for speech recognition in the weakly-supervised setting to reduce the need for scarce supervised data.
LMs can also adapt to numeric tasks~\citep{born2022regression} or other domains such as protein folding~\citep{jumper2021highly} or symbolic math solvers~\citep{noorbakhsh2021pretrained}.
Recent works~\citep{herzig-etal-2020-tapas,iida2021tabbie,liu2021tapex} pretrain LMs with large tabular datasets, improving the question answering systems by reasoning over the tables. 
Compared with these existing works, our work is unique in that it is based on \gpt{} language models trained \emph{only} on textual data.

\paragraph{Analyzing the adaptability of LMs.} 
Similar to our work, recent works~\citep{li2021quantifying,abnar2021exploring} have made efforts to understand and quantify the feasibility and limitations of the adaptability of large LMs for upstream performance and downstream tasks.
For instance, the recent work~\citep{li2021quantifying} built a benchmark of 500 small language tasks for testing the adaptability of LMs, observing that the LMs~\cite{JMLR:v21:20-074} can adapt well to an extensive range of complex tasks rather than just memorizing the learned patterns.
BIG-bench~\citep{bigbench} is recently introduced as a new benchmark for quantifying the capacity of LMs, consisting of more than 200 tasks on a diverse set of topics.
Another relevant work~\citep{lego} attempts to understand the effect of LM pretraining by studying how the transformer architecture, the backbone of LMs, succeeds at a designed synthetic task. 
Similar efforts in this line of  work are to analyze the behaviors, representations, and inductive bias of pretrained LMs~\citep{hupkes2018visualisation,warstadt2020learning,lovering2020predicting} or investigate different aspects of LMs~\citep{nogueira2021investigating,lample2019deep,wallace2019nlp}.
For instance, a recent work~\citep{nogueira2021investigating}, in the investigation of the difficulty of numeracy in LMs observes that transformer-based language models do not work well on complex numeric tasks and are sensitive to different formats of numeracy. 
Note that these existing works focus primarily on downstream language tasks, while we focus on adapting LMs on non-language tasks without any modification of the loss or architecture.

\paragraph{Methods for adapting LMs.}
The most common method for adapting LMs is \textit{fine-tuning}~\citep{li2017learning} which aims to slightly adjust pretrained parameters for learning the downstream tasks~\citep{zhang2019dialogpt,ribeiro2020investigating}.
Fine-tuning can involve simple architecture modifications, such as adding linear layers~\citep{donahue2016adversarial,chen2020simple} or freezing parts of the network~\citep{lu2021pretrained,perez2018film,chen2018self}.
Fine-tuning can be improved with advanced techniques, such as multi-stage methods~\citep{phang2018sentence}, intermediate fine-tuning~\citep{moghe2021cross,mao2019improving,fabbri2020improving}, or self-supervised training~\citep{rubino2020intermediate}.
The recent progress in fine-tuning LMs focuses on the \emph{parameter-efficient} techniques for minimizing the number of fine-tunable parameters, including adapter-based fine-tuning~\citep{houlsby2019parameter,rebuffi2017learning,wang2020k} that adds and trains small residual blocks between transformer layers, freezing-based fine-tuning~\citep{gheini2021cross,lu2021pretrained,dinh2022improved} that freezes most of the pretrained parameters and fine-tunes only tiny parts of the networks, and distillation-based fine-tuning~\citep{chen2019distilling}. 
In this line of work, LoRA~\citep{hu2021lora} further reduces the number of trainable parameters in large LMs by approximating the weight updates using low-rank matrices without changing pretrained parameters.
LoRA is used as the fine-tuning method for \gptj{} in our \lift{}/\gptj{} framework. 
To directly adopt these fine-tuning methods used in LMs for non-language tasks, it is common practice to modify the input/output layers and the loss functions.
However, these modifications might lead to undesired behaviors like catastrophic forgetting~\citep{li2017learning,chen2020recall}. 
On the other hand, our method \lift{} uses the language interface for fine-tuning without making any changes to the architecture or the loss function.

In-context few-shot learning paradigm~\citep{raffel2019exploring,liu2021makes,lu2021fantastically,min2021recent,scao2021many} suggests modifying only the inputs of LMs by adding a few examples of the downstream task.
A critical part of these methods is reformulating the downstream task samples to the language modeling inputs~\citep{reynolds2021prompt,min2022rethinking}, resulting in multiple efforts in generating~\citep{guo2021text}, searching~\citep{shin2020autoprompt}, and properly tuning the prompts~\citep{li2021prefix,lester2021power,liu2021p}.
While these methods have shown great effectiveness for multiple NLP downstream tasks, it is unclear how to apply them to downstream tasks of other modalities.
On the other hand, our work successfully adapts LMs for non-language tasks, further pushing the application boundaries of large-scale LMs.

\paragraph{Deep learning for tabular datasets. }
While deep neural networks have been successfully applied to various data types, such as images or text, they still face difficulties with a few classification and regression tasks on tabular data~\citep{borisov2021deep,padhi2021tabular}, one of the most popular data types  in practice.
This may be due to the heterogeneous nature of tabular data, with their features being sparse, type-mixed, and weaker in correlation than natural image-language data~\citep{shwartz2022tabular,borisov2021deep}.
Multiple deep learning methods and architectures have been proposed for tabular datasets, from making discrete decision trees more differentiable~\citep{hazimeh2020tree,popov2019neural}, regularizing neural networks' weights~\cite{shavitt2018regularization,kadra2021regularization}, to recent attempts using attention-based modules~\citep{huang2020tabtransformer,arik2021tabnet,somepalli2021saint}.
Though these transformer-based models are the closest works to us in this line of work, their works focus on designing and improving architecture designs for specifically learning the tabular data rather than adapting the LMs.
To the best of our knowledge, we are the first to thoroughly study large LM adaptation for tabular learning without architecture changes.
Our work shows promising results of LMs in closing the gap to the best-performing methods, including tree-based ensemble algorithms (Random Forest~\citep{ho1995random} or XGBoost~\citep{chen2015xgboost}).

\paragraph{General-purpose models (generalist models).}
A primary goal of our work is to push the limit of the existing generalist language models (\textit{e.g.,} \gptt{}~\citep{brown2020gpt3}) to other modalities and domains, supporting the idea of building a domain-and-modality agnostic generalist model. 
Early works~\citep{reed2015neural,kaiser2017one,keskar2019ctrl} explored this idea by developing and training multi-task and multi-modal models on a wide range of diverse tasks to obtain better generalization and adaptation.
The development of large-scale LMs has significantly contributed to the area of generalist models for languages~\citep{brown2020gpt3,rae2021scaling,chowdhary2020natural}, vision~\citep{li2020unicoder}, visual language~\citep{alayrac2022flamingo,radford2021learning,jia2021scaling}, and control problems~\citep{reed2022generalist}.
These generalist models are usually trained with the scale on an extensive range of corpora, probably containing multiple modalities and domains.
In this line of work, a general-purpose architecture~\citep{jaegle2021perceiver} has also been studied to handle different input and output data types.
Moreover, a recent work proposes a new language model as a general-purpose interface to connect multiple pretrained foundation models. This new language model is shown to work well on both language-only and vision-language tasks~\citep{hao2022language}.
Although \lift{} primarily focuses on the LMs, it can be applied to other generalist models with LM-like architectures, such as GATO~\cite{reed2022generalist}.
Furthermore, it is worth noticing that our work shares the general goal with automated machine learning (AutoML)~\citep{feurer-neurips15a,feurer-arxiv20a} in improving the usability of machine learning.
AutoML automates the standard machine learning pipeline for model selection and hyperparameter tuning from a set of existing algorithms. At the same time, \lift{} uses only a single pretrained LM for solving all tasks.

\section{Experimental Setup}\label{app:experiment_setup}

\subsection{Datasets}
\label{app:setup_datatset}

\paragraph{Classification datasets.} 

\begin{figure}[h]
    \centering
    \includegraphics[width=.8\textwidth]{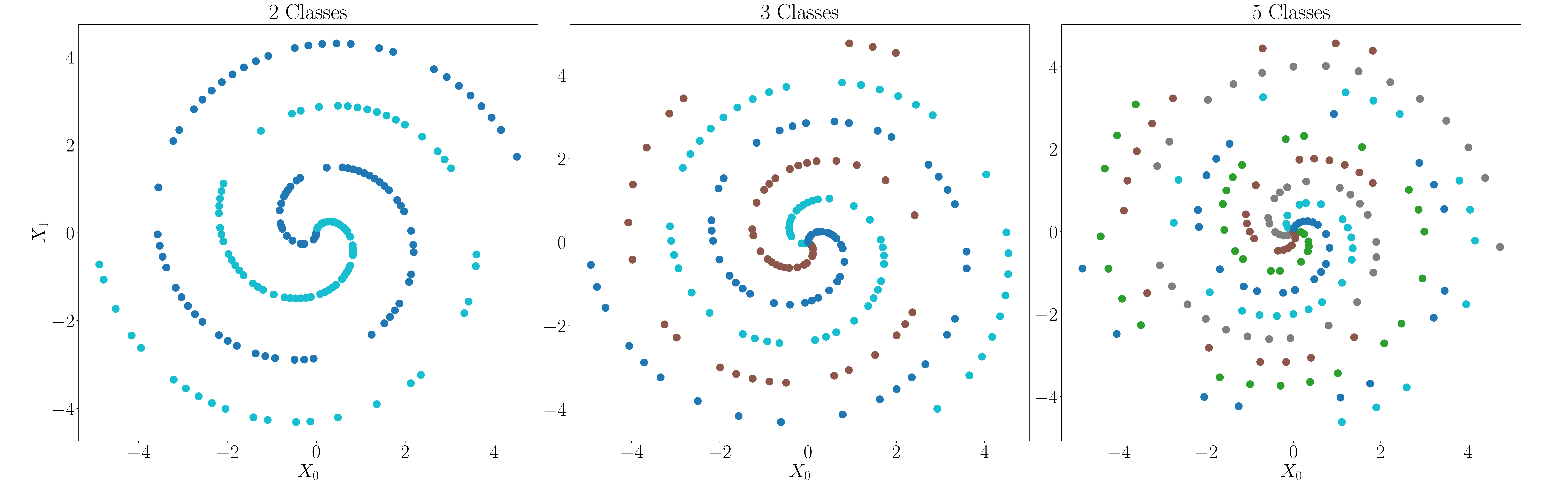}
    \vspace{-2mm}
    \caption{\textbf{Rolls dataset of 2, 3, 5 classes with 300 samples per dataset (all classes are balanced).
    }
    }
    \label{fig:nn_syn}
    \vspace{2mm}
\end{figure}

\begin{figure}[h]
    \vspace{4mm}
    \centering
     \includegraphics[width=\textwidth]{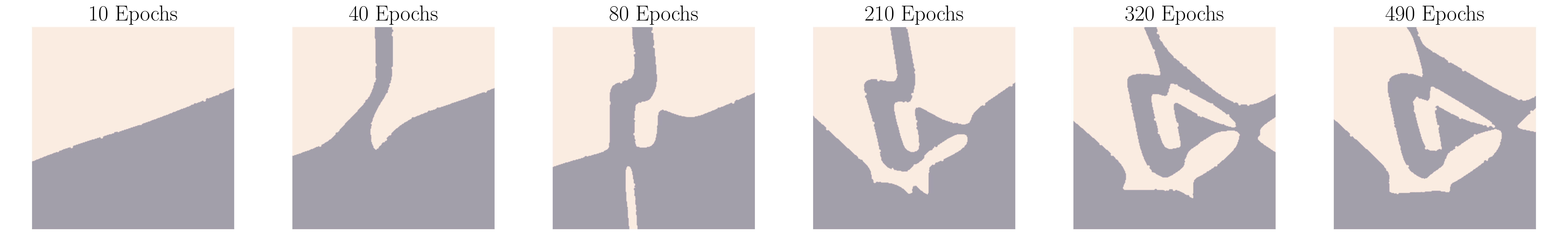}
    \caption{\textbf{Neural-net-based datasets.} Given \texttt{Rolls} dataset in Fig.~\ref{fig:nn_syn}, we train a 2-layer neural network for 10, 40, 80, 210, 320, 490 epochs, and get six decision boundaries at each column. We define six neural-net-based datasets from here: each decision boundary is used as a labeling function of each neural-net-based dataset. In the main manuscript, we used three out of six datasets, obtained by snapshots at 10, 80, and 490 epochs.
    }
    \label{fig:nn_syn_ex}    
    \vspace{4mm}
\end{figure}

Table~\ref{tab:clf_datasets} summarizes the datasets used for classification tasks.
We use two additional types of synthetic datasets: \textbf{neural-net-based datasets} used for understanding the inductive biases of algorithms (in Sec.~\ref{sec:inductive_bias} and Appendix~\ref{app:inductive_bias}) and \textbf{Gaussian pretext datasets} for the two-stage fine-tuning experiments (in Sec.~\ref{sec:improve_two_stage}).
The \textit{neural-net-based datasets} are generated as follows. For binary classification, we train a 2-layer neural network with \texttt{tanh} activation functions using the \textit{Rolls} dataset shown in the leftmost figure in Fig.~\ref{fig:nn_syn}, and take six snapshots of the decision boundary of the trained neural network shown in Fig.~\ref{fig:nn_syn_ex}; we took snapshots at training epochs 10, 40, 80, 210, 320 and 490. Then, for each snapshot, we define a synthetic dataset (what we call a neural-net-based dataset) having labels as the neural network's prediction for randomly chosen 2000 samples. 
For 3-class and 5-class classifications, we also use 2000 samples to train a 2-layer neural network using the \textit{Rolls} dataset shown as the second and the third figure in Fig.~\ref{fig:nn_syn}. 
The decision boundaries of networks trained on more epochs are visually more complex. Hence, the corresponding classification tasks are becoming more complicated, from the left column to the right column in Fig.~\ref{fig:nn_syn_ex}. In the manuscript, we tested on three out of six datasets, obtained by snapshots at 10, 80, and 490 epochs, respectively. 
Given a target dataset of $n$ classes and $d$ features, \textit{Gaussian pretext datasets} are generated as follows: using~\texttt{scikit-learn}\footnote{https://scikit-learn.org/stable/modules/generated/sklearn.datasets.make\_classification.html}, we randomly generate datasets of $n$ clusters, where each cluster has 100 normally distributed samples in the $d$-dimensional space. 

({\bf Image datasets}) For MNIST-based datasets, as the context length of LMs is limited, we (center) scale and crop the size of images from $28\times28$ into $18\times18$ and use the integral format of pixel values $[0, 255]$. 
We convert each image into a sequence of pixels in the order of left to right and top to bottom.
Each pixel sequence is then converted into the sentence input as of our language-interfaced framework.

\paragraph{Regression datasets.} We test \lift{}/\gpt{} on regression problems for both synthetic/real datasets. \\
({\bf Synthetic datasets})
To assess the regression performance of \lift{} in different datasets, we generate synthetic datasets based on six different  functions types, including smooth functions, non-smooth functions, and non-continuous functions: 
\begin{enumerate}
    \item Linear functions $y = f(\vx) =  \vx^\top \vone / p$
    \item Quadratic functions $y = f(\vx) = \vx^T \mI \vx /p$, where $\mI$ is the identity matrix
    \item Continuous exponential function $y = f(\vx) = \sum_{i=1}^p e^{0.2 x_i}/p$
    \item Cosine functions $y = f(\vx) = \sum_{i=1}^p \cos(0.5\pi x_i) /p$
    \item (Non-smooth) $\ell 1$-norm function $y = f(\vx) = \norm{\vx}_1/p$
    \item (Non-continuous) Piecewise linear function 
    \begin{equation}
f(\rvx) = \frac{1}{p} \sum_{i=1}^p \Tilde{f}(x_i) :=  \left\{\begin{array}{cc}
x_i-1 & -10 \leq x_i<-3 ,\\
0 & -3 \leq x_i<3, \\
x_i+1 & 3 \leq x_i \leq 10.
\end{array}\right\}.
    \end{equation}
\end{enumerate}
We let $\rx_i \sim \operatorname{Unif}(-10, 10)$ for each coordinate $i$, and the noise level $\sigma = 0.1$ by default.
We normalize all the functions above for fair comparison among different functions so that $y \in [-9,9]$ when $\vx\in [-10,10]^p$.
In particular, to assess whether \lift{} is better at dealing with positive numbers or integers, we generate additional datasets by further manipulating the  $(\rvx, \ry)$ distribution of linear and piecewise functions. 
For datasets with real numbers, we generate the 1D dataset $\ervx \sim \operatorname{Unif}(-150,-150)$.
For datasets with only positive numbers, we generate the dataset $\ervx_i \sim \operatorname{Unif}(0,300)$.
To generate the datasets with all integer prompts, we round down all the features to integers. 
For visualization, in addition to the training, validation, and test datasets, we generated grid datasets. 
Unless otherwise stated, we generate uniformly spaced 200 samples for 1D visualizations and 2,500 samples for 2D visualizations, with each coordinate $\ervx_i \in [-10, 10]$ for $i=1,\dots, p$. 
To visualize the extrapolation performance, we let the $\ervx_i \in [-15,15]$.

\begin{table}[t] 
    \caption{\textbf{Classification datasets.} We have three non-language types of data: synthetic data, real tabular data, and vision data.
    We use five synthetic datasets.
    For the real tabular data, we select datasets from OpenML with a wide range of number of features, types of features, number of classes, and number of training samples.
  We use MNIST, Fashion MNIST, and their permuted variants for the vision datasets.}
    \centering
    \scriptsize
\resizebox{\textwidth}{!}{
    \begin{tabular}{l|l|lccccc}
        \toprule[.5mm]
        \textbf { Data Type } & \textbf { Dataset } & \textbf{ID} & \textbf{Abbreviation} & \textbf { No. Features } & \textbf {No. Classes } & \textbf {No. Instances }  & \textbf { Note } \\ 
        \midrule
        \multirow{5}{*}{Synthetic}
        & 9Gaussians & 1 & - & 2& 9& 2000&- \\ 
        & Blobs  & 2& - &  2 & 4 &  2000&- \\
        & Circle  & 3 & - & 2& 2& 2000& non-linear boundary \\
        & TwoCircles & 6 & - & 2& 2& 2000& non-linear boundary \\ 
        & Moons & 4 & - & 2& 2& 2000&- \\
        \midrule
        \multirow{13}{*}{Tabular}
        & wholesale-customers & 1511 & Customers  & 8& 2 & 440& Imbalance \\
        & pollution & 882 &  Pollution & 15 & 2 & 60&  1 symbolic feature\\
        & spambase & 44 & Spambase & 57 & 2 & 4601 & 1 symbolic feature\\
        &  hill-valley & 1479 & Hill-Valley & 100 & 2 & 1212 & 1 symbolic feature\\
        & tae & 48 & TAE & 5 & 3 & 151 & Categorical data \\
        & cmc & 23 & CMC & 9& 3& 1473& Meaningful feature Names\\
        & wine & 187 & Wine & 13& 3&178 & Integral features\\ 
        (OpenML) & vehicle & 54 & Vehicle & 18& 4& 846& Meaningful feature Names\\ 
        & LED-display-domain-7digit & 40496& LED & 7 & 10 & 500 & 1 symbolic feature \\
        & optdigits & 28 & OPT&64 & 10 & 5620 & 1 symbolic feature\\
        & mfeat-factors & 12 & Mfeat &  216 & 10 & 2000& 1 symbolic feature\\
        & pollen & 871 & Pollen & 5 & 2 & 3848 & - \\
        & climate-model-simulation-crashes & 1467 & Climate & 20 & 2 & 540 & -\\
        & one-hundred-plants-margin & 1491& Margin & 64 & 100 & 1600 & 1 symbolic feature\\
        & one-hundred-plants-shape & 1492 & Shape & 64& 100 & 1600& 1 symbolic feature\\
        & one-hundred-plants-texture & 1493 & Texture& 64& 100 & 1599& 1 symbolic feature\\
        & breast-cancer & 13 & Breast & 9 & 2 & 286 & - \\ 
        & iris & 61 & Iris & 4 & 3 & 150 & - \\ 
        & visualizing\_hamster & 893 & Hamster & 5 & 2 & 73 & - \\
        & PizzaCutter3 & 1444 & Pizza & 37 & 2 & 1043 & - \\
        \midrule
        \multirow{3}{*}{Vision} & MNIST & - & - & 784& 10& 70k&- \\
         & Permuted MNIST & - & P-MNIST  & 784& 10& 70k&- \\
          & Fashion MNIST & - & FMNIST & 784& 10& 70k&- \\
         & Permuted Fashion MNIST & - & P-FMNIST & 784& 10& 70k&- \\
        \bottomrule[.5mm]
    \end{tabular}}
    \vspace{1mm}
    \label{tab:clf_datasets}
\end{table}

({\bf Real datasets}) We consider four real datasets: Medical Insurance dataset~\citep{medins} with 1,338 samples and 6 features, 
Combined Cycle Power Plant  (CCPP) dataset~\citep{cccp}  with 9,568 samples and 4 features, Servo~\citep{servo} dataset with 167 samples and 4 features,
and Student~\citep{cortez2008using} dataset with 649 samples and 33 features. 
The Medical Insurance and Student datasets contain feature names that can be interpreted using common knowledge (see feature lists in Table~\ref{table:reg_context}), while CCPP and Servo do not. 

\subsection{\lift{} and Baseline Implementation}
\label{app:setup_model}

This section provides details of our models and implementation.
We describe our pretrained language models and  the baseline implementations (\ref{app:exp_model_baseline}), the computing resources used for running experiments (\ref{app:computing_resource}), and how to fine-tune and select the hyperparameters (\ref{app:hyperparameter}).

\subsubsection{Pretrained Language Models and Baselines}
\label{app:exp_model_baseline}
\paragraph{Pretrained language models.} Our main results are with two pretrained language models: \gptj{}~\citep{gpt-j} and \gptt{}~\citep{brown2020gpt3}. 
We mainly focus on \gptj{} for the reproducibility purpose and provide additional results on \gptt{} as reference. 
For \gptj{}, 
we use a quantized version\footnote{https://huggingface.co/hivemind/gpt-j-6B-8bit} of 6 billion parameters with 8-bit weights.
For \gptt{}, we use the \gptt{} OpenAI API\footnote{https://openai.com/api/} and employ the \texttt{Ada} version by default. 
In Sec.~\ref{sec:different_gpt}, we compare two previously mentioned models with three bigger versions of \gptt{} (\texttt{Baggage, Curie, Davinci}). The largest one is
\texttt{Davinci}-\gptt{} containing approximately $175$ billions of parameters.

Since \gpt{}s can output any language token, the output might not be appropriate for the desired task. For example, \gpt{} might output non-numerical words for regression task approximating a function $f: \mathbb{R}^n \rightarrow \mathbb{R}$. In such a case, we categorize this as \textit{invalid output}. 

\paragraph{Baselines.} 
For XGBoost, we use the open-source \texttt{XGBoost} module\footnote{https://xgboost.readthedocs.io/}.
For other baselines, we implement them using \texttt{scikit-learn} module\footnote{https://scikit-learn.org/}.

\subsubsection{Computing Resources}
\label{app:computing_resource}
For experiments on \gptj{}, we use GPU computing from two 24GB-RTX3090 GPUs and AWS EC2 instances\footnote{https://aws.amazon.com/ec2/} (\texttt{p3.8xlarge},  \texttt{p3.2xlarge}).
For other models, we run experiments on CPU instances.

\subsubsection{Hyperparameter Selection} 
\label{app:hyperparameter}
In fine-tuning \gptj{}, we use Adam-8bit optimizer implementation\footnote{https://huggingface.co/hivemind/gpt-j-6B-8bit} with weight decay of $0.01$ and $6$ warm-up steps.
The learning rate is chosen from \texttt{1e-4} and \texttt{2e-4} for the synthetic/OpenML datasets, and \texttt{1e-5} for the vision datasets.
We use a linear learning scheduler for the optimizer.
For classification, the batch size depends on the number of features of the datasets.
We set the batch size to be 128, 32, 16, and 2 for datasets with the number of features being no greater than 2, between 2 and 6, between 6 and 20, and greater than 20, respectively.
For regression, we set batch size as $4$ by default and reduce it to 1 to avoid the memory issue when the number of features increases. 
For \gptt{},   we use the API provided by OpenAI to perform black-box \gptt{}
fine-tuning with the default setting.
Our implementation is with PyTorch framework\footnote{https://pytorch.org/}.

We perform hyperparameter selection based on validation results for all methods for a fair comparison. 
The hyperparameter tuning scheme for all methods is detailed as follows:
\paragraph{Classification methods.}
    \begin{itemize}
        \item \lift{}/\gptj{}: number of epochs $\in \set{5,10,15}$
        \item \lift{}/\gptt{}: learning rate multiplier $\in \set{0.05, 0.1, 0.2}$
        \item Random Forest (\textit{\randomf{}}): maximum depth $\in \set{3,5,10}$, minimum number of samples required to split an internal node $\in \set{2,5,10}$
        \item Decision Tree (\textit{DT}): maximum depth of the tree $\in \set{3,5,20}$ and criterion $\in \set{\text{Gini impurity, Shannon information gain}}$
        \item Support Vector Machine (\textit{SVM}): kernel $\in \set{\text{linear kernel, radial basis function}}$, regularization parameter $\in \set{1, 10, 100}$
        \item Multilayer Perceptron (\textit{\ann{}}): initial learning rate $\in \set{0.001, 0.01, 0.1}$
        \item Logistic regression (\textit{LogReg}): inverse of regularization strength $\in \set{1, 10, 100}$
        \item K-Nearest Neighbor (\textit{\knn{}}): number of neighbors to use $\in \set{1,3,5}$, power parameter for the Minkowski metric $\in \set{1,2}$
        \item XGBoost (\textit{\xg{}}): maximum depth $\in \set{3,5,10}$
    \end{itemize}
    
\paragraph{Regression methods}
    \begin{itemize}
        \item \lift{}/\gptj{}: number of epochs $\in \set{2,6,10}$
        \item \lift{}/\gptt{}: learning rate multiplier $\in \set{0.05, 0.1, 0.2}$
        \item Polynomial Regression (\textit{\qr{}}): no hyperparameter selection but fixed the degree at $3$ since higher-order polynomial regression introduces out-of-memory error, especially for high-dimensional datasets
        \item K-Nearest Neighbor (\textit{\knn{}}): number of neighbors $\in \set{2,5,8}$
        \item Kernel Regression (\textit{\krr{}}): Gamma parameter of Radial Basis Kernel $\in \set{0.01, 0.1, 1}$
        \item Multilayer Perceptron (\textit{\ann{}}): initial learning rate $\in \set{0.0001, 0.001, 0.01}$
        \item Gradient Boosting Tree (\textit{\gbt{}}): learning rate $\in \set{0.001, 0.01, 0.1}$
        \item Random Forest (\textit{\randomf{}}): maximum depth $\in \set{4,6}$
        \item Gaussian Process (\textit{\gp{}}): the number of optimizer restarts used to find parameters of the kernel that maximize the log marginal likelihood $\in \set{5,10}$.
    \end{itemize}
Note that we perform model selection based on validation RAE, instead of validation loss. 

\section{Detailed and Extended Results of Experiments in the Main Paper}
\label{app:full_findings}

In this section, we provide the extended version of experimental results presented in the main paper, including the primary findings of \lift{} (\ref{app:basic_findings}), the specific learning properties of \lift{} (\ref{app:lift_specific}), and improving techniques for \lift{} (\ref{app:lift_improve}).

\subsection{Results for Basic Findings of \lift{} (Section~\ref{sec:lift_basics})}
\label{app:basic_findings}

\subsubsection{How Well Does \lift{} Perform on Standard ML Tasks?}
\label{app:acc_full}

\begin{table}[!htbp]
    \caption{
    \textbf{Accuracies ($\uparrow$) on various classification datasets.} We evaluate \lift{}/\gpt{}s on different classification datasets: 2D synthetic datasets, tabular datasets in OpenML~\openml{}, and image datasets, varying number of features ($p$) and number of classes ($c$).  
    Overall, \lift{}/\gpt{}s perform comparably well across all tasks.
    \lift{}/\gpt{}s can be adapted well to non-linear datasets (circles, two circles), beyond the capacity of logistic regression.
    On the OpenML data, \lift{}/\gpt{}s achieves competitive performances with the best methods, such as XGBoost or RBF-SVM. 
    The performance of \lift{}/\gpt{}s degrades as the number of classes is large, \textit{e.g.}, when the number of classes $c$=100.
    \tuan{On the vision data, \lift{}/\gpt{}s perform comparably well, achieving highly competitive accuracies on both MNIST and Fashion MNIST.
    We note that the classes of MNIST are not fully balanced. Thus MCC gets 11.35\% instead of 10\% as MCC returns the optimal class learned from the training dataset. }
    }
\centering
    \begin{adjustbox}{width=\textwidth,center}
    \begin{tabular}{ccccccccccc|cc}
    \toprule[0.07cm]
    \textbf{Type} &    \textbf{Dataset (ID)} & 
\textbf{$p/c$} & \textbf{MCC} & \textbf{LogReg} & \textbf{KNN}   & \textbf{DT}  & \textbf{MLP}    & \textbf{RBF-SVM}   & \textbf{RF}    & \textbf{XG} & \textbf{\lift{}/GPT-J}  & \textbf{\lift{}/GPT-3} \\[0.05cm] \midrule 
    \multirow{5}[2]{*}{Synthetic} & circles (3) & 2 / 2     & 50.00    & 48.58\relscale{0.8}{$\pm$1.94} & 81.25\relscale{0.8}{$\pm$0.20} & 77.42\relscale{0.8}{$\pm$0.24} & {82.00\relscale{0.8}{$\pm$0.54}} & \textbf{83.08\relscale{0.8}{$\pm$0.59}} & {82.42\relscale{0.8}{$\pm$1.33}} & 81.42\relscale{0.8}{$\pm$0.31} & 79.95\relscale{0.8}{$\pm$1.53} & 81.17\relscale{0.8}{$\pm$0.42} \bigstrut[t]\\
          & two circles (6) & 2 / 2     & 50.00    & 49.83\relscale{0.8}{$\pm$4.18} & \textbf{81.83\relscale{0.8}{$\pm$0.62}} & 75.50\relscale{0.8}{$\pm$0.20} & 68.42\relscale{0.8}{$\pm$3.86} & {80.00\relscale{0.8}{$\pm$0.54}} & 76.08\relscale{0.8}{$\pm$0.59} & 79.25\relscale{0.8}{$\pm$0.35} & 75.92\relscale{0.8}{$\pm$1.65} & 81.42\relscale{0.8}{$\pm$0.82} \\
          & blobs (2) & 2 / 4     & 25.00    & {96.75\relscale{0.8}{$\pm$0.00}} & 95.50\relscale{0.8}{$\pm$0.20} & 96.08\relscale{0.8}{$\pm$0.82} & 96.58\relscale{0.8}{$\pm$0.42} & {96.75\relscale{0.8}{$\pm$0.00}} & \textbf{97.17\relscale{0.8}{$\pm$0.24}} & 96.17\relscale{0.8}{$\pm$0.12} & 96.17\relscale{0.8}{$\pm$0.59} & 96.67\relscale{0.8}{$\pm$0.24} \\
          & moons (4) & 2 / 4     & 50.00    & 88.58\relscale{0.8}{$\pm$0.12} & \textbf{100.00\relscale{0.8}{$\pm$0.00}} & 99.25\relscale{0.8}{$\pm$0.41} & 98.75\relscale{0.8}{$\pm$1.08} & \textbf{100.00\relscale{0.8}{$\pm$0.00}} & 99.75\relscale{0.8}{$\pm$0.00} & 99.83\relscale{0.8}{$\pm$0.12} & 99.58\relscale{0.8}{$\pm$0.42} & \textbf{100.00\relscale{0.8}{$\pm$0.00}} \\
          & 9Clusters (1) & 2 / 9     & 11.25 & 100.00\relscale{0.8}{$\pm$0.00} & 100.00\relscale{0.8}{$\pm$0.00} & 100.00\relscale{0.8}{$\pm$0.00} & 100.00\relscale{0.8}{$\pm$0.00} & 100.00\relscale{0.8}{$\pm$0.00} & \textbf{100.00\relscale{0.8}{$\pm$0.00}} & \textbf{100.00\relscale{0.8}{$\pm$0.00}} & 99.75\relscale{0.8}{$\pm$0.00} & \textbf{100.00\relscale{0.8}{$\pm$0.00}} \bigstrut[b]\\
    \midrule
   \multirow{13}[2]{*}{Tabular} 
          & Customers (1511)  & 9 / 2     & 68.18 & {87.12\relscale{0.8}{$\pm$0.54}} & \textbf{88.64\relscale{0.8}{$\pm$0.00}} & 85.98\relscale{0.8}{$\pm$0.53} & 86.36\relscale{0.8}{$\pm$1.86} & {86.36\relscale{0.8}{$\pm$0.00}} & 85.23\relscale{0.8}{$\pm$0.00} & 85.23\relscale{0.8}{$\pm$0.00} & 85.23\relscale{0.8}{$\pm$1.61} & 84.85\relscale{0.8}{$\pm$1.42} \\
          & Pollution (882)   & 16 / 2    & 50.00 & 58.33\relscale{0.8}{$\pm$11.79} & 66.67\relscale{0.8}{$\pm$6.81} & \textbf{77.78\relscale{0.8}{$\pm$3.93}} & {66.67\relscale{0.8}{$\pm$0.00}} & 58.33\relscale{0.8}{$\pm$6.81} & \textbf{77.78\relscale{0.8}{$\pm$3.93}} & 63.89\relscale{0.8}{$\pm$7.86} & 63.89\relscale{0.8}{$\pm$3.93} & 63.89\relscale{0.8}{$\pm$7.86} \\
          & Spambase (44)   & 58 / 2    & 60.59 & 93.27\relscale{0.8}{$\pm$0.00} & 90.77\relscale{0.8}{$\pm$0.00} & 90.7\relscale{0.8}{$\pm$0.14} & \textbf{94.35\relscale{0.8}{$\pm$0.00}} & 93.70\relscale{0.8}{$\pm$0.00} & {95.01\relscale{0.8}{$\pm$0.00}} & \textbf{95.87\relscale{0.8}{$\pm$0.00}} & 94.03\relscale{0.8}{$\pm$0.54}  & 94.90\relscale{0.8}{$\pm$0.36} \\
          & Hill-Valley (1479)  & 101 / 2   & 49.79 & {77.78\relscale{0.8}{$\pm$0.00}} & 56.38\relscale{0.8}{$\pm$0.00} & 56.38\relscale{0.8}{$\pm$0.89} & 50.21\relscale{0.8}{$\pm$0.00} & 68.72\relscale{0.8}{$\pm$0.00} & 51.44\relscale{0.8}{$\pm$0.00} & 59.26\relscale{0.8}{$\pm$0.00} & \textbf{100.00\relscale{0.8}{$\pm$0.20}} & \textbf{99.73\relscale{0.8}{$\pm$0.19}} \\
          & TAE (48) & 6 / 3     & 35.48 & 45.16\relscale{0.8}{$\pm$4.56} & 60.22\relscale{0.8}{$\pm$4.02} & {65.59\relscale{0.8}{$\pm$5.49}} & 54.84\relscale{0.8}{$\pm$2.63} & 53.76\relscale{0.8}{$\pm$6.63} & \textbf{67.74\relscale{0.8}{$\pm$7.90}} & {66.67\relscale{0.8}{$\pm$8.05}} & 61.29\relscale{0.8}{$\pm$6.97} & 65.59\relscale{0.8}{$\pm$6.63} \\
          & CMC (23)  & 10 / 3    & 42.71 & 49.49\relscale{0.8}{$\pm$0.83} & 50.85\relscale{0.8}{$\pm$1.91} & {56.72\relscale{0.8}{$\pm$0.32}} & 57.29\relscale{0.8}{$\pm$0.73} & 56.50\relscale{0.8}{$\pm$0.97} & 53.45\relscale{0.8}{$\pm$1.05} & 52.43\relscale{0.8}{$\pm$0.42} & 49.83\relscale{0.8}{$\pm$0.28} & \textbf{57.74\relscale{0.8}{$\pm$0.89}} \\
          & Wine (187) & 14 / 3    & 38.89 & \textbf{100.00\relscale{0.8}{$\pm$0.00}} & 96.29\relscale{0.8}{$\pm$1.31} & 93.52\relscale{0.8}{$\pm$2.62} & 98.15\relscale{0.8}{$\pm$2.62} & \textbf{100.00\relscale{0.8}{$\pm$0.00}} & \textbf{100.00\relscale{0.8}{$\pm$0.00}} & 97.22\relscale{0.8}{$\pm$0.00} & 93.52\relscale{0.8}{$\pm$1.31} & 92.59\relscale{0.8}{$\pm$1.31} \\
            (OpenML)          & Vehicle (54)   & 19 / 4    & 25.88 & {80.39\relscale{0.8}{$\pm$1.00}} & 69.61\relscale{0.8}{$\pm$0.74} & 63.92\relscale{0.8}{$\pm$2.37} & {79.21\relscale{0.8}{$\pm$0.28}} & \textbf{81.18\relscale{0.8}{$\pm$0.48}} & 75.88\relscale{0.8}{$\pm$1.27} & 73.14\relscale{0.8}{$\pm$0.28} &
            64.31\relscale{0.8}{$\pm$2.37} & 70.20\relscale{0.8}{$\pm$2.73} \\
          & LED (40496) & 8 / 10     & 11.00  & {68.67\relscale{0.8}{$\pm$0.94}} & 63.67\relscale{0.8}{$\pm$6.13} & 66.33\relscale{0.8}{$\pm$2.87} & \textbf{72.00\relscale{0.8}{$\pm$0.82}} & 68.00\relscale{0.8}{$\pm$0.82} & 64.33\relscale{0.8}{$\pm$0.94} & 66.00\relscale{0.8}{$\pm$0.82} & 65.33\relscale{0.8}{$\pm$0.47} & {69.33\relscale{0.8}{$\pm$2.05}} \\
          & OPT (28)   & 65 / 10     & 10.14  & 96.53\relscale{0.8}{$\pm$0.22} & 96.92\relscale{0.8}{$\pm$0.16} & 89.8\relscale{0.8}{$\pm$1.09} & 97.36\relscale{0.8}{$\pm$0.27} & {97.95\relscale{0.8}{$\pm$0.00}} & 97.69\relscale{0.8}{$\pm$0.14} & 97.48\relscale{0.8}{$\pm$0.17} & 98.22\relscale{0.8}{$\pm$0.11} & \textbf{98.99\relscale{0.8}{$\pm$0.30}} \\
          & Mfeat (12)  & 217 / 10    & 10.00    & 97.67\relscale{0.8}{$\pm$0.12} & 97.67\relscale{0.8}{$\pm$0.31} & 87.67\relscale{0.8}{$\pm$1.05} & 96.5\relscale{0.8}{$\pm$0.35} & \textbf{98.83\relscale{0.8}{$\pm$0.24}} & {97.75\relscale{0.8}{$\pm$0.35}} & 96.75\relscale{0.8}{$\pm$0.00} & {94.17\relscale{0.8}{$\pm$1.75} } & 93.08\relscale{0.8}{$\pm$0.24} \\
          & Margin (1491) & 65 / 100   & 0.94  & {81.35\relscale{0.8}{$\pm$0.15}} & 77.60\relscale{0.8}{$\pm$0.97} & 43.86\relscale{0.8}{$\pm$1.21} & 77.71\relscale{0.8}{$\pm$1.91} & \textbf{81.98\relscale{0.8}{$\pm$0.30}} & {77.71\relscale{0.8}{$\pm$1.98}} & 70.21\relscale{0.8}{$\pm$0.29} & 50.23\relscale{0.8}{$\pm$1.33} & 59.37\relscale{0.8}{$\pm$0.92} \\
          & Texture (1493) & 65 / 100    & 0.94  & {81.67\relscale{0.8}{$\pm$0.97}} & 80.62\relscale{0.8}{$\pm$0.76} & 46.88\relscale{0.8}{$\pm$1.93} & 76.88\relscale{0.8}{$\pm$2.44} & \textbf{83.44\relscale{0.8}{$\pm$0.89}} & 73.12\relscale{0.8}{$\pm$0.76} & 70.73\relscale{0.8}{$\pm$1.41} & 50.32\relscale{0.8}{$\pm$2.18} & 67.50\relscale{0.8}{$\pm$1.42} \bigstrut[b]\\
    \midrule
    \multirow{3}[2]{*}{Images} & MNIST & \multirow{3}[2]{*}{784 / 10} & 11.35 & 91.95\relscale{0.8}{$\pm$0.69} & 96.71\relscale{0.8}{$\pm$0.11}  & 87.42\relscale{0.8}{$\pm$0.64} & 97.30\relscale{0.8}{$\pm$0.16} &  {97.70\relscale{0.8}{$\pm$0.97}} & 94.91\relscale{0.8}{$\pm$0.18}  & 97.69\relscale{0.8}{$\pm$0.04} & 97.01\relscale{0.8}{$\pm$1.15} & \textbf{98.15\relscale{0.8}{$\pm$0.67} } \bigstrut[t]\\
      &  P-MNIST  &       & 11.35 & 92.58\relscale{0.8}{$\pm$0.04} & 96.74\relscale{0.8}{$\pm$0.08} &  87.87\relscale{0.8}{$\pm$0.69} & 97.39\relscale{0.8}{$\pm$0.14}&  \textbf{98.06\relscale{0.8}{$\pm$0.31}} & 94.59\relscale{0.8}{$\pm$0.18}& {97.62\relscale{0.8}{$\pm$0.09}} & 95.80\relscale{0.8}{$\pm$ 0.07}& 96.25\relscale{0.8}{$\pm$0.35}  \\
       &  FMNIST  &       & 10.00 & 85.59\relscale{0.8}{$\pm$0.09} & 85.59\relscale{0.8}{$\pm$0.03}& 80.52\relscale{0.8}{$\pm$0.40}  & 88.86\relscale{0.8}{$\pm$0.02}& \textbf{90.59\relscale{0.8}{$\pm$0.02}} & 85.25\relscale{0.8}{$\pm$0.13}& 90.19\relscale{0.8}{$\pm$0.04} & 85.10 \relscale{0.8}{$\pm$ 0.19}& {90.18 \relscale{0.8}{$\pm$0.12}}  \\
      &  P-FMNIST  &       & 10.00    & 84.95\relscale{0.8}{$\pm$0.84} & 85.15\relscale{0.8}{$\pm$0.61}& 79.91\relscale{0.8}{$\pm$0.93} &  88.86\relscale{0.8}{$\pm$0.61}& {88.04\relscale{0.8}{$\pm$1.69}} &  84.93\relscale{0.8}{$\pm$0.59}& \textbf{89.93\relscale{0.8}{$\pm$0.14}} & 82.25\relscale{0.8}{$\pm$0.27}  & 88.92\relscale{0.8}{$\pm$0.71}  \bigstrut[b]\\
    \bottomrule[.7mm]
    \end{tabular}
    \end{adjustbox}
    \vspace{1mm}
\label{tab:classification_accuracy_full}
\end{table}
We now provide full results of classification and regression performances with all baselines, including the investigation of the interpolation and extrapolation performance of \lift{} for regression tasks. 

\begin{table}[t]
  \centering
  \caption{\bmt{\textbf{Comparison of accuracies($\uparrow$) between \lift{} and deep neural network models designed for tabular datasets.} 
  We consider two baselines here: TabNet~\citep{arik2021tabnet} and TabTransformer~\citep{huang2020tabtransformer}.
  We observe that \lift{} achieves comparable performance to TabNet and TabTransformer, which is more evidence of the good performance of \lift{}. 
  }
  }
    \begin{tabular}{c|ccccccccc}
 \toprule[0.05cm]
  Dataset (ID)    &  \textbf{MCC}   &  \textbf{\lift{}/GPT-3}      &  \textbf{\lift{}/GPT-J}   & \textbf{TabNet} & \textbf{TabTransformer}   \\ \midrule
 Blobs (2)        &  25.00 &  96.67$\pm$ 0.24 &  96.17$\pm$ 0.59  & 96.75$\pm$ 0.00&50.00$\pm$ 0.00  \\
 Two Circles (6)  &  50.00 &  81.42$\pm$ 0.82 &  75.92$\pm$ 1.65 & 74.25$\pm$ 12.39&49.25$\pm$ 1.29 \\
 Iris (61)        &  33.33 &  97.0$\pm$ 0.00  &  96.67$\pm$ 0.00 &  97.78$\pm$ 1.92  & 72.22$\pm$ 5.09       \\        
 Customers (1511) &  68.18 &  84.85$\pm$ 1.42 &  85.23$\pm$ 1.61  & 85.22$\pm$ 3.93  & 87.12$\pm$ 0.66      \\
 Wine (187)       &  38.89 &  92.59$\pm$ 1.31 &  93.52$\pm$ 1.31  & 94.44$\pm$ 5.56  & 90.74$\pm$ 13.70 \\
 LED (40496)      &  11.0  &  69.33$\pm$ 2.05 &  65.33$\pm$ 0.47& 67.00$\pm$ 2.46  & 41.00$\pm$ 12.49  \\
    \bottomrule[0.05cm]
    \end{tabular}
\vspace{-1mm}
\label{tab:tabtransoformer}
\end{table}

\paragraph{Classification. }
Table~\ref{tab:classification_accuracy_full} presents the classification performance with other baselines, including KNN, MLP, and Random Forest.
We further consider two additional baselines here, which has larger model sizes compared to the baselines we discussed in Sec.~\ref{sec:methodology}.
TabNet~\citep{arik2021tabnet} and TabTransformer~\citep{huang2020tabtransformer} are deep neural network models based on architectures specifically designed for tabular data. 
The results are presented in Table~\ref{tab:tabtransoformer}. 
We observe that \lift{} achieves comparable performance to TabNet and TabTransformer.
This further highlight the good performance of \lift{}. 

\paragraph{Regression. } Table~\ref{table:reg_all} provides the regression evaluation with all regression baselines on synthetic datasets, and Table~\ref{table:reg_real} provides the results for real datasets. 
Since experiments with \lift{}/\gptj{} are conduced on AWS and local server and due to this limitation of memory resources, we fail to run experiments of \lift{}/\gptj{} on high-dimensional datasets. 
Therefore, for 50D and 120D synthetic datasets, only results of \lift{}/\gptt{} are reported.

\begin{figure}
    \centering
    \includegraphics[width=\textwidth]{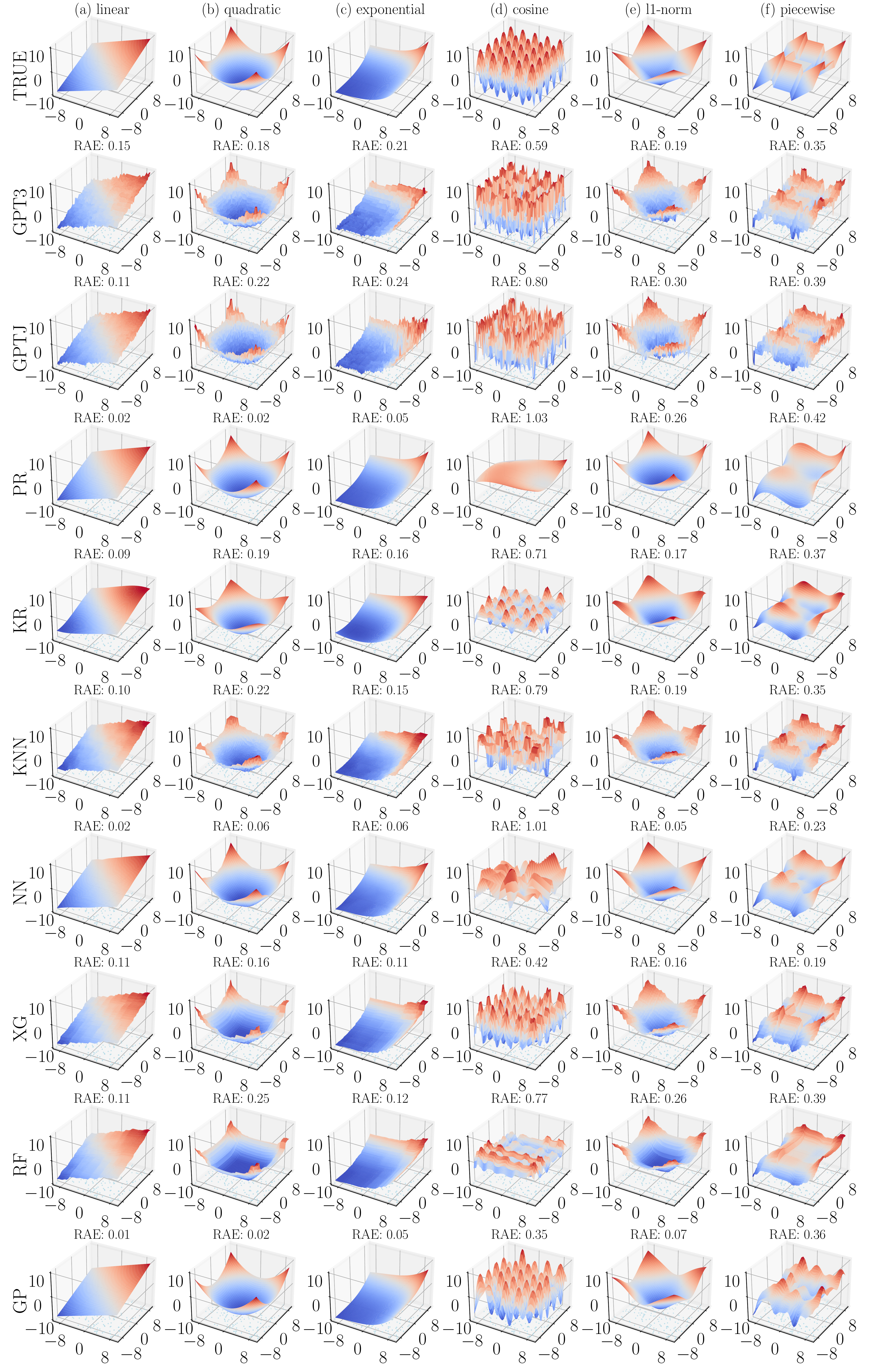}
    \caption{\textbf{Performance of \lift{}/\gpt{}s and baselines in approximating various functions. }
    The first row visualizes the true values of the functions,
    and the second \& third rows visualize the predicted values of \lift{}/\gpt{}s after fine-tuning for the corresponding regression tasks with \textbf{200 training samples}. We compared with other baselines with the same training samples.
    }
    \label{fig:reg_200}
\end{figure}

\begin{figure}
    \centering
    \includegraphics[width=\textwidth]{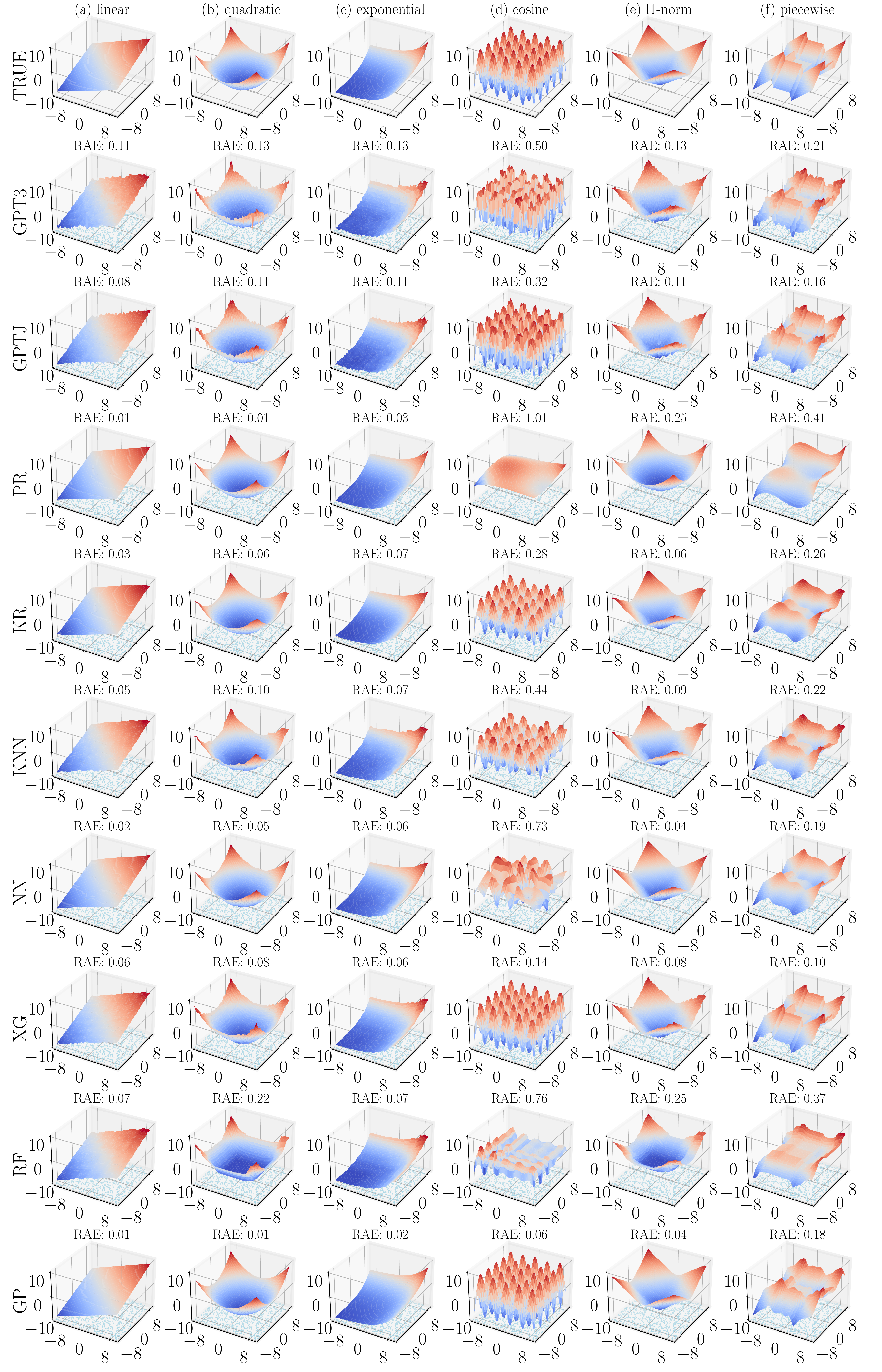}
    \caption{\textbf{Performance of \lift{}/\gpt{}s in approximating various functions. }
    The first row visualizes the true values of the functions,
    and the second \& third rows visualize the predicted values of \lift{}/\gpt{}s after fine-tuning for the corresponding regression tasks with \textbf{1000 training samples}. We compared with other baselines with the same training samples.
    }
    
    \label{fig:reg_1000}
\end{figure}

We further provide the visualization of regression models. Fig.~\ref{fig:reg_200} and ~\ref{fig:reg_1000} visualize the 2D predictions for various functions with 200 and 1000 samples training datasets, respectively.
Each coordinate of the training sample is drawn uniformly from $[-10,10]$.
Specifically, the prediction is performed on the interval $[-12,12]$.

\begin{table}[!htbp]
\caption{ \textbf{Comparison of regression methods in approximating various functions.}
The regression performance is measured by RAE($\downarrow$), and we tested on six functions with various $p$, the number of features. 
\lift{} can approximate different types of functions in low-dimensional cases ($p=1, 2$), although it fails to achieve performance comparable to that of strong baselines.
We observed that \lift{} fails to achieve satisfying regression performance in high-dimensional cases ($p=50, 100$). 
\tuan{
\textit{Results of \lift{}/\gptj{} on high-dimensional datasets are not available due to the resource limitation.}
}
}
\centering
\resizebox{\textwidth}{!}{
\begin{tabular}{llccccccccc}
\toprule[.5mm]
\multicolumn{2}{l}{\diagbox{\textbf{Dataset}}{\textbf{Method}}} &  \textbf{\qr{}} & \textbf{\krr{}} & \textbf{\knn{}}  & \textbf{\ann{}} & \textbf{\gbt{}} & \textbf{\randomf{}} & \textbf{\gp{}} & \textbf{\lift{}/\gptj{}} & \textbf{\lift{}/\gptt{}}\\ \midrule
    \multirow{4}{*}{\linear{}} & $p=1$ & $0.01 \pm 0.0$ & $0.05 \pm 0.0$ & $0.04 \pm 0.0$ & $0.03 \pm 0.0$ & $0.05 \pm 0.0$ & $0.04 \pm 0.0$ & $0.01 \pm 0.0$ & $0.08 \pm 0.0$ & $0.06 \pm 0.0$ \\ 
    & $p = 2$ & $0.03 \pm 0.0$ & $0.09 \pm 0.0$ & $0.12 \pm 0.0$ & $0.04 \pm 0.0$ & $0.12 \pm 0.0$ & $0.12 \pm 0.0$ & $0.01 \pm 0.0$ & $0.12 \pm 0.0$  & $0.19 \pm 0.0$ \\ 
    & $p = 50$ & $0.71 \pm 0.0$ & $1.02 \pm 0.0$ & $0.78 \pm 0.0$ & $1.85 \pm 0.1$ & $0.97 \pm 0.0$ & $0.87 \pm 0.0$ & $0.13 \pm 0.0$ & - & $1.18 \pm 0.2$ \\ 
    & $p = 100$ & $0.95 \pm 0.0$ & $1.02 \pm 0.0$ & $0.88 \pm 0.0$ & $3.02 \pm 0.0$ & $0.99 \pm 0.0$ & $0.94 \pm 0.0$ & $0.64 \pm 0.0$& - & $2.14 \pm 0.5$ \\ \midrule
    
    \multirow{4}{*}{\quadr{}}& $p=1$& $0.01 \pm 0.0$ & $0.05 \pm 0.0$ & $0.05 \pm 0.0$ & $0.03 \pm 0.0$ & $0.06 \pm 0.0$ & $0.05 \pm 0.0$ & $0.01 \pm 0.0$ & $0.11 \pm 0.0$ & $0.13 \pm 0.0$ \\ 
     & $p = 2$ &$0.03 \pm 0.0$ & $0.16 \pm 0.0$ & $0.17 \pm 0.0$ & $0.06 \pm 0.0$ & $0.15 \pm 0.0$ & $0.25 \pm 0.0$ & $0.02 \pm 0.0$ & $0.28 \pm 0.1$  & $0.22 \pm 0.0$ \\ 
     & $p = 50$ & $1.12 \pm 0.0$ & $5.19 \pm 0.0$ & $1.33 \pm 0.0$ & $2.28 \pm 0.0$ & $0.98 \pm 0.0$ & $0.96 \pm 0.0$ & $0.69 \pm 0.0$ & - & $0.99 \pm 0.2$ \\ 
     & $p = 100$ &$1.02 \pm 0.0$ & $7.30 \pm 0.0$ & $1.29 \pm 0.0$ & $2.89 \pm 0.0$ & $1.01 \pm 0.0$ & $0.98 \pm 0.0$ & $0.89 \pm 0.0$& - & $1.06 \pm 0.1$ \\ \midrule
     
\multirow{4}{*}{\expo{}}& $p=1$& $0.04 \pm 0.0$ & $0.07 \pm 0.0$ & $0.05 \pm 0.0$ & $0.02 \pm 0.0$ & $0.05 \pm 0.0$ & $0.04 \pm 0.0$ & $0.01 \pm 0.0$ & $0.11 \pm 0.0$ & $0.09 \pm 0.0$ \\ 
 & $p = 2$ &$0.04 \pm 0.0$ & $0.15 \pm 0.0$ & $0.13 \pm 0.0$ & $0.07 \pm 0.0$ & $0.09 \pm 0.0$ & $0.11 \pm 0.0$ & $0.04 \pm 0.0$ & $0.19 \pm 0.0$ & $0.20 \pm 0.0$ \\ 
 & $p = 50$ & $0.94 \pm 0.0$ & $10.23 \pm 0.0$ & $1.04 \pm 0.0$ & $3.18 \pm 0.2$ & $1.05 \pm 0.0$ & $0.96 \pm 0.0$ & $0.53 \pm 0.0$ & - & $1.15 \pm 0.0$ \\ 
 & $p = 100$ &  $0.96 \pm 0.0$ & $14.12 \pm 0.0$ & $1.03 \pm 0.0$ & $4.14 \pm 0.0$ & $0.97 \pm 0.0$ & $0.93 \pm 0.0$ & $0.79 \pm 0.0$& - & $1.03 \pm 0.0$ \\ \midrule
 
\multirow{4}{*}{\cosi{}}& $p=1$& $1.05 \pm 0.0$ & $0.12 \pm 0.0$ & $0.14 \pm 0.0$ & $0.38 \pm 0.1$ & $0.15 \pm 0.0$ & $0.35 \pm 0.0$ & $0.04 \pm 0.0$ & $0.38 \pm 0.1$ & $0.44 \pm 0.1$ \\ 
 & $p = 2$ & $1.04 \pm 0.0$ & $0.74 \pm 0.0$ & $0.83 \pm 0.1$ & $1.06 \pm 0.0$ & $0.41 \pm 0.0$ & $0.80 \pm 0.0$ & $0.31 \pm 0.0$ & $0.82 \pm 0.2$ &  $0.65 \pm 0.1$ \\ 
 & $p = 50$ & $1.01 \pm 0.0$ & $1.01 \pm 0.0$ & $1.00 \pm 0.0$ & $1.59 \pm 0.0$ & $1.00 \pm 0.0$ & $0.99 \pm 0.0$ & $1.01 \pm 0.0$ & - & $1.25 \pm 0.1$ \\ 
 & $p = 100$ & $1.02 \pm 0.0$ & $1.00 \pm 0.0$ & $1.09 \pm 0.0$ & $2.43 \pm 0.1$ & $1.04 \pm 0.0$ & $1.06 \pm 0.0$ & $1.00 \pm 0.0$ & -& $1.20 \pm 0.3$ \\ \midrule
 
\multirow{4}{*}{\lone{}}& $p=1$& $0.23 \pm 0.0$ & $0.06 \pm 0.0$ & $0.05 \pm 0.0$ & $0.03 \pm 0.0$ & $0.06 \pm 0.0$ & $0.06 \pm 0.0$ & $0.03 \pm 0.0$ & $0.10 \pm 0.0$ & $0.09 \pm 0.0$ \\ 
 & $p = 2$ & $0.24 \pm 0.0$ & $0.17 \pm 0.0$ & $0.19 \pm 0.0$ & $0.06 \pm 0.0$ & $0.15 \pm 0.0$ & $0.29 \pm 0.0$ & $0.07 \pm 0.0$ & $0.24 \pm 0.0$  & $0.20 \pm 0.0$ \\ 
 & $p = 50$ & $1.09 \pm 0.0$ & $1.00 \pm 0.0$ & $1.28 \pm 0.0$ & $1.97 \pm 0.1$ & $0.98 \pm 0.0$ & $0.94 \pm 0.0$ & $0.96 \pm 0.0$ & - & $1.12 \pm 0.1$ \\ 
 & $p = 100$ & $1.01 \pm 0.0$ & $1.01 \pm 0.0$ & $1.22 \pm 0.0$ & $2.80 \pm 0.1$ & $1.03 \pm 0.0$ & $1.01 \pm 0.0$ & $0.99 \pm 0.0$& - & $1.27 \pm 0.2$ \\ \midrule
 
\multirow{4}{*}{\pw{}}& $p=1$& $0.45 \pm 0.0$ & $0.17 \pm 0.0$ & $0.08 \pm 0.0$ & $0.08 \pm 0.0$ & $0.06 \pm 0.0$ & $0.07 \pm 0.0$ & $0.10 \pm 0.0$ & $0.15 \pm 0.0$ & $0.17 \pm 0.0$ \\ 
 & $p = 2$ & $0.39 \pm 0.0$ & $0.34 \pm 0.0$ & $0.33 \pm 0.0$ & $0.20 \pm 0.0$ & $0.19 \pm 0.0$ & $0.38 \pm 0.0$ & $0.29 \pm 0.0$ & $0.40 \pm 0.1$  &$0.40 \pm 0.1$ \\ 
 & $p = 50$ & $0.93 \pm 0.0$ & $1.00 \pm 0.0$ & $0.97 \pm 0.0$ & $2.11 \pm 0.0$ & $1.00 \pm 0.0$ & $0.94 \pm 0.0$ & $0.93 \pm 0.0$ & - & $1.35 \pm 0.1$ \\ 
 & $p = 100$ & $1.01 \pm 0.0$ & $1.00 \pm 0.0$ & $1.08 \pm 0.0$ & $4.20 \pm 0.1$ & $1.02 \pm 0.0$ & $1.01 \pm 0.0$ & $1.01 \pm 0.0$& - & $1.11 \pm 0.0$ \\ 
 
\bottomrule
\end{tabular} }
\vspace{1in}
\label{table:reg_all}
\end{table}

\begin{table}[!htbp]
\caption{
\textbf{Comparison of regression methods in real datasets.} 
The regression performance is measured by RAE($\downarrow$). 
We observe that \lift{}/\gptt{} achieves the top 2 regression performance among all the real datasets. 
}
\centering
\begin{adjustbox}{width=\textwidth,center}
\begin{tabular}{l|ccccccc|cc}
\toprule[.5mm]
\diagbox{\textbf{Dataset}}{\textbf{Method}} &  \textbf{\qr{}} & \textbf{\krr{}} & \textbf{\knn{}}  & \textbf{\ann{}} & \textbf{\gbt{}} & \textbf{\randomf{}} & \textbf{\gp{}} & \textbf{\lift{}/\gptj{}} & \textbf{\lift{}/\gptt{}}\\ \midrule
ccpp & $0.22 \pm 0.00$ &$21.60 \pm 0.00$ &$0.45 \pm 0.00$ &$0.30 \pm 0.00$ & \textbf{$0.17 \pm 0.00$} & \textbf{$0.21 \pm 0.00$} &$0.69 \pm 0.00$& $0.24 \pm 0.01$ & \textbf{$0.18 \pm 0.01$} \\ \midrule
servo& $0.92 \pm 0.00$ &$0.95 \pm 0.00$ &$0.86 \pm 0.00$ &$0.82 \pm 0.00$ &$0.25 \pm 0.00$ &$0.25 \pm 0.00$ &$1.03 \pm 0.00$& $1.17 \pm 0.16$ &  $0.29 \pm 0.02$   \\ \midrule
insurance& $0.48 \pm 0.00$ &$1.48 \pm 0.00$ &$1.03 \pm 0.00$ &$0.44 \pm 0.00$ &$0.25 \pm 0.00$ &$0.26 \pm 0.00$ &$1.30 \pm 0.00$& $0.53 \pm 0.11$ &  $0.14 \pm 0.05$ \\ \midrule
student& $0.47 \pm 0.00$ &$1.56 \pm 0.00$ &$0.66 \pm 0.00$ &$0.37 \pm 0.00$ &$0.39 \pm 0.00$ &$0.36 \pm 0.00$ &$0.45 \pm 0.00$& $0.36 \pm 0.02$  & $0.27 \pm 0.01$  \\ 
\bottomrule[.5mm]
\end{tabular} 
\end{adjustbox}
\label{table:reg_real}
\end{table}

\begin{figure}
    \centering
    \includegraphics[width=\textwidth]{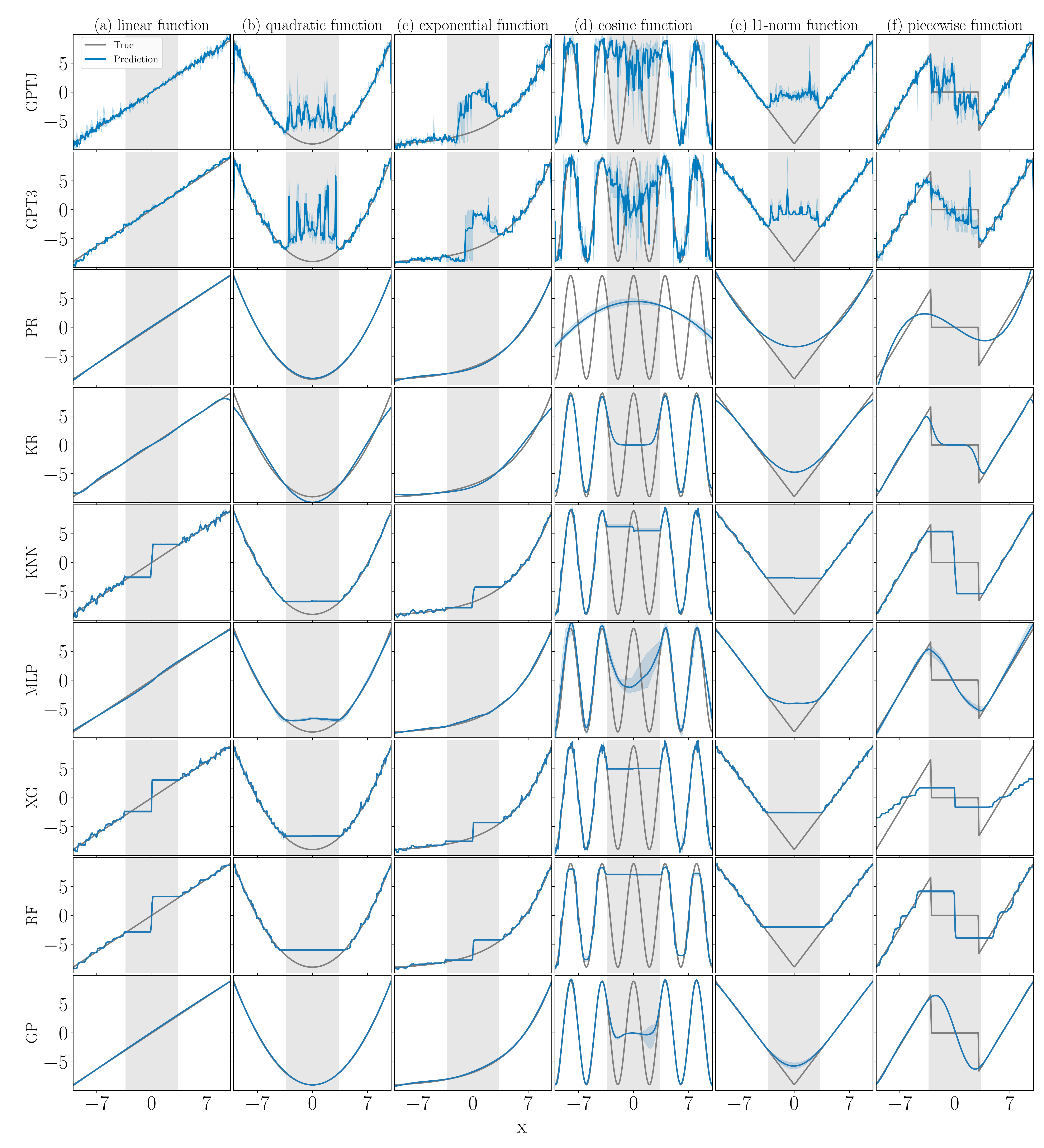}
    \caption{
    \textbf{Interpolation performance on synthetic regression tasks.}
    Each algorithm is trained with samples in the white background region ($3 \leq \lvert x \rvert \leq 10$), and tested on the interpolation area $\lvert x \rvert \leq 3$. \lift/\gpt{}s are having worse interpolation performances compared with existing methods.
    }
    \label{fig:reg_donut_1d}
\end{figure}

\begin{figure}[h]
    \centering
    \includegraphics[width=\textwidth]{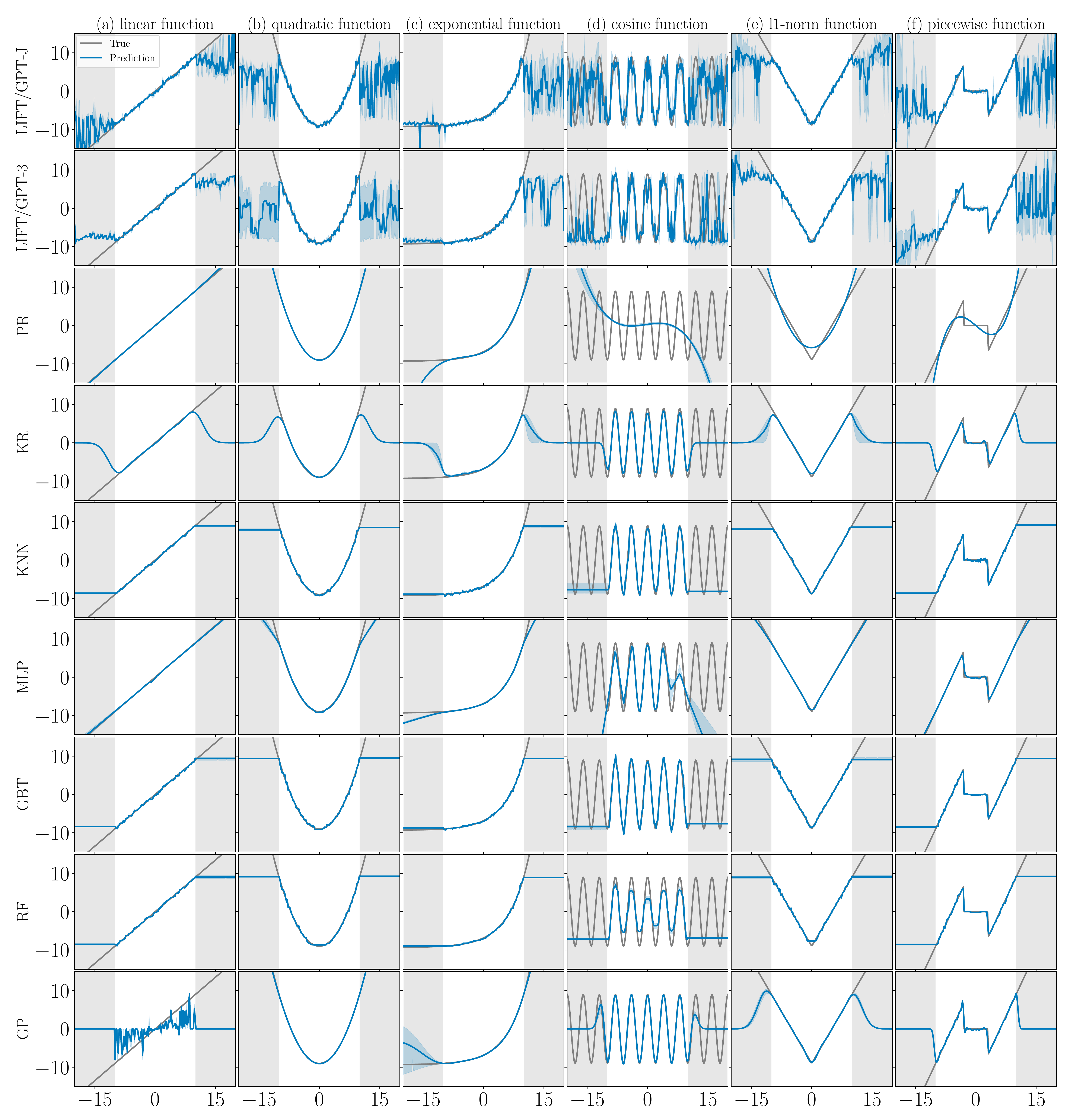}
    \caption{\textbf{Comparison of the extrapolation performance of \lift{} and various baselines on synthetic regression tasks of approximating six functions $f$.}
    Each algorithm is trained by 200 samples $(x,y)$ where the input $x$ is drawn from interval $[-10, 10]$ and the output is defined as $y=f(x)$. We test the how each algorithm perform regression for $x \notin [-10, 10]$.
    }
    \label{fig:reg_ex_1d}
\end{figure}

Fig.~\ref{fig:reg_donut_1d} and Fig.~\ref{fig:reg_ex_1d} visualize the interpolation and extrapolation of various methods. 
All methods fail to extrapolate and interpolate well for all functions. It turns out that \lift{} is not having good interpolation performance except in the linear regression case.  
An interesting observation is that \lift{} tends to output seen values (from training data) for extrapolation. For example, in Fig.~\ref{fig:reg_ex_1d}b, the outputs of \lift{}s for $x \notin [-10, 10]$ (extrapolation) lie in the range of outputs for  $x \in [-10, 10]$ (trained data), and similar behaviors are observed for other functions as well.

\subsubsection{How Many Samples Does \lift{} Need?}
\label{app:sample_complexity}
\begin{figure}[!htbp]
     \centering
     \begin{subfigure}[a]{\textwidth}
         \centering
         \includegraphics[width=0.9\textwidth]{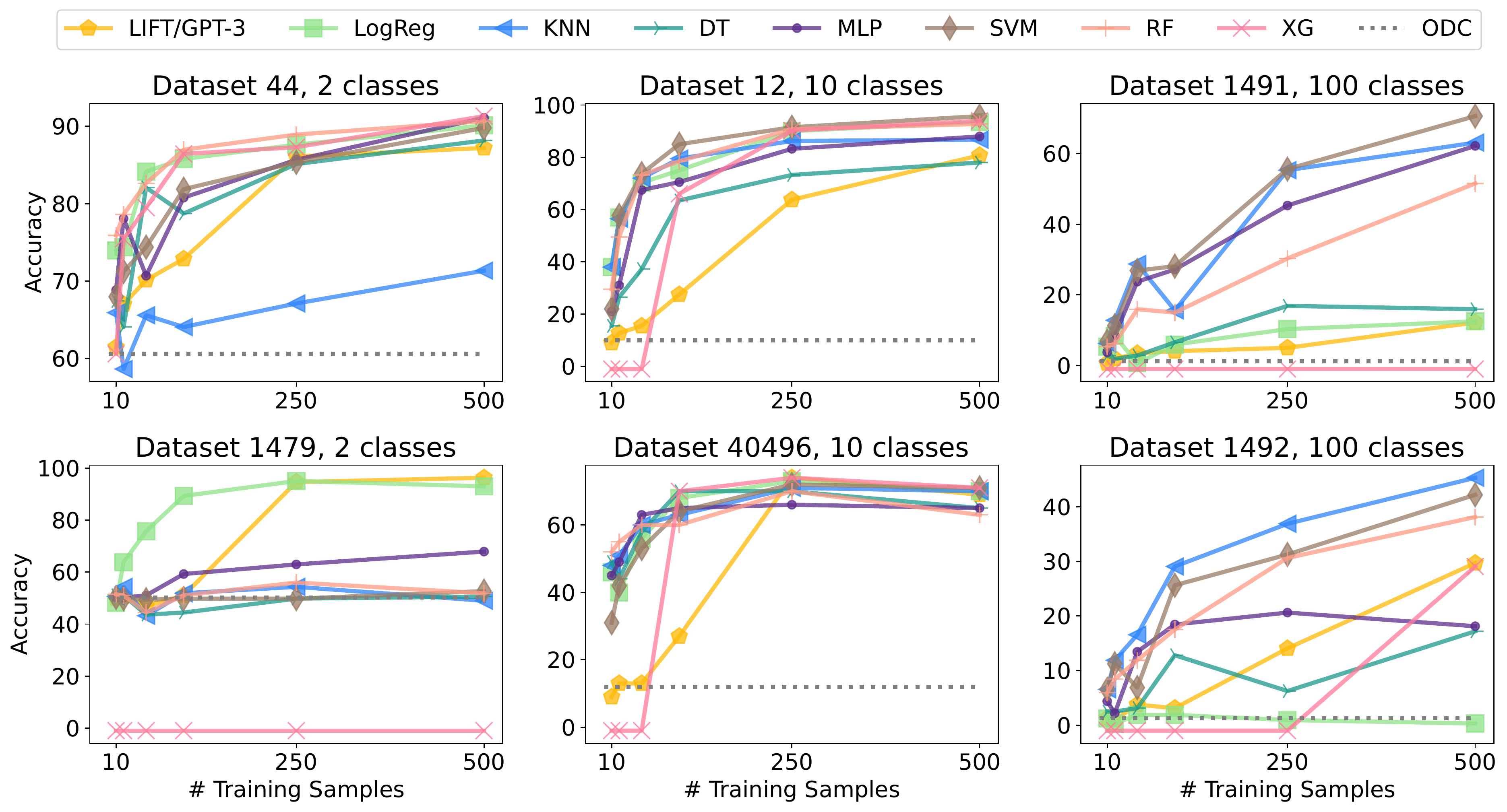}
         \caption{Classification tasks on OpenML tabular datasets. The classification performance is measured in terms of accuracy ($\uparrow$). \textit{Here, ODC denotes optimal deterministic classifier, which is identical to the majority class classifier (MCC) in the main paper.}
         }
         \label{fig:app_sample_complexity_clf}
     \end{subfigure}
     \vfill
    \vspace{1mm}
     \begin{subfigure}[b]{\textwidth}
         \centering
         \includegraphics[width=\textwidth]{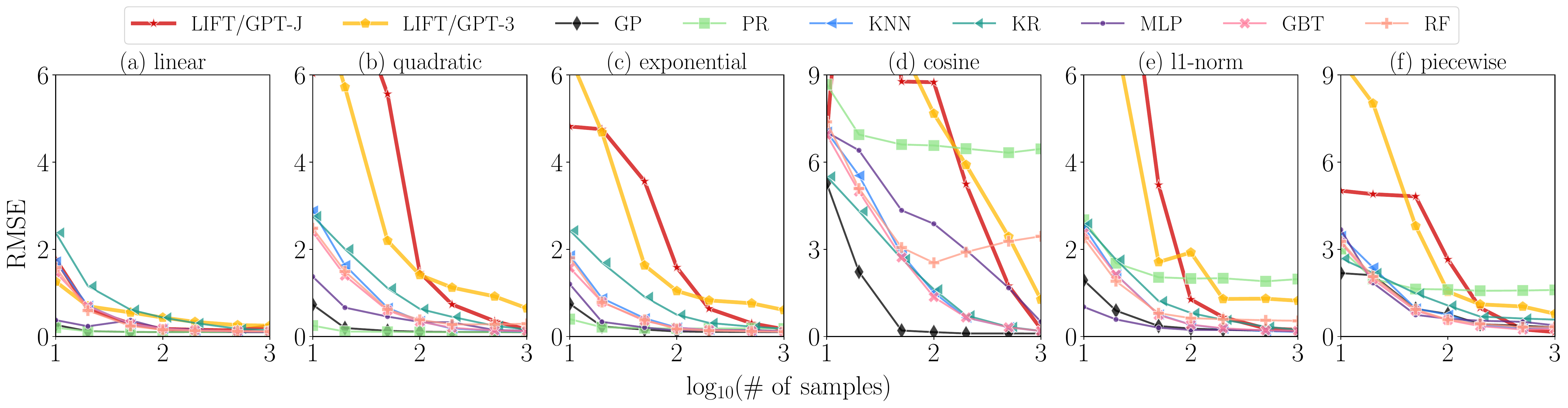}
         \caption{Regression tasks (function approximation). The regression performance is reported in RMSE ($\downarrow$).
         }
         \label{fig:reg:sample_complexity}
     \end{subfigure}
    \caption{
    \textbf{Sample complexity evaluations on classification and regression tasks. }
     Each figure presents the comparison of performance evaluated on \lift{}/\gpt{}s and baselines varying numbers of training samples (10--500 for classification and 10--1000 for regression).
    \lift{} needs a slightly larger sample complexity to start achieving similar performances to the best baseline methods.
    For regression tasks, we note that \lift{} achieves competitive or even better performance when around 1000s of samples are given, especially for the discontinuous functions, \textit{e.g.}, piecewise function. 
    }
    \label{fig:sample_complexity}
\end{figure}

Fig.~\ref{fig:app_sample_complexity_clf} and Fig.~\ref{fig:reg:sample_complexity} provide the 
sample complexity comparisons between evaluated methods in the classification and regression settings.

\subsubsection{Can We Understand the Inductive Biases of Language Models via \lift{}?}\label{app:inductive_bias}

Continuing from Sec.~\ref{sec:inductive_bias}, here we provide more experiments with detailed measurements that quantify the similarity between decision boundaries.

\begin{figure}[htp]
\vspace{-6mm}
    \centering
    \begin{subfigure}[a]{\textwidth}
         \centering
        \includegraphics[width=\textwidth]{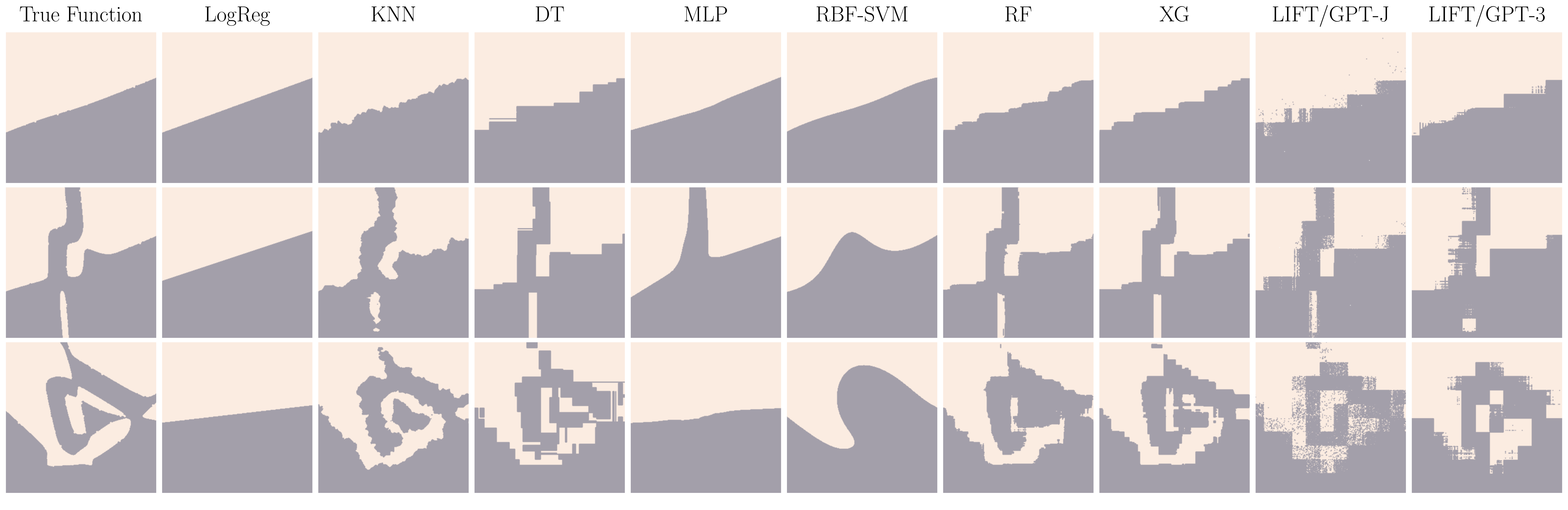}
        \vspace{-6mm}
        \caption{Binary classification}
    \end{subfigure}
    \begin{subfigure}[b]{\textwidth}
         \centering
        \includegraphics[width=\textwidth]{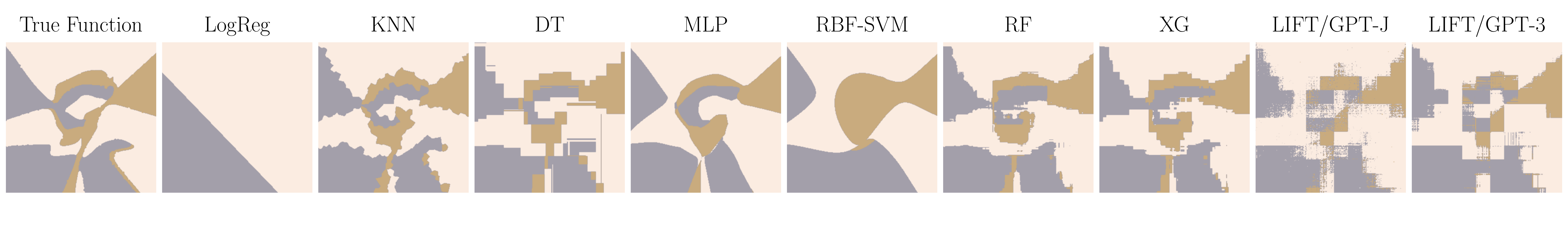}
        \vspace{-8mm}
        \caption{3-way classification} 
    \end{subfigure}
    \begin{subfigure}[b]{\textwidth}
         \centering
        \includegraphics[width=\textwidth]{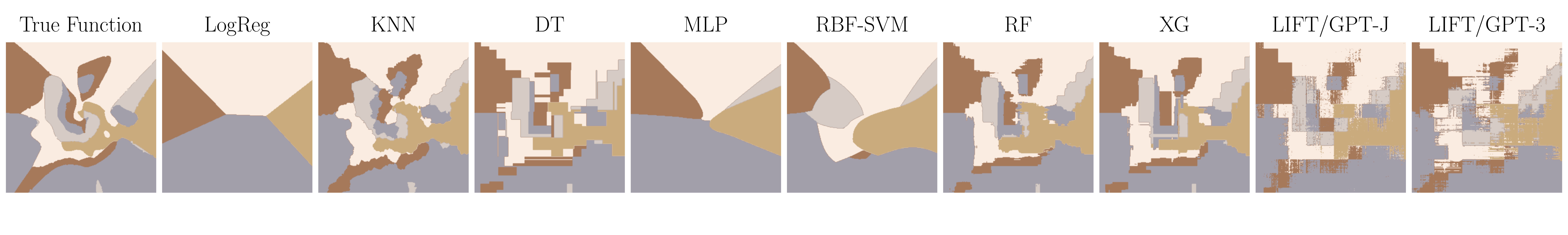}
        \vspace{-8mm}
        \caption{5-way classification} 
    \end{subfigure}
    \begin{subfigure}[a]{\textwidth}
         \centering
        \includegraphics[width=\textwidth]{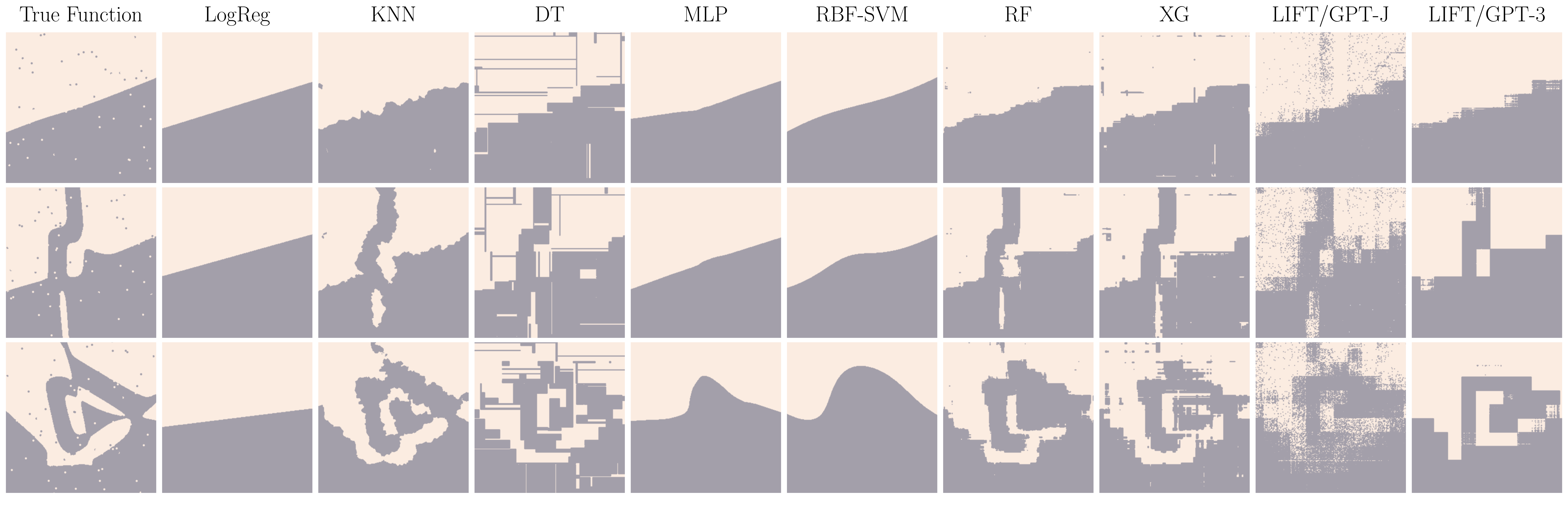}
        \vspace{-6mm}
        \caption{Binary classification with label corruption (corruption ratio is 5\%)}
    \end{subfigure}
    \begin{subfigure}[b]{\textwidth}
         \centering
        \includegraphics[width=\textwidth]{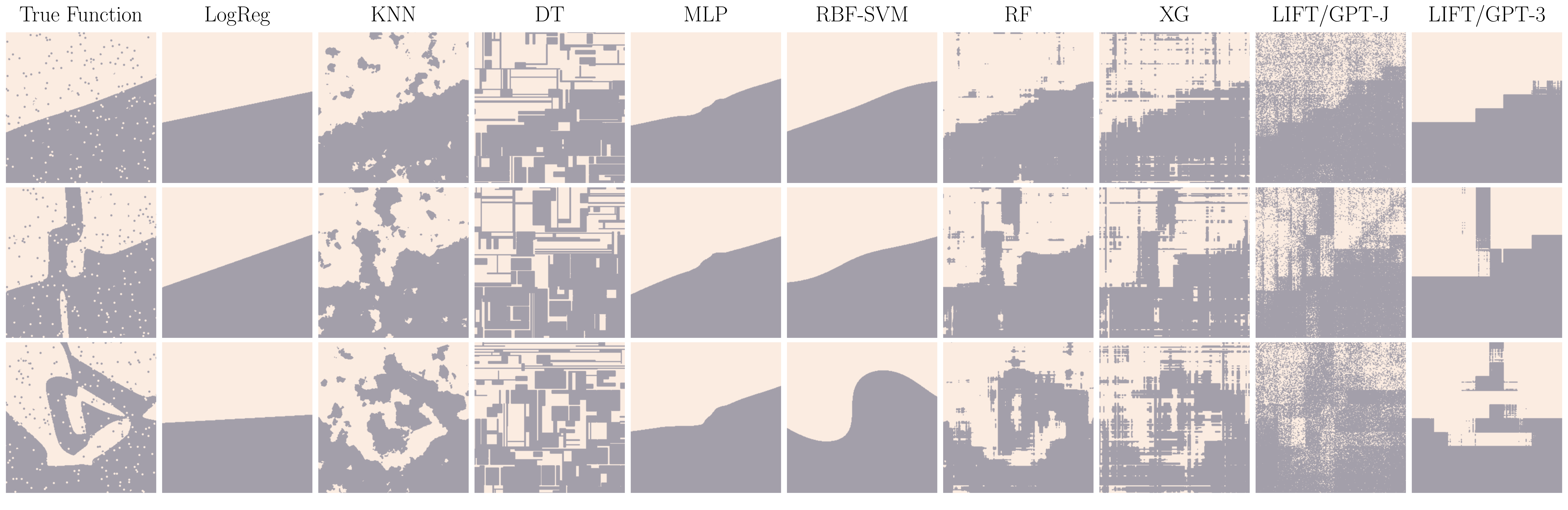}
        \vspace{-6mm}
        \caption{Binary classification with label corruption (corruption ratio is 20\%)}
    \end{subfigure}
    \caption{
    \textbf{Classification decision boundary visualizations on neural-net-based synthetic datasets.} 
    The first column of each row shows the decision boundary of the training datasets.
    In (a), (b), and (c), we visualize the decision boundaries of models trained on datasets with two, three, and five classes. 
    In (d) and (e), we consider the label-corrupted version of the binary-class datasets, with corruption probabilities of 5\% and 20\%.
    We find that \lift{}/\gpt{}s adapt well and roughly estimate the true decision boundaries.
    The shapes of \lift{}'s decision boundary are likely to be axis-parallel and show multiple fractals.}
    \label{fig:clf_decision_boundary}
\end{figure}

\paragraph{Visualizing decision boundary.}
We construct datasets with various classification complexities to investigate the adaptability of \lift{}. 
In particular, we construct three datasets: a binary classification dataset, a 3-class and a 5-class  dataset (shown in the first column of Fig.~\ref{fig:clf_decision_boundary}a, Fig.~\ref{fig:clf_decision_boundary}b, and Fig.~\ref{fig:clf_decision_boundary}c).
We call these datasets \textit{neural-net-based synthetic datasets} since we generate them using a 2-layer neural network. See Fig.~\ref{fig:nn_syn} and Appendix~\ref{app:setup_datatset} for detailed explanations of how we generated these datasets. 
Note that Fig.~\ref{fig:clf_decision_boundary}a is the same as Fig.~\ref{fig:clf_decision_boundary_reduced} in Sec.~\ref{sec:inductive_bias}, which we put for completeness here.

Fig.~\ref{fig:clf_decision_boundary} visualizes the decision boundaries of models trained on the neural-net-based synthetic data.
In addition, we also visualize the decision boundaries of models trained on the \textit{label-corrupted} versions of three binary classification datasets, with the corruption probabilities being 5\% and 20\% (see details in Sec.~\ref{sec:label_corruption}), shown in  Fig.~\ref{fig:clf_decision_boundary}d and  Fig.~\ref{fig:clf_decision_boundary}e.
Specifically, we consider the binary classification tasks and flip the training data labels with the provided probabilities.
Overall, the same observation of Sec.~\ref{sec:inductive_bias} also holds for the 3-class and 5-class datasets — both \gptj{} and \gptt{} models fine-tuned with \lift{} can adapt well to different boundaries. 
They can capture the rough shapes of the decision boundaries in all three settings.

Besides, when the level of corruption increases, decision trees and XGboost are the most affected baselines.
While roughly capturing the boundary, \lift{}/\gptj{} also shows more noisy predictions.
In contrast, \lift{}/\gptt{} displays great robustness against the corrupted labels while capturing the correct boundary shapes.
Nevertheless, this experiment indicates the different behaviors of \lift{}s from the baseline algorithms and their adaptability to different types of decision boundary.

One interesting observation here is that \lift{}'s decision boundaries are axis-parallel and show a lot of fractals.
The axis-parallel boundary looks similar to the boundary of tree-based classifiers, and the fractal shapes of \lift{}'s boundaries are similar to the observations on the decision boundaries of some convolution neural networks~\citep{somepalli2022can}.
However, the main reason why \lift{}'s decision boundary has such patterns seems to be due to the way it interprets numbers. 
Since we rely solely on the language interface, there are some artifacts due to the decimal numeral system.
For instance, $0.98$ and $0.99$ are only one-character different, but $0.99$ and $1.00$ are three-characters different.
We believe that such an artifact is the reason behind axis-parallel decision boundaries and fractal-like patterns.

\begin{table}[t]
\caption{
\textbf{Quantifying the similarity between decision boundaries of \lift{}/\gptt{} and those of various baselines. }
We use different settings of the baselines, where their hyperparameters are given with the baseline name, and the selected values of hyperparameters are specified in the second line.
Each column reports the matching accuracy ($\uparrow$) between the predictions of \lift{}/\gptt{} with those of the baseline.
Each score is a percentage similarity of both \lift{}/\gptt{} and the baseline classifying a point with the same class. 
For example, a score of $100$ for model A signifies that \lift{}/\gptt{} classified all sampled test points in the same manner as model A, regardless of their true dataset accuracy.
The last row reports the average matching accuracy.
We highlight the highly matched algorithms, namely RBF-SVM, MLP (W=100), and Random Forest (E=100).
}
\begin{center}
\setlength\tabcolsep{2pt}
\resizebox{\textwidth}{!}{
\begin{tabular}{l|ccc|c|ccc|cc|ccc|c|ccc}
\toprule
\multirow{2}{*}{\diagbox{Dataset (ID)}{Similarity}{Method}} & \multicolumn{3}{c|}{\textbf{SVM} (kernel)} & \multicolumn{1}{c|}{\textbf{LogReg} 
} & \multicolumn{3}{c|}{\textbf{KNN} ($k$)} & \multicolumn{2}{c|}{\textbf{DT} (depth $D$)} & \multicolumn{3}{c|}{\textbf{\ann{} (width $W$)}} & \multicolumn{1}{c|}{\textbf{XG}} & \multicolumn{3}{c}{\textbf{RF} (\# estimators $E$)} \\ [.15in]
& \textbf{poly} & \textbf{rbf} & \textbf{sigmoid} & 
& \textbf{K=1} & \textbf{K=3} & \textbf{K=5}& \textbf{D=3} & \textbf{D=5} & \textbf{W=10} & \textbf{W=100} & \textbf{W=200} & & \textbf{E=20} & \textbf{E=50} & \textbf{E=100} \\ \midrule
9clusters (1) & 100.00 & 100.00 & 100.00 & 100.00 & 100.00 & 100.00 & 100.00 & 76.00 & 100.00 & 100.00 & 100.00 & 100.00 & 97.50 & 100.00 & 100.00 & 100.00 \\
blobs (2) & 98.50 & 97.50 & 92.00 & 97.50 & 94.00 & 95.50 & 96.00 & 94.50 & 91.00 & 97.00 & 97.00 & 97.00 & 93.50 & 94.50 & 94.00 & 94.00 \\
circles (3) & 54.00 & 93.50 & 48.00 & 51.00 & 85.50 & 88.50 & 88.50 & 67.00 & 80.00 & 84.00 & 92.50 & 92.50 & 87.50 & 85.00 & 87.50 & 89.00 \\
moons (4) & 90.50 & 97.50 & 74.50 & 87.50 & 99.00 & 99.00 & 99.00 & 91.00 & 98.50 & 92.50 & 98.50 & 98.50 & 97.50 & 95.50 & 94.50 & 96.00 \\
two circles (6) & 63.00 & 62.50 & 48.50 & 62.50 & 59.50 & 58.00 & 59.00 & 57.50 & 59.50 & 60.50 & 64.00 & 65.00 & 58.50 & 64.00 & 63.50 & 65.00 \\
\midrule
CMC (23) & 59.50 & 61.00 & 68.00 & 63.50 & 50.50 & 52.50 & 51.00 & 65.00 & 72.00 & 62.50 & 53.50 & 53.50 & 63.00 & 56.00 & 55.50 & 59.00 \\
Pollen (871) & 66.00 & 74.50 & 66.00 & 67.50 & 60.00 & 63.00 & 64.50 & 69.50 & 66.00 & 60.00 & 59.50 & 58.00 & 63.50 & 64.50 & 62.50 & 67.00 \\
Climate (1467) & 100.00 & 100.00 & 100.00 & 94.50 & 92.50 & 98.50 & 100.00 & 94.50 & 91.00 & 98.00 & 97.50 & 97.50 & 94.50 & 99.50 & 99.50 & 100.00 \\
LED (40496) & 88.00 & 87.00 & 82.50 & 91.50 & 74.00 & 76.50 & 84.50 & 69.00 & 84.50 & 85.50 & 94.50 & 91.50 & 92.00 & 91.50 & 91.50 & 90.50 \\
\midrule
Average & 79.94 & \textbf{85.94} & 75.5 & 79.5 & 79.4 & 81.28 & 82.50 & 76.00 & 82.50 & 82.22 & \textbf{84.11} & 83.72 & 83.06 & 83.39 & 81.17 & \textbf{84.5} \\
\bottomrule
\end{tabular} 
}
\end{center}
\label{table:clf_inductive_bias}
\end{table}

\paragraph{Quantifying the similarity of decision boundaries.}
To further verify whether \lift{}s behave similarly to any standard algorithm, we quantify the similarity between the decision boundaries of \lift{}/\gptt{} and those of the baselines.
Specifically, the similarity score is  the percentage of the exact classification matches between \lift{}/\gptt{} and the compared method.
We randomly sample two sets of $200$ data points from the original dataset for training and evaluation, respectively, and the results of all methods are reported in Table~\ref{table:clf_inductive_bias}.
Based on the similarity score, while we observe no similar discernible pattern between \lift{}/\gptt{} and the baselines, 
we find that \lift{}/\gptt{} appear to share the most similar behavior pattern to RBF-SVM, random forest (E=100), and MLP (W=100).

\subsubsection{How Robust Is \lift{}?}\label{app:robustness}

\begin{figure}[h]
    \begin{subfigure}[a]{\textwidth}
        \centering
        \includegraphics[width=\textwidth]{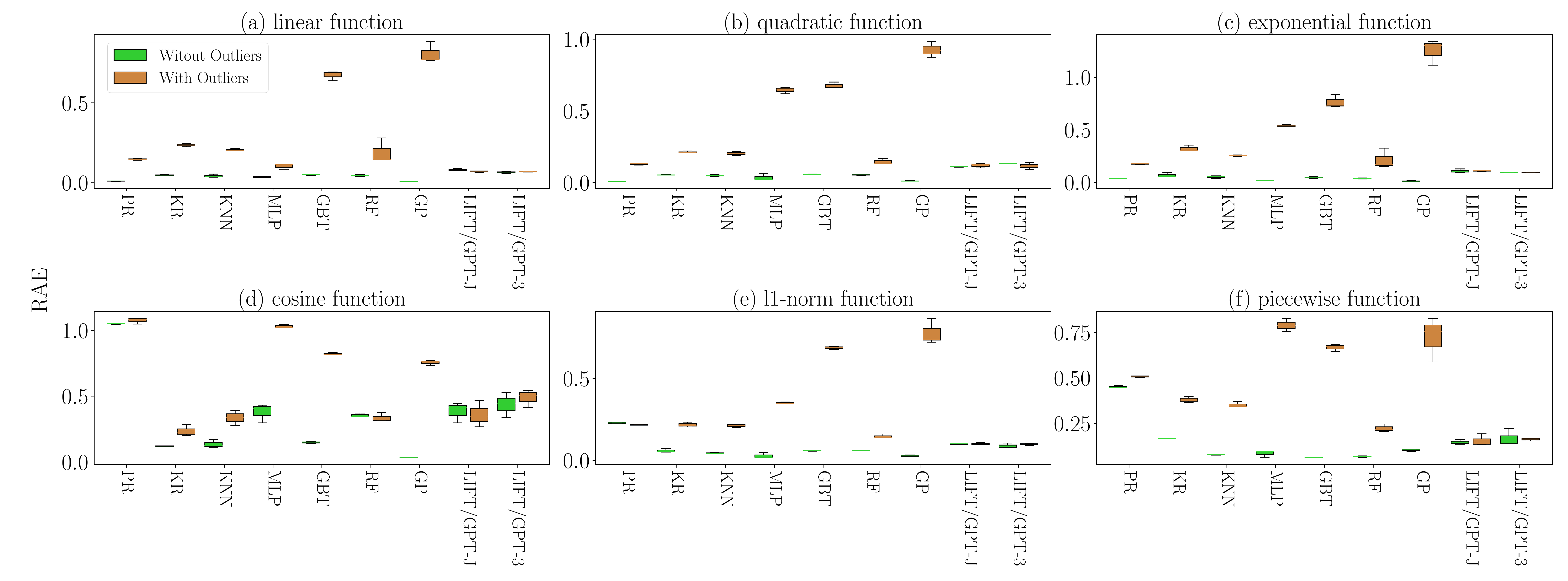}
        \caption{When 2\% of datasets are outliers}
    \end{subfigure}
    \begin{subfigure}[b]{\textwidth}
         \centering
        \includegraphics[width=\textwidth]{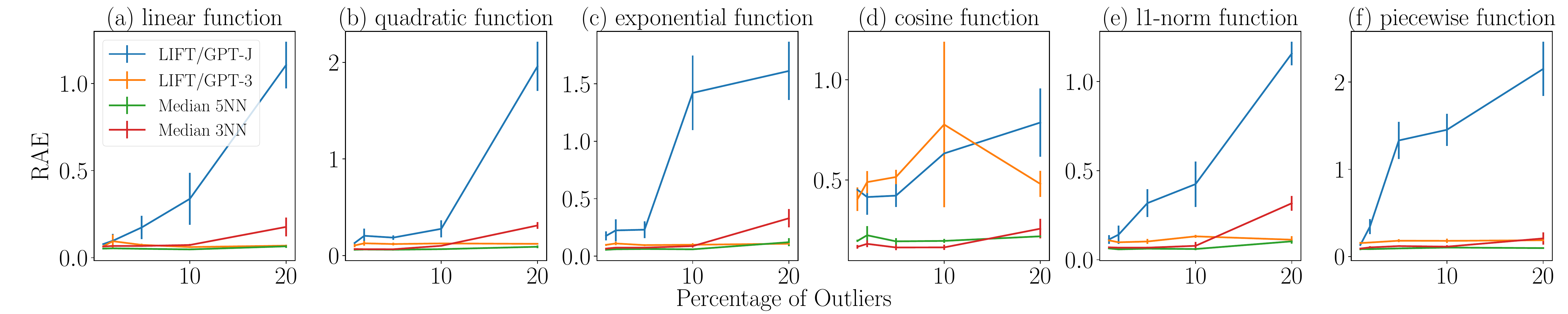}
        \caption{For various portions of outliers  (1\%, 2\%, 5\%, 10\%, 20\%)} 
    \end{subfigure}
    \caption{
    \textbf{Comparing robustness of methods against outliers on regression tasks when the datasets contain (a) 2\% outliers and (b) various portions of outliers.}
    We report each algorithm's regression error measured by Related Absolute Error (RAE).
  \textit{ (a)} When training datasets contain 2\% outliers, all \lift{} models are highly robust against outliers compared with baselines. 
    \textit{(b)} When we increase the fraction of outliers (up to 20\%), \lift{}/\gptt{} is comparable to the strong baseline (median KNN), while \lift{}/\gptj{} fails.
    }
    \label{fig:app_outliers}
\end{figure}

\paragraph{Robustness to outliers in training data.}
Fig.~\ref{fig:app_outliers} visualizes the outlier robustness results discussed in  Sec.~\ref{sec:robutness:outlier}. 

\begin{figure}[h]
    \centering
    \includegraphics[width=\textwidth]{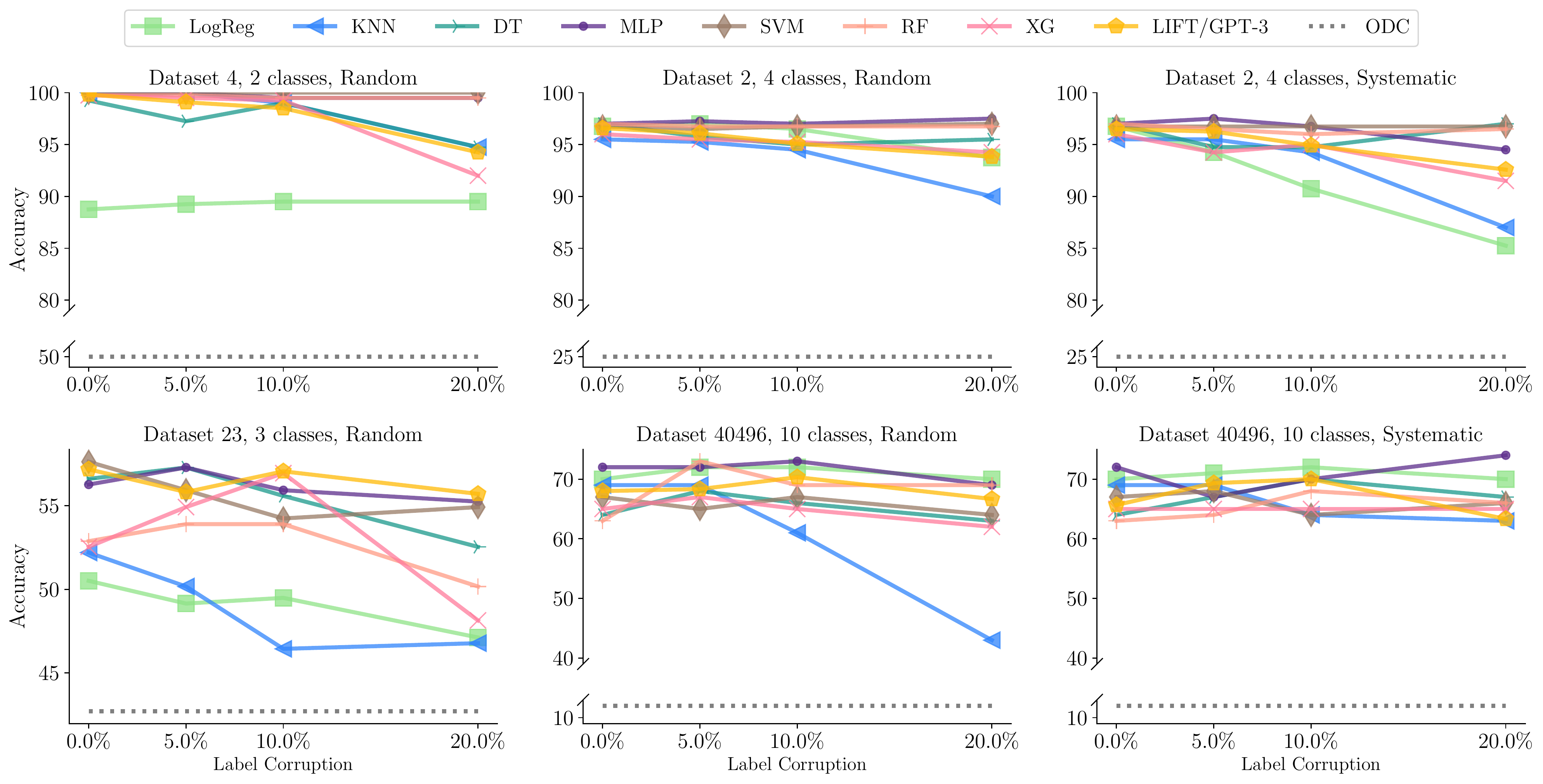}
    \caption{
    \textbf{Robustness against label corruption.}
    Each figure presents classification accuracies ($\uparrow$) evaluated  under different percentages of corruption in the training data (0\% -- 20\%).
    We use synthetic data \texttt{Blobs} and \texttt{Moons} (ID 2 \& 4) and real OpenML datasets \texttt{CMC} and \texttt{LED} (ID 23 \& 40496). 
  We simulate \textit{random errors} (the first two columns) and \textit{systematic errors} (the last column). 
  \lift{}/\gptt{} displays robustness across the datasets.
  \textit{Here, ODC denotes optimal deterministic classifier, which is identical to the majority class classifier (MCC) in the main paper.}
    }
    \label{fig:app_label_corruption}
\end{figure}

\paragraph{Robustness to label corruption.}
\label{sec:label_corruption}
We choose a subset of samples and corrupt the label of the chosen samples, using two corruption schemes: (1) \textit{random errors} (randomly select another label with an equal probability for all labels) and (2) \textit{systematic errors}~\cite{shen2019learning} (replace a label with its next label in the target label list, \textit{e.g.}, $0 \rightarrow 1, 1\rightarrow 2, 2\rightarrow 0$ for a 3-way classification).
As shown in Fig.~\ref{fig:app_label_corruption} , \lift{}/\gptt{} can perform well under label corruption; it follows the general trend of other baselines, not outperforming or underperforming. 
Note that \lift{}/\gptt{} almost always displays greater robustness than KNN.

Table~\ref{table:clf_label_corruption} and Table~\ref{table:clf_systematic_label_corruption} extend the results reported in Fig.~\ref{fig:app_label_corruption}. These additional datasets follow a similar trend to what was discussed in Sec.~\ref{sec:robustness}.

\begin{table}[h]
\caption{\textbf{Accuracy($\uparrow$) comparison of various methods fitted to \textit{randomly} corrupted classification labels.}
In this regime, we corrupt a sample by assigning it another random label in the label space.
}
\begin{center}
\begin{small}
\setlength\tabcolsep{3pt}
\tiny{
\begin{tabularx}{0.87\textwidth}{llcccccccccc}
\toprule[.4mm]
Dataset (ID)& Corruption & \textbf{MCC} & \textbf{LogReg} & \textbf{KNN} & \textbf{DT} & \textbf{MLP} & \textbf{SVM} & \textbf{RF} & \textbf{XG} & \textbf{\lift{}/GPT-3} & \textbf{\lift{}/GPT-J}\\ \midrule
\multirow{4}{*}{Blobs (2)}
& 0\%  & 25.00 & 96.75 & 95.50 & 97.00 & 97.00 & 96.75 & 97.00 & 96.00 & 96.58 & 96.17 \\
& 5\%  & 25.00 & 97.00 & 95.25 & 95.75 & 97.25 & 96.50 & 96.75 & 95.50 & 96.08 & 94.83 \\
& 10\% & 25.00 & 96.50 & 94.50 & 95.00 & 97.00 & 96.75 & 96.75 & 95.25 & 95.08 & 91.38 \\
& 20\% & 25.00 & 93.75 & 90.00 & 95.50 & 97.50 & 97.00 & 96.75 & 94.25 & 93.83 & 83.12 \\
\midrule
\multirow{4}{*}{Moons (4)}
& 0\%  & 50.00 & 88.75 & 100.00 & 99.25 & 99.75 & 100.00 & 99.75 & 99.75 & 99.83 & 99.58 \\
& 5\%  & 50.00 & 89.25 & 100.00 & 97.25 & 100.00 & 100.00 & 99.75 & 99.50 & 99.08 & 96.50 \\
& 10\% & 50.00 & 89.50 & 99.00 & 99.00 & 99.50 & 100.00 & 99.50 & 99.25 & 98.50 & 94.00 \\
& 20\% & 50.00 & 89.50 & 94.75 & 94.75 & 99.50 & 100.00 & 99.50 & 92.00 & 94.25 & 79.88 \\
\midrule
\multirow{4}{*}{CMC (23)}
& 0\%  & 42.71 & 50.51 & 52.20 & 56.61 & 56.27 & 57.63 & 52.88 & 52.54 & 57.18 & 49.83 \\
& 5\%  & 42.71 & 49.15 & 50.17 & 57.29 & 57.29 & 55.93 & 53.90 & 54.92 & 55.82 & 50.28 \\
& 10\% & 42.71 & 49.49 & 46.44 & 55.59 & 55.93 & 54.24 & 53.90 & 56.95 & 57.06 & 48.47 \\
& 20\% & 42.71 & 47.12 & 46.78 & 52.54 & 55.25 & 54.92 & 50.17 & 48.14 & 55.71 & 45.42 \\
\midrule
\multirow{4}{*}{TAE (48)}
& 0\%  & 35.48 & 51.61 & 61.29 & 67.74 & 58.06 & 61.29 & 77.42 & 64.52 & 50.54 & 61.29 \\
& 5\%  & 35.48 & 54.84 & 61.29 & 67.74 & 45.16 & 67.74 & 64.52 & 74.19 & 45.16 & 53.76 \\
& 10\% & 35.48 & 41.94 & 45.16 & 54.84 & 32.26 & 32.26 & 51.61 & 54.84 & 52.69 & 46.24 \\
& 20\% & 35.48 & 29.03 & 48.39 & 48.39 & 32.26 & 45.16 & 48.39 & 45.16 & 47.31 & 35.48 \\
\midrule
\multirow{4}{*}{Pollen (871)}
& 0\%  & 50.00 & 49.09 & 46.88 & 48.96 & 49.22 & 51.56 & 45.97 & 48.31 & 49.57 & 50.39 \\
& 5\%  & 50.00 & 51.43 & 48.18 & 49.22 & 50.26 & 49.74 & 48.44 & 46.62 & 50.65 & 48.61 \\
& 10\% & 50.00 & 48.70 & 48.70 & 47.27 & 50.00 & 46.49 & 51.17 & 47.01 & 48.96 & 48.66 \\
& 20\% & 50.00 & 50.39 & 49.22 & 50.52 & 47.01 & 50.52 & 47.14 & 49.61 & 50.74 & 50.82 \\
\midrule
\multirow{4}{*}{Climate (1467)}
& 0\%  & 91.67 & 89.81 & 89.81 & 91.67 & 91.67 & 87.96 & 91.67 & 90.74 & 91.67 & 87.04 \\
& 5\%  & 91.67 & 89.81 & 91.67 & 87.04 & 91.67 & 91.67 & 90.74 & 87.96 & 91.36 & 85.49 \\
& 10\% & 91.67 & 90.74 & 88.89 & 90.74 & 88.89 & 91.67 & 91.67 & 88.89 & 91.67 & 83.80 \\
& 20\% & 91.67 & 90.74 & 81.48 & 83.33 & 88.89 & 91.67 & 89.81 & 87.04 & 91.67 & 76.39 \\
\midrule
\multirow{4}{*}{LED (40496)}
& 0\%  & 12.00 & 70.00 & 69.00 & 64.00 & 72.00 & 67.00 & 63.00 & 65.00 & 68.00 & 67.33 \\
& 5\%  & 12.00 & 72.00 & 69.00 & 68.00 & 72.00 & 65.00 & 73.00 & 67.00 & 68.33 & 60.33 \\
& 10\% & 12.00 & 72.00 & 61.00 & 66.00 & 73.00 & 67.00 & 69.00 & 65.00 & 70.33 & 56.00 \\
& 20\% & 12.00 & 70.00 & 43.00 & 63.00 & 69.00 & 64.00 & 69.00 & 62.00 & 66.67 & 47.00 \\
\bottomrule[.4mm]
\end{tabularx} }
\end{small}
\end{center}
\label{table:clf_label_corruption}
\end{table}
\begin{table}[h]
\caption{
\textbf{Accuracies($\uparrow$) of various methods fitted to \textit{systematically} corrupted classification labels.}
In this regime, we corrupt a label by assigning all corrupted labels of one class to a single label.}
\centering
\setlength\tabcolsep{3pt}
\resizebox{0.9\textwidth}{!}{
    \begin{tabular}{llcccccccccc}
    \toprule[.4mm]
    Dataset ID & Corruption & \textbf{MCC} & \textbf{LogReg} & \textbf{KNN} & \textbf{DT} & \textbf{MLP} & \textbf{SVM} & \textbf{RF} & \textbf{XG} & \textbf{\lift{}/GPT-3} & \textbf{\lift{}/GPT-J}\\ \midrule
    \multirow{4}{*}{Blobs (2)}
    & 0\%  & 25.00 & 96.75 & 95.50 & 97.00 & 97.00 & 96.75 & 97.00 & 96.00 & 96.50 & 96.17 \\
    & 5\%  & 25.00 & 94.25 & 95.50 & 94.75 & 97.50 & 96.75 & 96.50 & 94.25 & 96.25 & 94.75 \\
    & 10\% & 25.00 & 90.75 & 94.25 & 94.75 & 96.75 & 96.75 & 96.00 & 95.00 & 94.92 & 90.17 \\
    & 20\% & 25.00 & 85.25 & 87.00 & 97.00 & 94.50 & 96.75 & 96.50 & 91.50 & 92.58 & 81.07 \\
    \midrule
    \multirow{4}{*}{LED (40496)}
    & 0\%  & 12.00 & 70.00 & 69.00 & 64.00 & 72.00 & 67.00 & 63.00 & 65.00 & 65.67 & 67.33 \\
    & 5\%  & 12.00 & 71.00 & 69.00 & 67.00 & 67.00 & 68.00 & 64.00 & 65.00 & 69.33 & 58.00 \\
    & 10\% & 12.00 & 72.00 & 64.00 & 70.00 & 70.00 & 64.00 & 68.00 & 65.00 & 70.00 & 55.67 \\
    & 20\% & 12.00 & 70.00 & 63.00 & 67.00 & 74.00 & 66.00 & 66.00 & 65.00 & 63.33 & 53.00 \\
    \bottomrule[.4mm]
    \end{tabular}
    }
\label{table:clf_systematic_label_corruption}
\end{table}

\paragraph{Robustness to class-imbalance of training data.}
\label{sec:class_imbalance}
\begin{table}[!htb]
\vspace{1mm}
\caption{
\textbf{Comparing accuracy ($\uparrow$), F1 ($\uparrow$), Precision ($\uparrow$), and Recall ($\uparrow$) on imbalanced datasets in OpenML (\texttt{Pizza}, \texttt{Climate}, \texttt{Customers}). }
All datasets are for binary classification and are highly imbalanced. The class-imbalance ratio (Imb. Ratio) is defined as 
the ratio of the number of samples in the majority class and that in the minority class. 
Here, DC-0 and DC-1 refer to deterministic classifiers that constantly predict all samples as class $0$ and $1$ respectively.
MCC refers to the \mcc{} that returns the major class learned from the training dataset.
\lift{}/\gpt{}s achieve comparably high scores across the three tasks.
For instance, \lift{}/\gptj{} achieves the best F1 on datasets \texttt{Pizza} and \texttt{Customers}. 
}
  \centering
  \resizebox{\textwidth}{!}{
    \begin{tabular}{ccl|ccc|ccccccc|cc}
    \toprule[.5mm]
    Dataset (ID)   & Imb. Ratio &       & \textbf{MCC} & \textbf{DC-0} & \textbf{DC-1} & \textbf{LogReg} & \textbf{KNN}  & \textbf{DT}  & \textbf{MLP}   & \textbf{RBF-SVM}   & \textbf{\randomf{}}   & \textbf{\xg{}} & \textbf{\lift{}/\gptj{}}  & \textbf{\lift{}/\gptt{}} \bigstrut\\
    \midrule
    \multirow{4}[2]{*}{Pizza (1444)} & \multirow{4}[2]{*}{7.36} & Accuracy & \textbf{88.04} & \textbf{88.04} & 11.96 & 86.92\relscale{0.8}{$\pm$0.23} & 87.56\relscale{0.8}{$\pm$0.68} & 87.24\relscale{0.8}{$\pm$0.60} & 86.28\relscale{0.8}{$\pm$1.37} & \textbf{88.04\relscale{0.8}{$\pm$0.00}} & \textbf{88.04\relscale{0.8}{$\pm$1.04}} & \textbf{88.04\relscale{0.8}{$\pm$0.68}} & 83.89\relscale{0.8}{$\pm$0.45} & 85.17\relscale{0.8}{$\pm$1.35} \bigstrut[t]\\
          & & F1    & 0.00 & 0.00  & 21.37  & 10.77\relscale{0.8}{$\pm$2.74} & 16.84\relscale{0.8}{$\pm$5.26} & 9.01\relscale{0.8}{$\pm$12.74} & 11.77\relscale{0.8}{$\pm$3.43} & 0.00\relscale{0.8}{$\pm$0.00} & 15.79\relscale{0.8}{$\pm$6.45} & 35.50\relscale{0.8}{$\pm$3.68} & \textbf{35.83\relscale{0.8}{$\pm$3.61}} & 24.52\relscale{0.8}{$\pm$1.78} \\
          & & Precision & 0.00 & 0.00  & 11.96 & 28.97\relscale{0.8}{$\pm$3.41} & 42.86\relscale{0.8}{$\pm$10.10} & 13.89\relscale{0.8}{$\pm$19.64} & 32.69\relscale{0.8}{$\pm$12.26} & 0.00\relscale{0.8}{$\pm$0.00} & 51.67\relscale{0.8}{$\pm$23.21} & \textbf{52.21\relscale{0.8}{$\pm$7.28}} & 38.84\relscale{0.8}{$\pm$5.24} &  32.41\relscale{0.8}{$\pm$6.55} \\
          & & Recall & 0.00 & 0.00  & \textbf{100.00} & 6.67\relscale{0.8}{$\pm$1.89} & 10.67\relscale{0.8}{$\pm$3.77} & 6.67\relscale{0.8}{$\pm$9.43} & 8.00\relscale{0.8}{$\pm$3.27} & 0.00\relscale{0.8}{$\pm$0.00} & 9.33\relscale{0.8}{$\pm$3.77} & 28.00\relscale{0.8}{$\pm$5.66} & 33.33\relscale{0.8}{$\pm$2.36} & 20.00\relscale{0.8}{$\pm$0.00} \bigstrut[b]\\
    \midrule
        \multirow{4}[2]{*}{Climate (1467)} & 
    \multirow{4}[2]{*}{11.00} &
    Accuracy & \textbf{91.67} & 8.33 &\textbf{91.67} & 88.89\relscale{0.8}{$\pm$0.76} & 90.74\relscale{0.8}{$\pm$0.76} & 88.89\relscale{0.8}{$\pm$2.27} & \textbf{91.67\relscale{0.8}{$\pm$0.00}} & 87.96\relscale{0.8}{$\pm$0.00} & 91.36\relscale{0.8}{$\pm$0.44} & 89.51\relscale{0.8}{$\pm$0.87} & 87.04\relscale{0.8}{$\pm$2.27} & \textbf{91.67\relscale{0.8}{$\pm$0.00}} \bigstrut[t]\\
          & & F1    & \textbf{95.65} & 0.00 &\textbf{95.65} & 94.00\relscale{0.8}{$\pm$0.43} & 95.13\relscale{0.8}{$\pm$0.40} & 94.04\relscale{0.8}{$\pm$1.27} & \textbf{95.65\relscale{0.8}{$\pm$0.00}} & 93.47\relscale{0.8}{$\pm$0.00} & 95.48\relscale{0.8}{$\pm$0.24} & 94.37\relscale{0.8}{$\pm$0.48} & 94.51\relscale{0.8}{$\pm$1.08} & \textbf{95.65\relscale{0.8}{$\pm$1.00}} \\
          & & Precision & 91.67 & 0.00 & 91.67 & \textbf{93.07\relscale{0.8}{$\pm$0.06}} & 91.85\relscale{0.8}{$\pm$0.43} & 92.26\relscale{0.8}{$\pm$1.20} & 91.67\relscale{0.8}{$\pm$0.00} & 93.00\relscale{0.8}{$\pm$0.00} & 91.64\relscale{0.8}{$\pm$0.04} & 92.83\relscale{0.8}{$\pm$0.43} & 92.10\relscale{0.8}{$\pm$0.36} & 91.67\relscale{0.8}{$\pm$0.00} \\
          & & Recall & \textbf{100.00}  & 0.00 &\textbf{100.00} & 94.95\relscale{0.8}{$\pm$0.82} & 98.65\relscale{0.8}{$\pm$0.48} & 95.96\relscale{0.8}{$\pm$2.86} & \textbf{100.00\relscale{0.8}{$\pm$0.00}} & 93.94\relscale{0.8}{$\pm$0.00} & 99.66\relscale{0.8}{$\pm$0.48} & 95.96\relscale{0.8}{$\pm$0.82} & 97.08\relscale{0.8}{$\pm$2.57} & \textbf{100.00\relscale{0.8}{$\pm$0.00}} \bigstrut[b]\\
    \midrule
    \multirow{4}[2]{*}{Customers (1511)} & 
    \multirow{4}[2]{*}{2.14} &
    Accuracy & 68.18  & 68.18 & 31.82 & 87.12\relscale{0.8}{$\pm$0.54} & \textbf{88.64\relscale{0.8}{$\pm$0.00}} & 85.98\relscale{0.8}{$\pm$0.53} & 86.36\relscale{0.8}{$\pm$1.86} & 86.36\relscale{0.8}{$\pm$0.00} & 85.23\relscale{0.8}{$\pm$0.00} & 85.23\relscale{0.8}{$\pm$0.00} & 85.23\relscale{0.8}{$\pm$1.61} & 84.85\relscale{0.8}{$\pm$1.42} \bigstrut[t]\\
          & & F1    & 0.00 & 0.00 & 48.28 & 79.76\relscale{0.8}{$\pm$0.89} & 80.51\relscale{0.8}{$\pm$0.36} & 78.60\relscale{0.8}{$\pm$1.00} & 77.80\relscale{0.8}{$\pm$2.79} & 78.82\relscale{0.8}{$\pm$0.35} & 76.64\relscale{0.8}{$\pm$0.39} & 76.91\relscale{0.8}{$\pm$0.39} & \textbf{84.43\relscale{0.8}{$\pm$1.43}} & 75.28\relscale{0.8}{$\pm$2.60} \\
          & & Precision & 0.00 & 0.00 & 31.82 & 79.79\relscale{0.8}{$\pm$1.23} & \textbf{88.64\relscale{0.8}{$\pm$1.61}} & 76.40\relscale{0.8}{$\pm$0.38} & 81.00\relscale{0.8}{$\pm$4.65} & 77.94\relscale{0.8}{$\pm$0.90} & 77.14\relscale{0.8}{$\pm$0.91} & 76.50\relscale{0.8}{$\pm$0.91} & 87.57\relscale{0.8}{$\pm$7.11} & 78.18\relscale{0.8}{$\pm$1.97} \\
          & & Recall & 0.00 & 0.00 & 100.00 & 79.76\relscale{0.8}{$\pm$1.68} & 73.81\relscale{0.8}{$\pm$1.68} & 80.95\relscale{0.8}{$\pm$1.68} & 75.00\relscale{0.8}{$\pm$2.91} & 79.76\relscale{0.8}{$\pm$1.68} & 76.19\relscale{0.8}{$\pm$1.68} & 77.38\relscale{0.8}{$\pm$1.68} & \textbf{82.61\relscale{0.8}{$\pm$7.10}} & 72.62\relscale{0.8}{$\pm$3.37} \bigstrut[b]\\
    \bottomrule[.5mm]
    \end{tabular}
}
\vspace{1mm}
 \label{tab:imbalanced}
\end{table}
We evaluate \lift{} on class-imbalanced classification tasks (OpenML datasets \texttt{Pizza}, \texttt{Climate}, and \texttt{Customers} having IDs 1444, 1467, and 1511), shown in Table~\ref{tab:imbalanced}.
We use additional metrics: F1, precision, and recall (higher scores indicate better performance), which are considered as better measurements for the imbalanced data than the accuracy.
The higher values of the MCC's accuracy imply the higher levels of imbalance in the data (50\% shows the perfect balance).
For the reference, we report the performances of the deterministic classifiers that always return the label of class $0$ (DC-0) and class $1$ (DC-1). 

Though evaluated datasets all have high class-imbalance ratios, 
we find that \lift{} can perform well, achieving high F1, precision, and recall scores across the tasks.
For instance, on the \texttt{Customers} dataset (the class-imbalance ratio is nearly $8$), MCC gets $0$ for both precision and recall as all predicted labels of MCC are $0$ (\textit{that is}, the major class), while \lift{}/\gptj{} achieves the best recall ($82.61\pm7.10$) and F1 scores ($84.43\pm1.43$).
Here, the $0$ value of precision and recall in MCC means that
MCC classifies all samples as negative, which is the major class in the training dataset.

\paragraph{Robustness to feature corruption on test data.}\label{sec:feat_corrupt}
Here we provide detailed experiment setting and experiment results on random noise perturbation. 
Given a perturbation budget $\eps \geq 0$, we consider two types of perturbation $\boldsymbol{\delta}$ with $\lVert \boldsymbol{\delta} \rVert_{\infty} \leq \eps$:
random noise and adversarial perturbation~\citep{szegedy2013intriguing}. 
For random noise $\boldsymbol{\delta}$, we test on two types: (1) random Gaussian noise $\boldsymbol{\delta} \sim \mathcal{N}(\mathbf{0}, \mathbf{I}_{p})$ scaled to satisfy $\lVert \boldsymbol{\delta} \rVert_{\infty} = \eps$, and (2) signed constant noise $\boldsymbol{\delta}$ where each element $\delta_i$ has magnitude $\eps$ and random sign. 
For adversarial perturbation $\boldsymbol{\delta}$, we test on the transfer attack~\citep{kurakin2016adversarial}, \ie, we generate an adversarial example $(\vx+\boldsymbol{\delta}, y)$ for a source neural network (that we can access) with constraint $\lVert \boldsymbol{\delta} \rVert \leq \varepsilon$,
and test whether the target network correctly classifies the adversarial examples.

Table~\ref{tab:adv_rob} shows the results of \lift{} and baselines for the MNIST classification problem. 
We test on random noise (Gaussian and signed constant) and PGD attacks transferred from LeNet-5 and MLP. 
We compare the results for three networks: LeNet-5, MLP (having 2 hidden layers, each with 300 neurons and 100 neurons), and \lift{}/\gptt{}. 
\lift{}/\gptt{} is observed to tolerate random noise (both Gaussian and signed constant) for small perturbation radius $\eps=0.01$. 

We do not include the result for \lift{}/\gptj{} since it is not even robust against simple noise. Please refer to Section~\ref{sec:improve_augmentation} to check the vulnerability of \lift{}/\gptj{} against test-time noise and how data augmentation improves the robustness of \lift{}/\gptj{}.

\begin{table}[t]
\vspace{1mm}
\caption{
\textbf{Accuracies ($\uparrow$) of \lift{} and baselines (LeNet-5, MLP) under the perturbation on the input feature of MNIST data.} Given the perturbation budget $\varepsilon \in [0,1]$, we test on four types of perturbations within $L_{\infty}$ ball of radius $\varepsilon$. (1): adding random Gaussian noise that is scaled to reach the $L_{\infty}$ ball, (2): adding \emph{signed constant} noise vector where each element has magnitude $\eps$ and random sign, (3) \& (4): adversarial examples generated from a source network (LeNet-5 \& MLP, respectively) using PGD attack~\citep{madry2017towards} from foolbox~\citep{rauber2017foolbox}.
For small perturbation radii ($\eps=0.01$), 
\lift{}/\gptt{} maintains high accuracy for random noise, both for Gaussian and signed constant noise types. When $\eps = 0.01$ or $\eps=0.1$, the performance of \lift{}/\gptt{} for random noise and transferred adversarial attacks have significant gap, showing that the adversarial examples generated at LeNet-5 and MLP are transferred to \lift{}/\gptt{}. 
}
\centering
\setlength\tabcolsep{2pt}
\scriptsize
\begin{tabular}{c|ccc|ccc|ccc|ccc}
\toprule
Source   & 
\multicolumn{3}{c|}{\textbf{Random noise (Gaussian)}} & 
\multicolumn{3}{c|}{\textbf{Random noise (signed const.)}} &
\multicolumn{3}{c|}{\textbf{PGD attack on LeNet-5}} & 
\multicolumn{3}{c}{\textbf{PGD attack on MLP}} 
\\ [.05in] 
Target   & 
LeNet-5 & MLP 
& \lift{}/\gptt{} & 
LeNet-5 & MLP 
& \lift{}/\gptt{} & 
LeNet-5 & MLP 
& \lift{}/\gptt{} & 
LeNet-5 & MLP 
& \lift{}/\gptt{} \\
\midrule
$\varepsilon=0$  
& 99.22 & 98.09
& 98.15
& 99.22 & 98.09
& 98.15
& 99.22 & 98.09
& 98.15
& 99.22 & 98.09
& 98.15
\\
$\varepsilon = 0.01$  
& 99.25 & 98.05
& 98.28
& 99.26 & 98.08
& 88.05
& 97.27  & 97.77
&  44.88  
& 99.15 & 96.89
& 44.46
 \\
$\varepsilon=0.1$ 
& 99.20 & 97.70
& 88.38
& 99.06 & 97.39
& 68.80
& 26.80  & 93.99
&   33.66  
& 96.98 & 23.12
& 23.62
\\

$\varepsilon=0.3$  
& 98.01 & 87.69
& 54.80
& 79.80 & 74.20
& 29.68
& 0.00  &   36.62
& 20.31   
& 41.51 & 0.00
&  20.29
\\
\bottomrule
\end{tabular}
\vspace{1mm}
\label{tab:adv_rob}
\end{table}

\subsubsection{Does \lift{} Need Large-Scale Models Pretrained on Natural Language Data?}
\label{app:dependency_lms}
Continuing from Sec.~\ref{sec:dependency_lms}, we provide the detailed setup of this experiment. 
We obtain the Gibberish model by fine-tuning the entire \gptj{} model (rather than LoRA~\citep{hu2021lora}) on the Gibberish dataset~\citep{gibberish} for 10 epochs at learning rate \texttt{0.1}.
For \lift{}/Code-Gen, \lift{}/CodeParrot, and \lift{}/Gibberish, we follow the same e as \lift{}/\gptj{}, which we have discussed in Appendix~\ref{app:experiment_setup}.
To fine-tune \lift{}/Rand-\gptj{} for specific tasks, we set the temperature as 1 instead, as we observed the 0 temperature consistently gives us poor performances. 
Note that only 10\%--15\% of \lift{}/Rand-\gptj{}'s outputs of are valid. 
The accuracies listed in the table are computed among valid outputs. 
Other settings of \lift{}/RAnd-\gptj{} are the same as \lift{}/\gptj{}.

\subsection{Results for \lift{}-Specific Learning Properties (Section~\ref{sec:lift_specific})}
\label{app:lift_specific}

\subsubsection{Does \lift{} Benefit from Incorporating Feature Names?}
\label{app:context}
Continuing from Sec.~\ref{sec:feature_names}, we provide more details of experiment settings, further evaluations with \gptj{} models (Table~\ref{tab:clf_feature_name_gptj}), and results on regression tasks (Table~\ref{table:reg_context}) here.

\paragraph{Prompt templates.}
We design five prompts templates to assess how incorporating feature names affects the performance of \lift{}. 
For instance, consider a data sample ``$\vx=(\texttt{English speaker},23,3,\texttt{summer},19)$, $y=3$'' from \texttt{TAE} dataset where the feature names are ``\texttt{native speaker, instructor, course, semester, class size}'', and the target attribute is \texttt{teaching performance}.
We can incorporate the contextual information by either simply replacing the ``$x_i$'' in the prompts with the corresponding feature names or converting this sample into a coherent sentence.
Meanwhile, we also investigate how shuffled feature names affect the performance of \lift{} by designing the prompts accordingly. 
For illustration purposes, we provide the example of the five prompt templates as below.
\begin{itemize}\setlength{\itemindent}{-2em}
    \item \texttt{(W/O Names)} ``\texttt{When we have $x_1=1, x_2=23, x_3=3, x_4=1, x_5=19$, what should be $y$ value?}''
    \item \texttt{(Correct-Names I)} ``\texttt{When we have native speaker=English speaker, course instructor=$23$, course=$3$, semester=summer, class size=19, how is the teaching performance?}''
    \item \texttt{(Correct-Names II)} ``\texttt{In the course 3 offered in the summer semester, there was a native English-speaking teaching assistant and an instructor whose ID is 23. How is the teaching performance?}''
    \item \texttt{(Shuffled-Names I)} ``\texttt{When we have semester=English speaker, class size=$23$, semester=$3$, course instructor=summer, native speaker=19, how is the teaching performance?}''
    \item \texttt{(Shuffled-Names II)} ``\texttt{In the course summer offered in the 3 semester, there was a 19 teaching assistant and an instructor whose ID is summer. How is the teaching performance?}''
\end{itemize}
We note that the sentence generated using the \texttt{(Shuffled-Names II)} template can be incoherent.

\paragraph{Settings.}
\textit{(Datasets)} Among the OpenML datasets evaluated in Table~\ref{tab:classification_accuracy}, we select three datasets: \texttt{CMC}, \texttt{TAE}, and \texttt{Vehicle} (with IDs being 23, 48, and 54) whose all provided feature names are meaningful and relevant to the prediction task and the response values.
\textit{(Baselines)}
We compare our target model \lift{} when feature names are correctly incorporated (\texttt{Correct-Names I, II}) with the versions of \lift{} when feature names are incorrectly incorporated with randomly shuffled orders (\texttt{Shuffled-Names I, II}) and when feature names are not included (\texttt{W/o Names}).
Also, we compare all models with the simple baseline MCC and the strong baseline XGBoost.

\paragraph{Classification.} We provide additional evaluation of \lift{}/\gptj{} on three datasets used in the main paper.
Table~\ref{tab:clf_feature_name_gptj} presents our result with the same settings in the main paper.
We can see that correctly using feature names helps improve the performance of \lift{} from the models without feature names or the models with randomly shuffled feature names.
This finding is consistent with the finding in the main papers on the usefulness of incorporating the feature names.

\begin{table}[!htp]
  \centering
   \caption{\textbf{The effect of using feature names on \lift{}/\gptj{}}. We compare the classification accuracy ($\uparrow$) of \lift{}/\gptj{} when feature names \textit{are and are not} incorporated in  prompts.
  } 
\resizebox{\textwidth}{!}{
    \begin{tabular}{cccccccc}
    \toprule
    \multirow{2}{*}{Dataset (ID)} & \multirow{2}{*}{MCC} & \multirow{2}{*}{XGBoost} & \multicolumn{5}{c}{{\lift{}/\gptj{}}} \bigstrut[t]\\
    [0.1cm]
          &  & & W/o Names & Shuffled-Names I & Shuffled-Names II & Correct-Names I &   Correct-Names II  \bigstrut[b]\\[0.05cm]
    \midrule[0.05cm]
    {CMC (23)} &  {42.71}  &  52.43$\pm$0.42 & 49.49$\pm$0.56 & \textbf{51.30$\pm$1.05} & \textbf{51.30$\pm$2.51} & 48.82$\pm$3.12 & 50.39$\pm$1.05 \bigstrut[t]\\
    {TAE (48)}  & {35.48} &  66.67$\pm$8.05 & 60.22$\pm$4.02 & 63.44$\pm$6.08 & 58.06$\pm$7.90 & 60.21$\pm$10.64 & \textbf{65.59$\pm$8.47} \\
    {Vehicle (54)} &  {25.88} &  73.14$\pm$0.28 & 64.31$\pm$2.37 & 66.87$\pm$1.54 & 65.49$\pm$1.69 &   \multicolumn{2}{c}{\textbf{69.02$\pm$3.67*}} \\
    \bottomrule
    \end{tabular}
    }
\vspace{1mm}
\label{tab:clf_feature_name_gptj}
\end{table}

\paragraph{Regression.} To investigate whether incorporating feature names in prompts improves the regression performance of \lift{}, similar to the datasets selection process of classification tasks, we evaluate the effect of feature names on the datasets \texttt{Insurance} and \texttt{Student}, whose tasks can be helped by common knowledge.
To be more specific, while the task of \texttt{Insurance} dataset is to predict the insurance costs, the key features of \texttt{Insurance} dataset are \texttt{age}, \texttt{body mass index}, and \texttt{smoke or not}, which are intuitively closely related to the task. 
For the \texttt{Student} dataset, the task is to predict students' grades based on their weekly study time, previous grades, etc. Therefore, the features and task of \texttt{Student} are also highly correlated.
Table~\ref{table:reg_context} presents our evaluation of regression tasks.
We find that fine-tuning with feature names does not necessarily help with the regression tasks.

\begin{table}[!htb]
\caption{
\textbf{Investigating if incorporating feature names to \lift{} improves sample efficiency in regression tasks.}
The experiments are conducted on \texttt{Insurance} and \texttt{Student} datasets. 
The second column indicates the fraction of samples used for training the model. 
We observe no significant improvements in the performance when feature names are properly included. 
    }
\begin{center}
\begin{small}
\tiny{
\begin{tabularx}{0.93\textwidth}{l|p{0.5cm}|c|ccccc}
\toprule
\multirow{2}{*}{Dataset}  & \multirow{2}{*}{Frac.}& \multirow{2}{*}{\textbf{\randomf{}}}  & \multicolumn{4}{c}{\textbf{\lift{}/\gptt{}}} \\ 
& &  & W/O Names & Shuffled-Names I & Shuffled-Names II & Correct-Names I & Correct-Names II \\ \toprule
\multirow{5}{*}{insurance} & 0.2 & \textbf{0.31 $\pm$ 0.00} & 0.89 $\pm$0.03 & 0.76 $\pm$0.11 & 0.59 $\pm$0.09 & 0.59 $\pm$0.11 & 0.89 $\pm$0.03 \\
& 0.4  & 0.26 $\pm$ 0.00 &  0.42 $\pm$0.15  &  0.30 $\pm$0.02 & \textbf{0.20 $\pm$0.03}& 0.35 $\pm$0.10 & 0.21 $\pm$0.01\\
& 0.6 & 0.26 $\pm$ 0.00 & 0.30 $\pm$0.10 & 0.24 $\pm$0.03 & \textbf{0.19 $\pm$0.02}  & 0.30 $\pm$0.12 & 0.22 $\pm$0.08  \\
& 0.8 & 0.27 $\pm$ 0.00 & 0.31 $\pm$0.07 &0.19 $\pm$0.04 & 0.18 $\pm$0.03   & 0.14 $\pm$0.01 & \textbf{0.11 $\pm$0.02} \\
& 1.0 & 0.26 $\pm$ 0.00  & 0.14 $\pm$0.05 &  0.17 $\pm$0.03 & 0.19 $\pm$0.01  & 0.17 $\pm$0.04 & \textbf{0.10 $\pm$0.03}\\ \midrule

\multirow{5}{*}{student} & 0.2 & 0.40 $\pm$ 0.00 &  0.32 $\pm$0.01 &0.32 $\pm$0.01 & 0.34 $\pm$0.02   & \textbf{0.31 $\pm$0.01} & \textbf{0.31 $\pm$0.01} \\
& 0.4  & 0.36 $\pm$ 0.00 & 0.32 $\pm$0.02 &  0.31 $\pm$0.01 & \textbf{0.30 $\pm$0.00} & 0.32 $\pm$0.01 & 0.35 $\pm$0.01\\
& 0.6 & 0.36 $\pm$ 0.00  & 0.31 $\pm$0.01 & 0.31 $\pm$0.01 & 0.31 $\pm$0.01 & 0.31 $\pm$0.01 &  \textbf{0.30 $\pm$0.00} \\
& 0.8 & 0.38 $\pm$ 0.00 & 0.28 $\pm$0.01  &\textbf{0.27 $\pm$0.01} & 0.29 $\pm$0.02  & 0.28 $\pm$0.01 & 0.28 $\pm$0.00  \\
& 1.0 & 0.35 $\pm$ 0.00 &\textbf{0.27 $\pm$0.01} & 0.28 $\pm$0.01 & 0.28 $\pm$0.01 & 0.28 $\pm$0.01  & 0.35 $\pm$0.02\\
\bottomrule
\end{tabularx} }
\end{small}
\end{center}
\label{table:reg_context}
\end{table}

\subsubsection{Is \lift{} Calibrated?}\label{app:bayesian}

\begin{figure}
    \centering
    \includegraphics[width=0.95\textwidth]{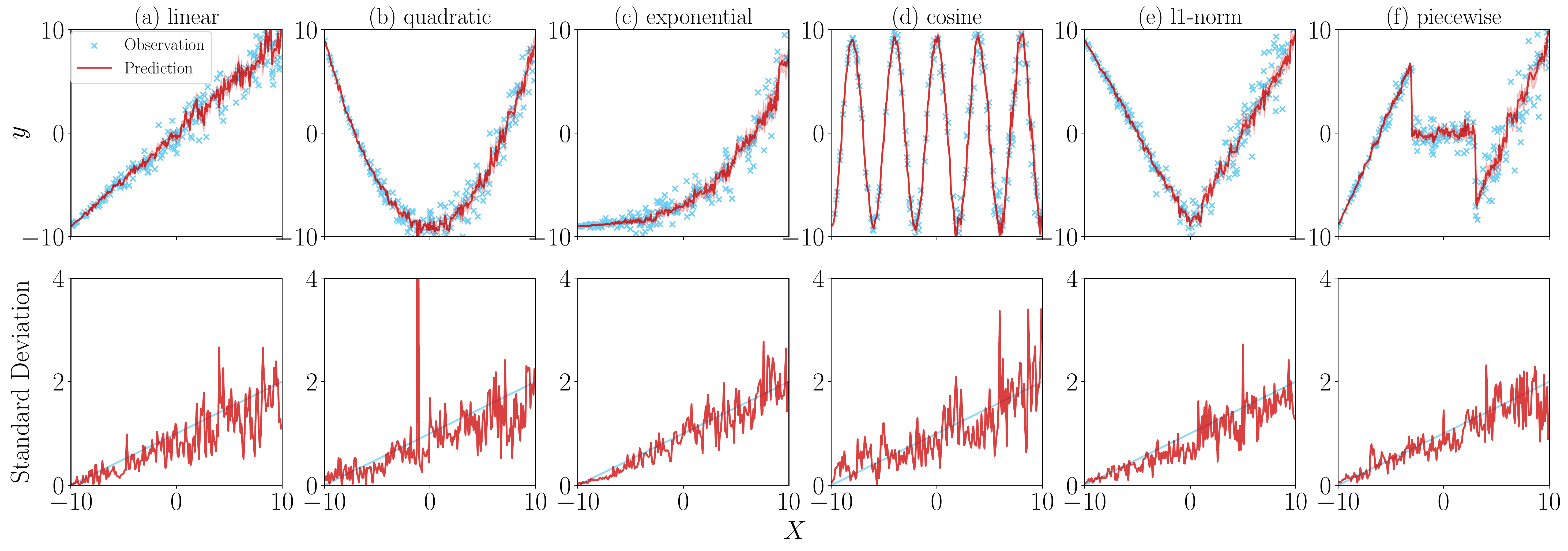}
    \caption{
    \textbf{Investigating calibrated prediction effect of \lift{}.}
    Prediction standard deviations of \lift{}/\gpt{} align well to the observations (top), across datasets, implying the well calibration.
    }
    \label{fig:bayesian}
\end{figure}

\begin{figure}[h]
    \centering
    \includegraphics[width=\textwidth]{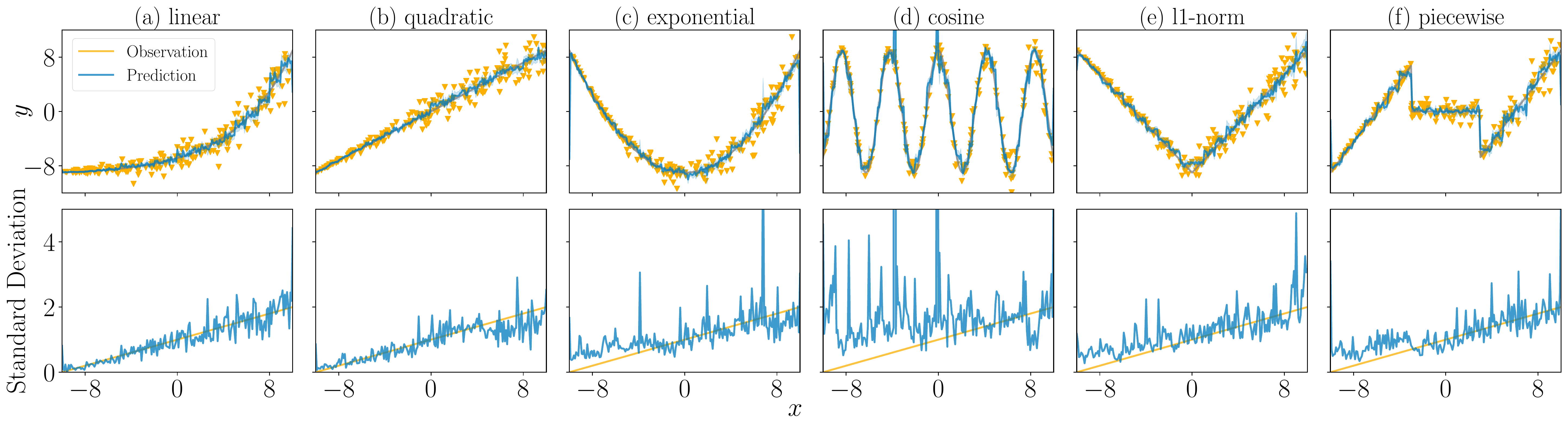}
    \caption{
    \textbf{Visualization of \lift{}/\gptt{} predictions under varying noise levels.}
    The predictions are made on grid datasets consisting of 103 eveny-spaced samples in $[-10,10]$.
    The standard deviation of \lift{}/\gptj{} predictions are computed based 20 repeated random predictions. 
    We observe that the standard deviations of predictions from \lift{}/\gptt{} aligns well with that of noisy training samples (observations), implying that \lift{}/\gptt{} can reflect the confidence, \ie, \lift{}/\gptt{} is calibrated.
    }
    \label{fig:bayesian_gpt3}
\end{figure}

Continuing the discussion in Sec.~\ref{sec:Bayesian}, Fig.~\ref{fig:bayesian_gpt3} indicates that \lift{}/\gptt{} is calibrated.

\subsubsection{Can we use \lift{} for Generation?}\label{app:generator}
Here we provide the detailed experiment setting of Sec.~\ref{sec:generator}. 

\paragraph{Data pre-processing.}
We preprocess the MNIST dataset as below. 
First, we crop each $28 \times 28$ image at the center to make an $18 \times 18$ image. Then, we represent the cropped image as a sequence of $324$ pixel values, where each pixel is an integer in $\{0, 1, \cdots, 255\}$.

The training process for both tasks (i) and (ii) are identical, detailed as below. Given an image in the training set, we put the digit of the image into a prompt and put the sequence of $324$ pixel values into the completion, as in the example shown below. 

\begin{itemize}\setlength{\itemindent}{-2em}
    \item \texttt{(Train Prompt)} ``\texttt{Generate an image of digit 9.\#\#\#}''
    \item \texttt{(Train Completion)} ``\texttt{0 0 0 0 0 0 0 0 0 0 0 0 0 0 0 0 0 0 0 0 0 0 0 0 1 90 255 254 255 96 0 0 0 0 0 0 0 0 0 0 0 24 139 253 253 253 253 247 95 0 0 0 0 0 0 0 0 0 24 211 253 231 134 12 125 251 175 0 0 0 0 0 0 0 0 0 196 253 215 26 0 0 0 63 47 0 0 0 0 0 4 61 21 48 239 211 28 0 0 0 0 14 59 181 206 19 0 0 26 181 64 230 249 63 0 0 0 0 0 165 253 253 253 36 0 0 0 0 166 253 98 0 0 0 0 33 198 252 210 242 253 36 0 0 0 0 234 253 31 0 0 13 63 250 253 207 44 235 253 36 0 0 0 0 234 253 162 6 137 221 253 250 169 25 79 253 224 21 0 0 0 0 148 253 253 253 253 243 161 42 0 0 100 253 186 0 0 0 0 0 6 81 155 155 111 41 0 0 0 0 182 253 186 0 0 0 0 0 0 0 0 0 0 0 0 0 0 18 223 253 84 0 0 0 0 0 0 0 0 0 0 0 0 0 0 32 253 246 54 0 0 0 0 0 0 0 0 0 0 0 0 0 0 129 253 233 0 0 0 0 0 0 0 0 0 0 0 0 0 0 0 135 253 201 0 0 0 0 0 0 0 0 0 0 0 0 0 0 0 186 253 129 0 0 0 0 0 0 0 0 0 0 0 0 0 0 0 239 253 59 0 0 0 0 @@@}''
\end{itemize}

The testing process is different for tasks (i) and (ii). For task (i), we put the prompt as above and check the image drawn by pixel values generated in the completion. For task (ii), the test prompt is set as follows. First, we randomly sample an image with the target digit from the MNIST set. We then include its digit number and the top half pixels (first 162 pixels) in the prompt. 
Once this test prompt is given, we observe the remaining 162 pixel values generated in the completion, and check how the entire 324 pixels look like. Examples of the test prompt are given below.

\begin{itemize}\setlength{\itemindent}{-2em}
    \item \texttt{(Test Prompt for task (i))} ``\texttt{Generate an image of digit 9.\#\#\#}''
    \item \texttt{(Test Prompt for task (ii))} ``\texttt{Generate an image of digit 9.\#\#\# 0 0 0 0 0 0 0 0 0 0 0 0 0 0 0 0 0 0 0 0 0 0 0 0 1 90 255 254 255 96 0 0 0 0 0 0 0 0 0 0 0 24 139 253 253 253 253 247 95 0 0 0 0 0 0 0 0 0 24 211 253 231 134 12 125 251 175 0 0 0 0 0 0 0 0 0 196 253 215 26 0 0 0 63 47 0 0 0 0 0 4 61 21 48 239 211 28 0 0 0 0 14 59 181 206 19 0 0 26 181 64 230 249 63 0 0 0 0 0 165 253 253 253 36 0 0 0 0 166 253 98 0 0 0 0 33 198 252 210 242 253 36 0 0 0 0 234 253 31 0 0 13 63 250 253 207 44 235 253 36 0 0}''
\end{itemize}

\paragraph{Experiment setup.}
For GPT-3, we fine-tune with the pre-processed MNIST training set for 15 epochs. For GPT-J, we fine-tune for three epochs. We use the learning rate $10^{-5}$. 
In the inference phase, as described above, we use different prompts for the two tasks and collect the first 324 and 162 white space-separated tokens in the output for tasks (i) and (ii), respectively.
In addition, for task (ii), we concatenate the provided 162 tokens in the prompt and the 162 tokens from the model output into a sequence of 324 tokens. 
The collected 324 tokens are converted to an $18 \times 18$ image if all of them are valid numbers in $\{0, 1, \cdots, 255\}$. 
We assess the performance of \lift{} as a generator under different temperatures $\in [0, 0.3, 0.5, 0.7, 0.9, 1]$.
Under a fixed temperature, we generate images of each digit 5 times. 
We use the standard perplexity score to evaluate the performance of the obtained generative models.
Here perplexity is a standard metric for assessing the performance of a generative model, which reflects the inverse of the probability of the given data samples being produced by the model.
In this experiment, we assess the generalization performance of models by comparing the perplexity scores between training and test sets. 
If the gap between the training perplexity and the test perplexity is small, it indicates that the model generalizes well.~\citep{blei2003latent}.

\begin{figure}
    \vspace{-0.2in}
    \centering
    \begin{subfigure}[a]{\textwidth}
        \includegraphics[width=0.16\textwidth]{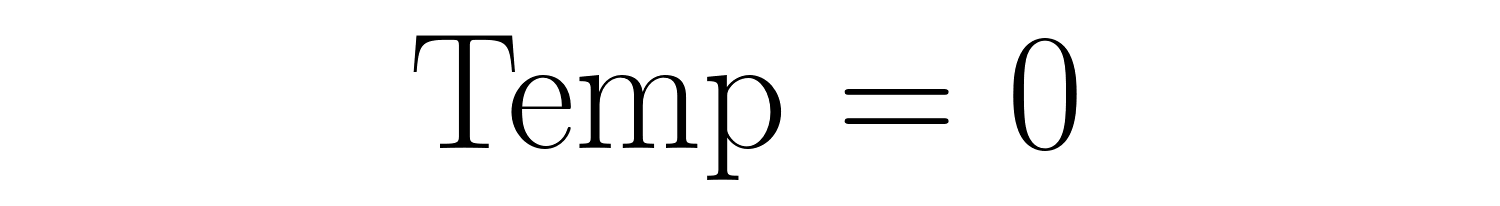}
        \includegraphics[width=0.16\textwidth]{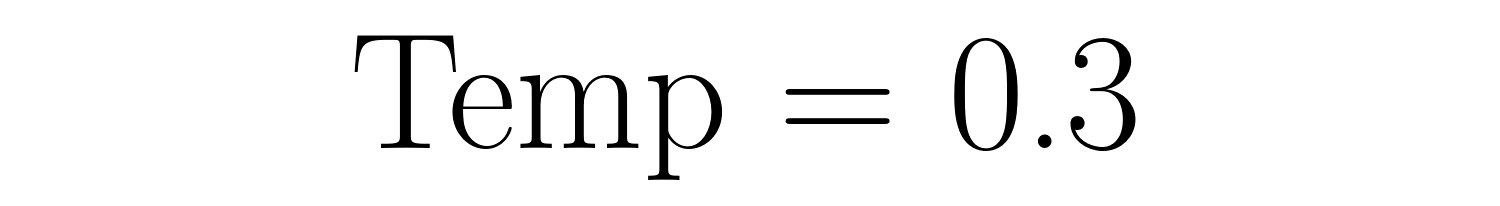}
        \includegraphics[width=0.16\textwidth]{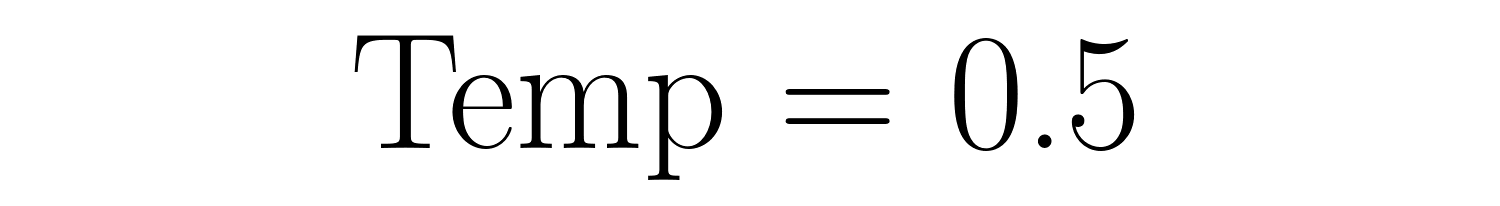}
        \includegraphics[width=0.16\textwidth]{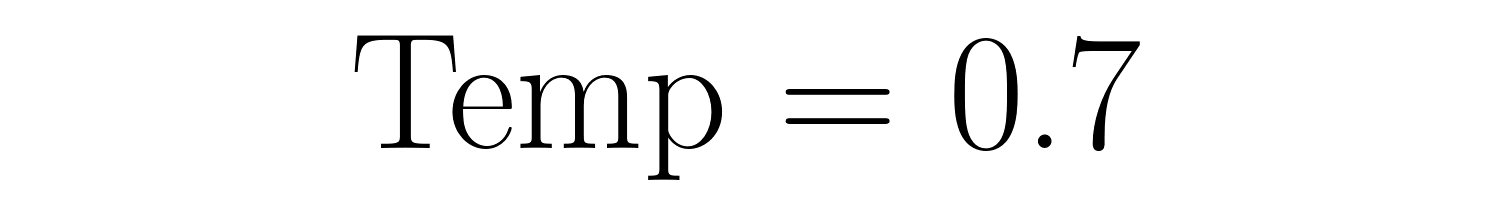}
        \includegraphics[width=0.16\textwidth]{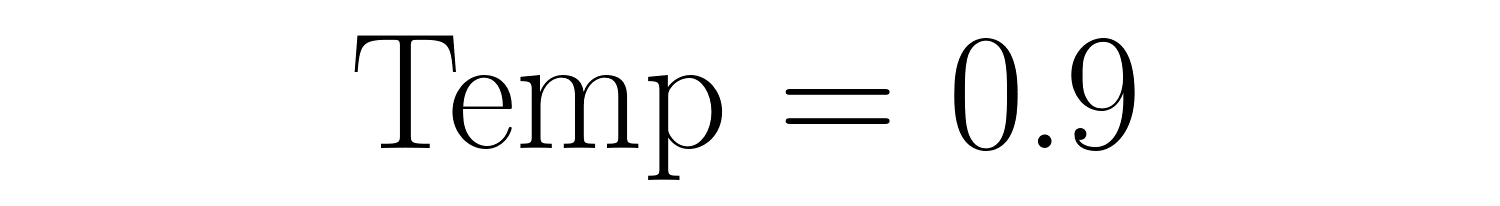}
        \includegraphics[width=0.16\textwidth]{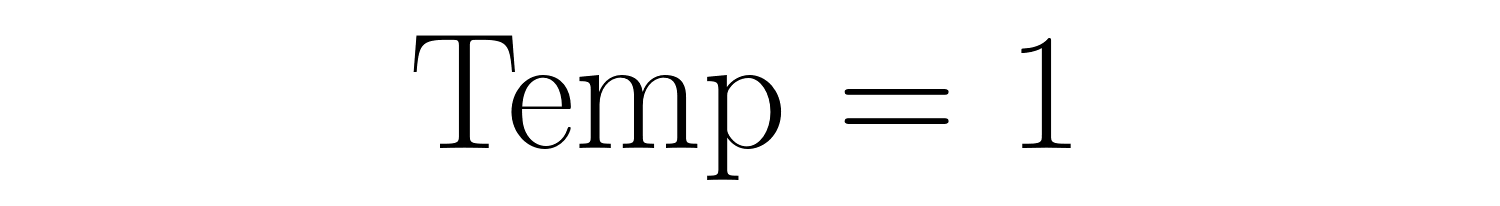}
        \includegraphics[width=0.16\textwidth]{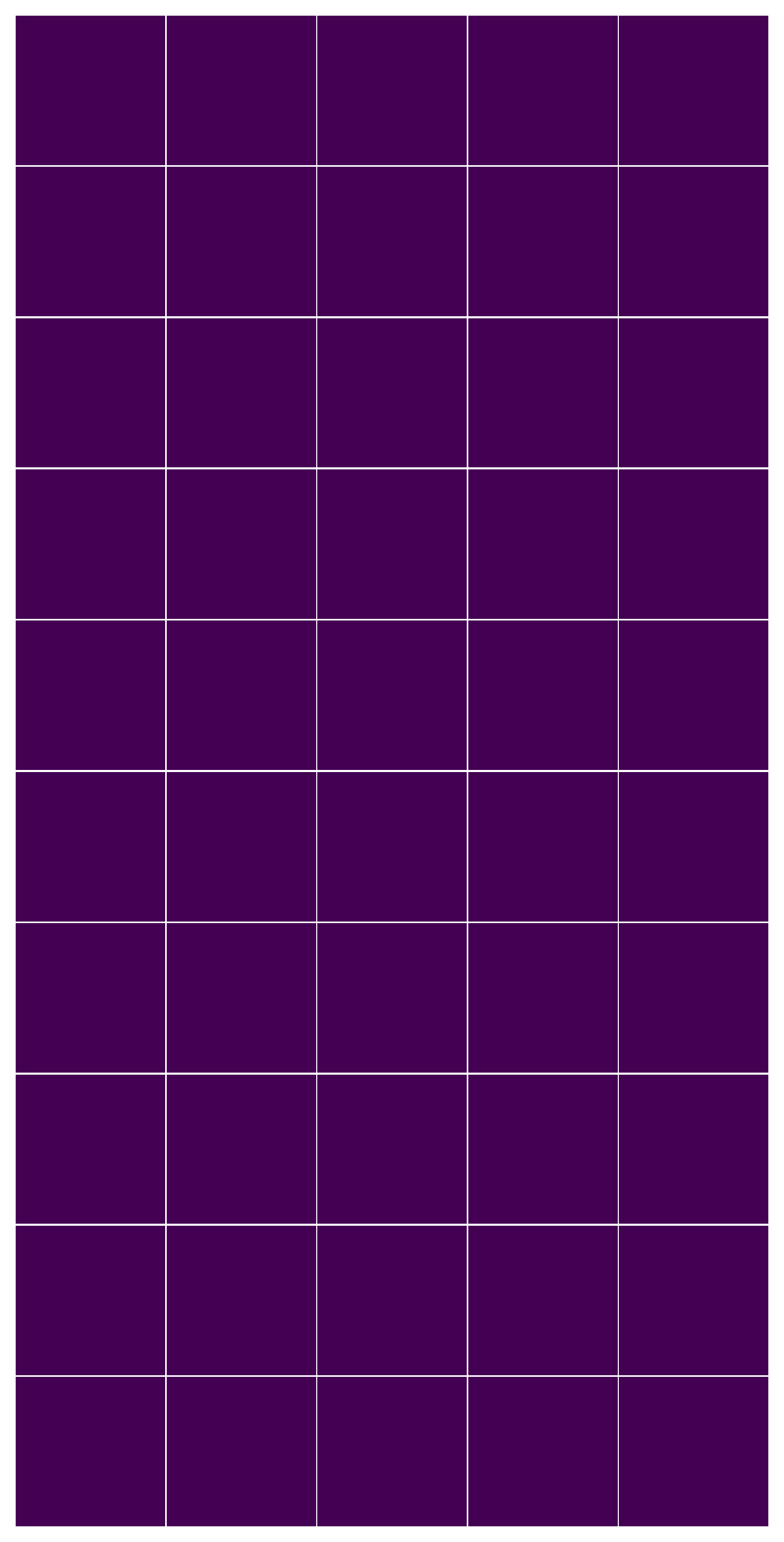}
        \includegraphics[width=0.16\textwidth]{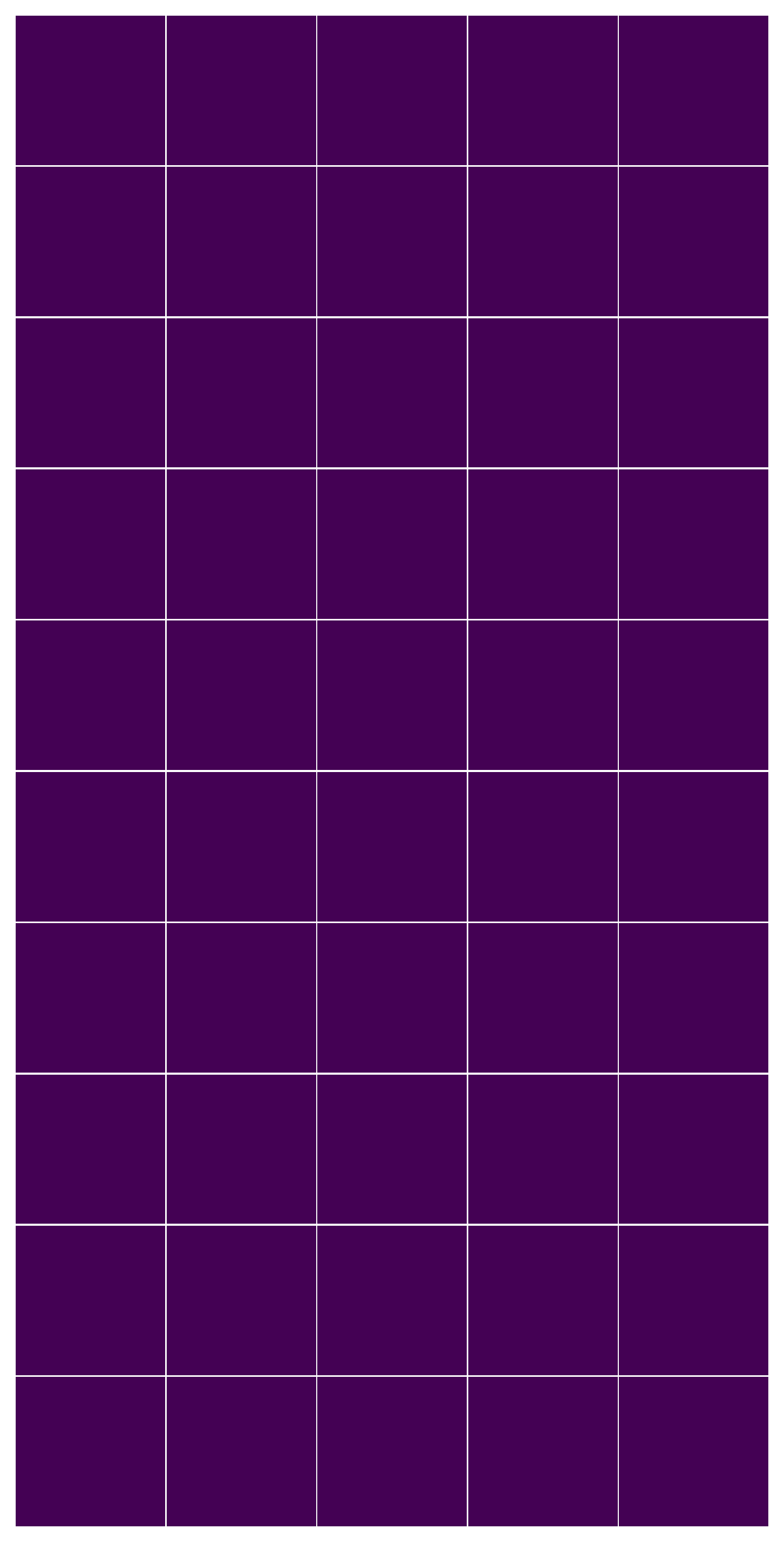}
        \includegraphics[width=0.16\textwidth]{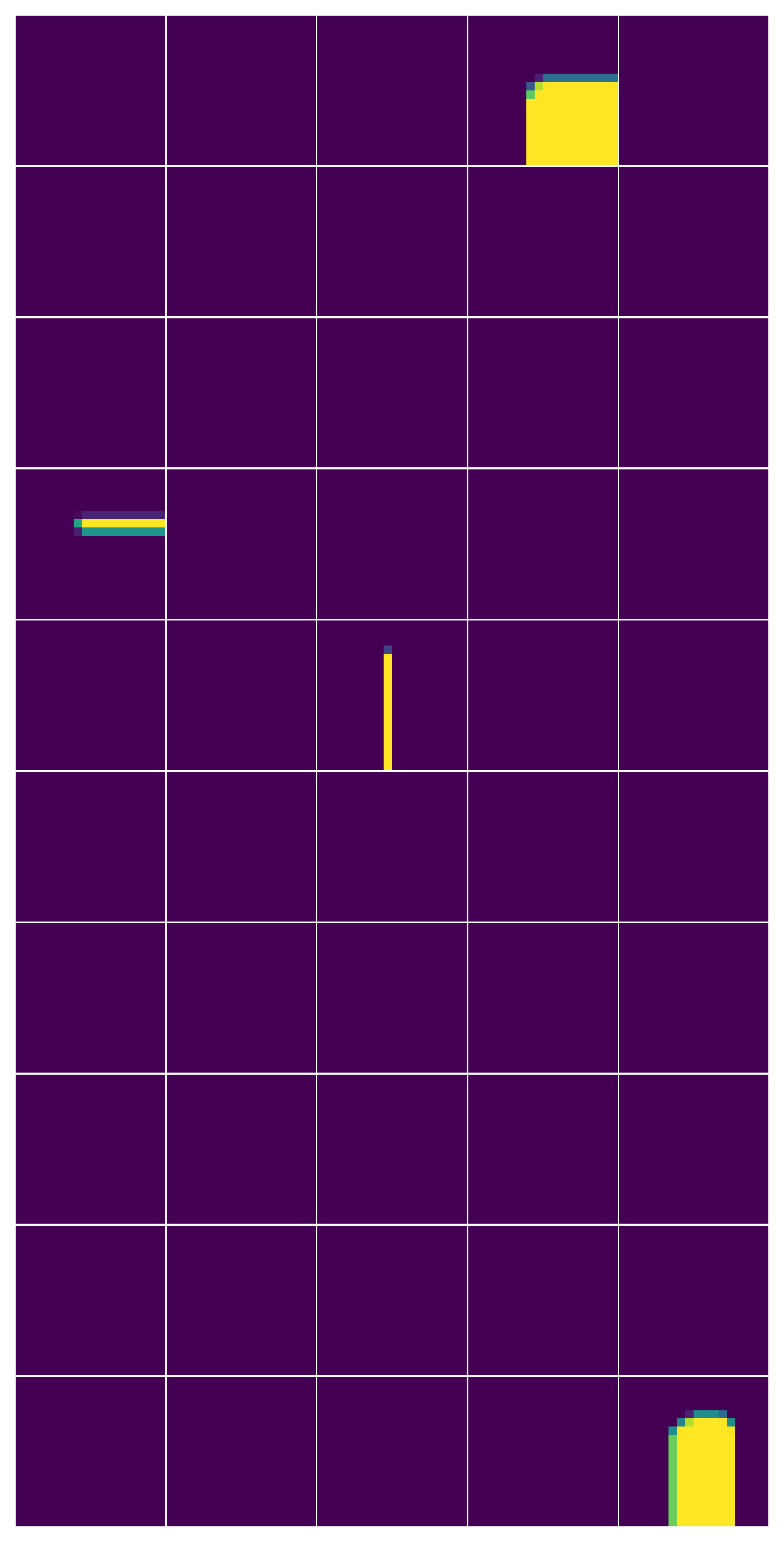}
        \includegraphics[width=0.16\textwidth]{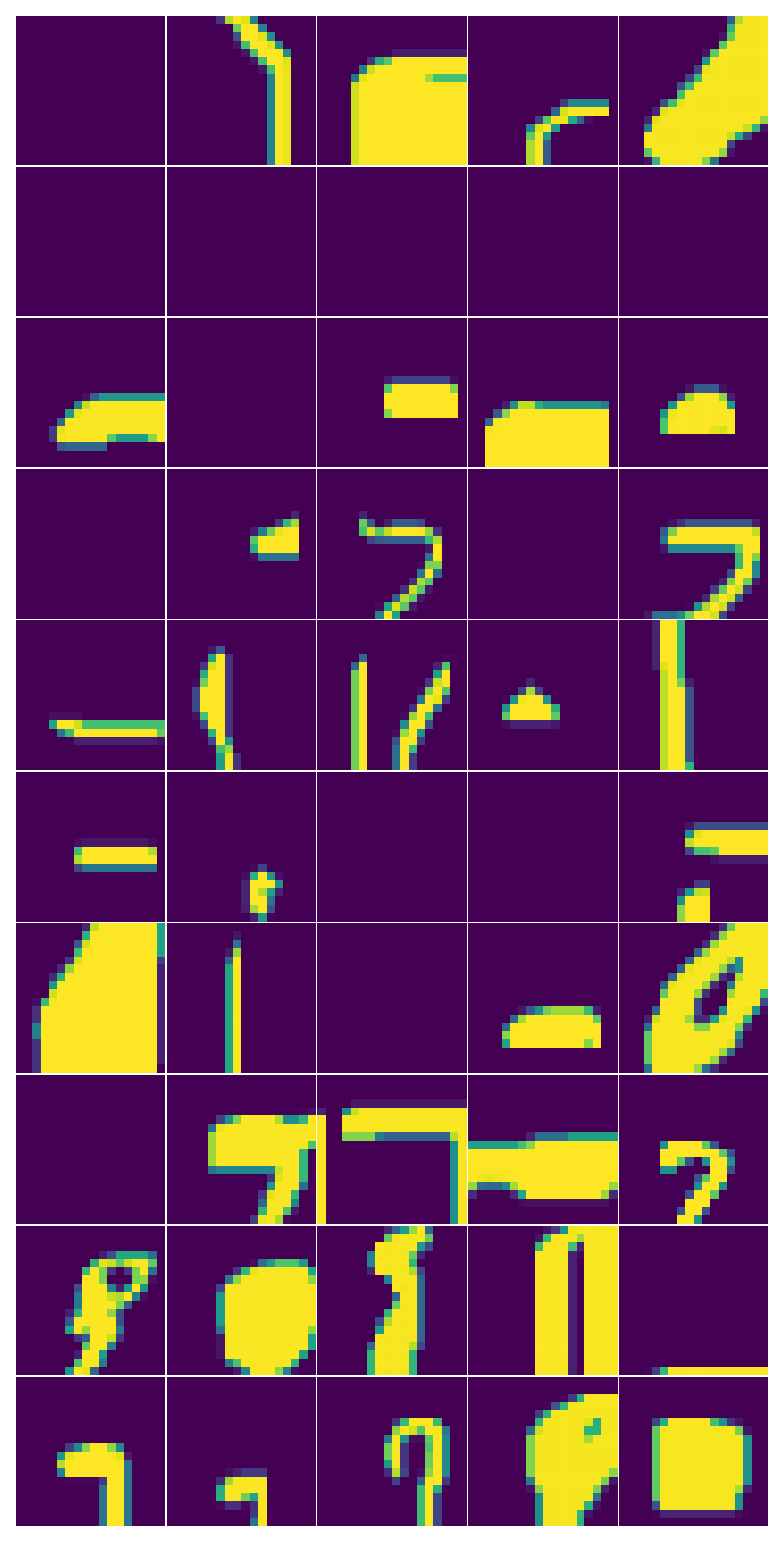}
        \includegraphics[width=0.16\textwidth]{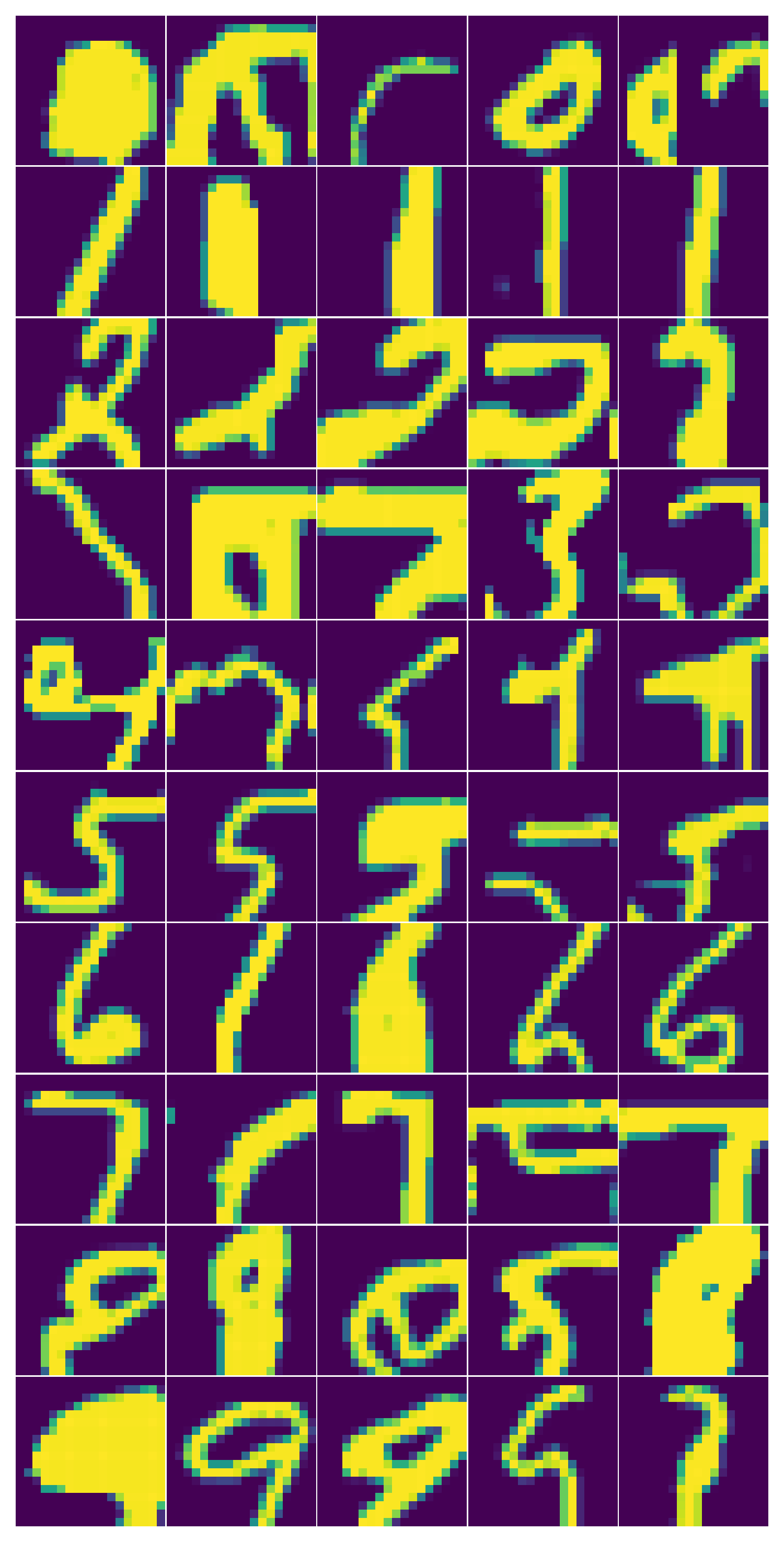}
        \includegraphics[width=0.16\textwidth]{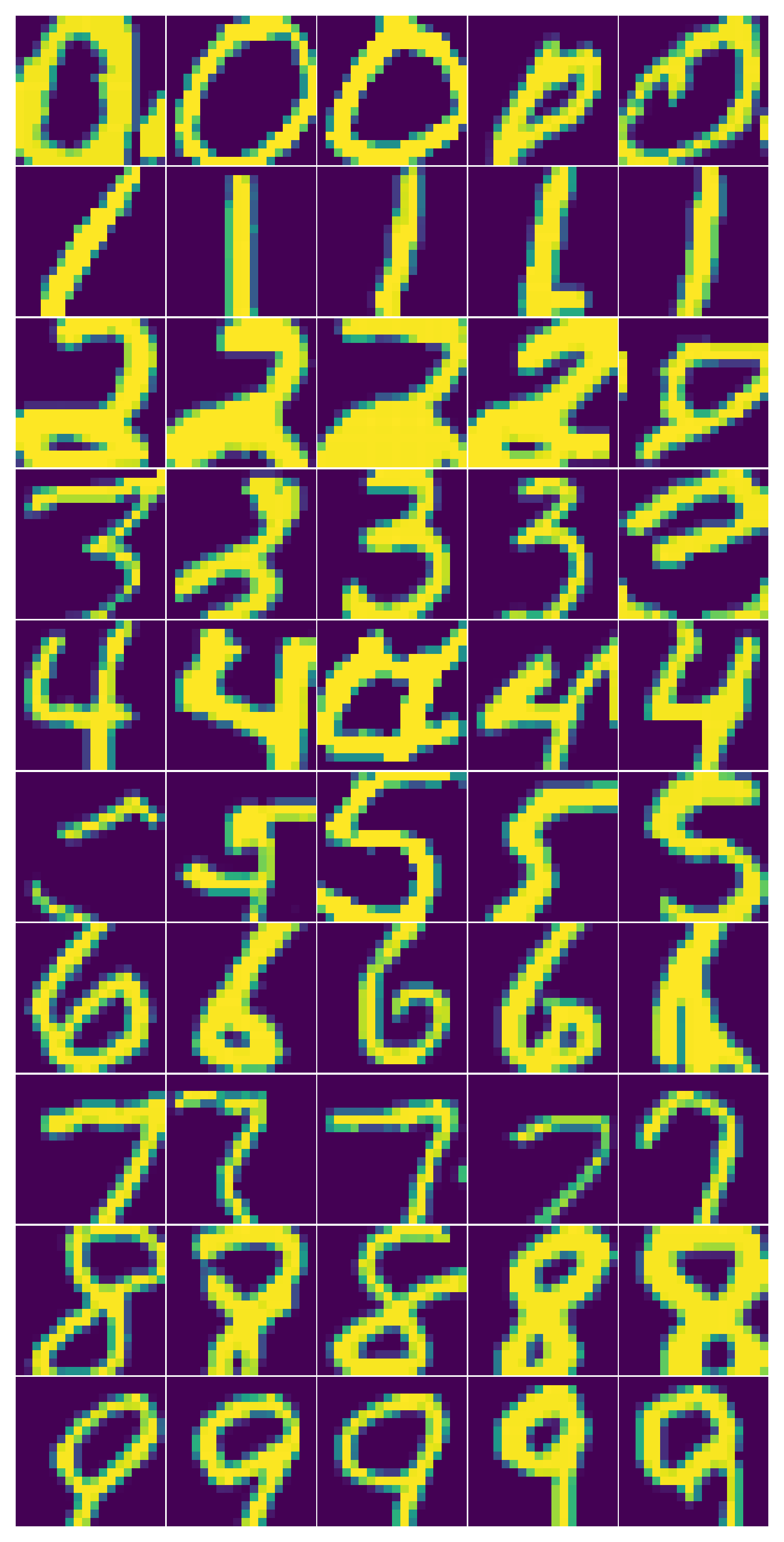}
    \caption{Images generated by \lift{}/\gptj{} given the digit number in the prompt.}
    \label{fig:mnist_gptj}
    \end{subfigure}
    \begin{subfigure}[a]{\textwidth}
        \includegraphics[width=0.16\textwidth]{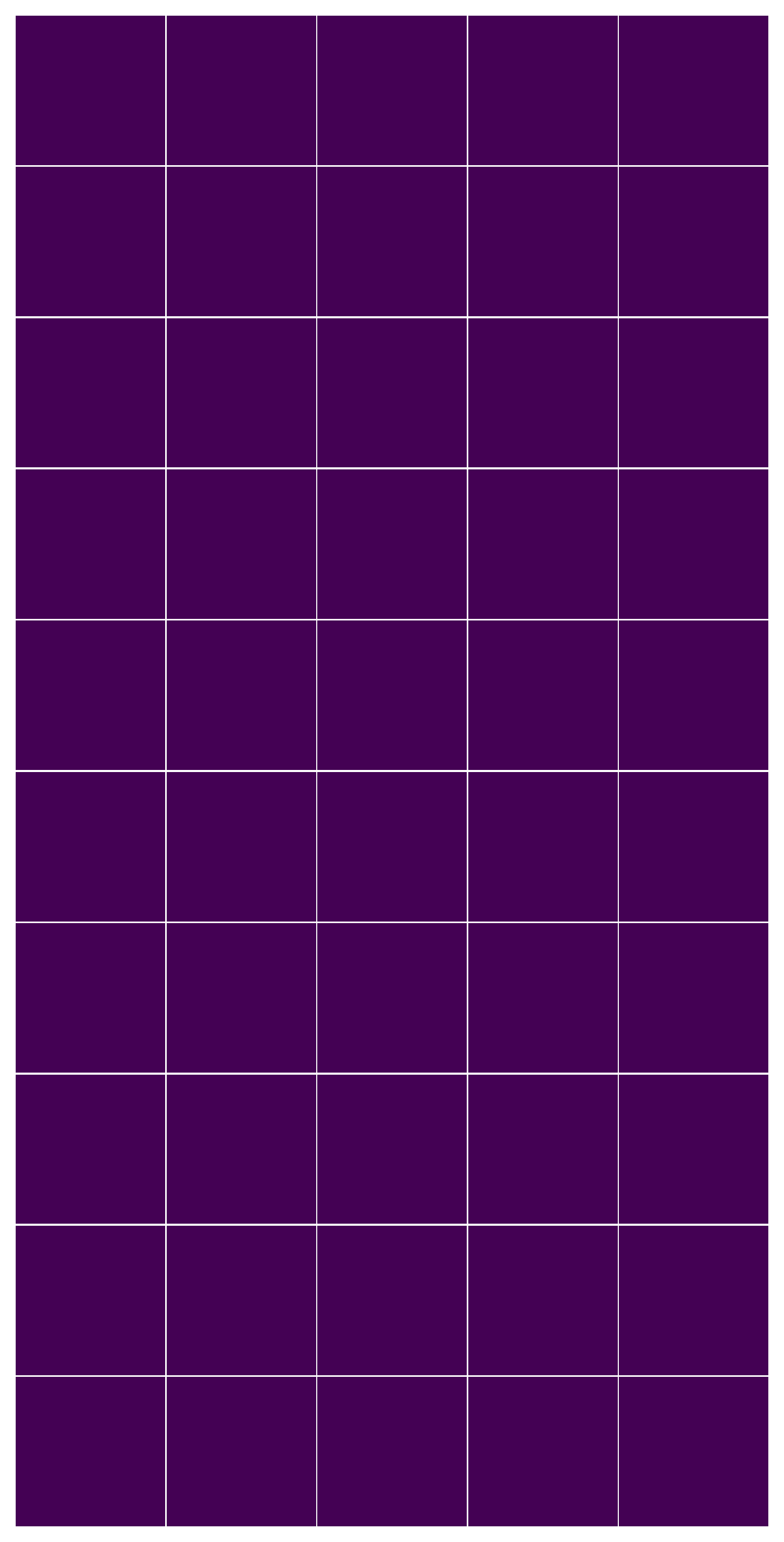}
        \includegraphics[width=0.16\textwidth]{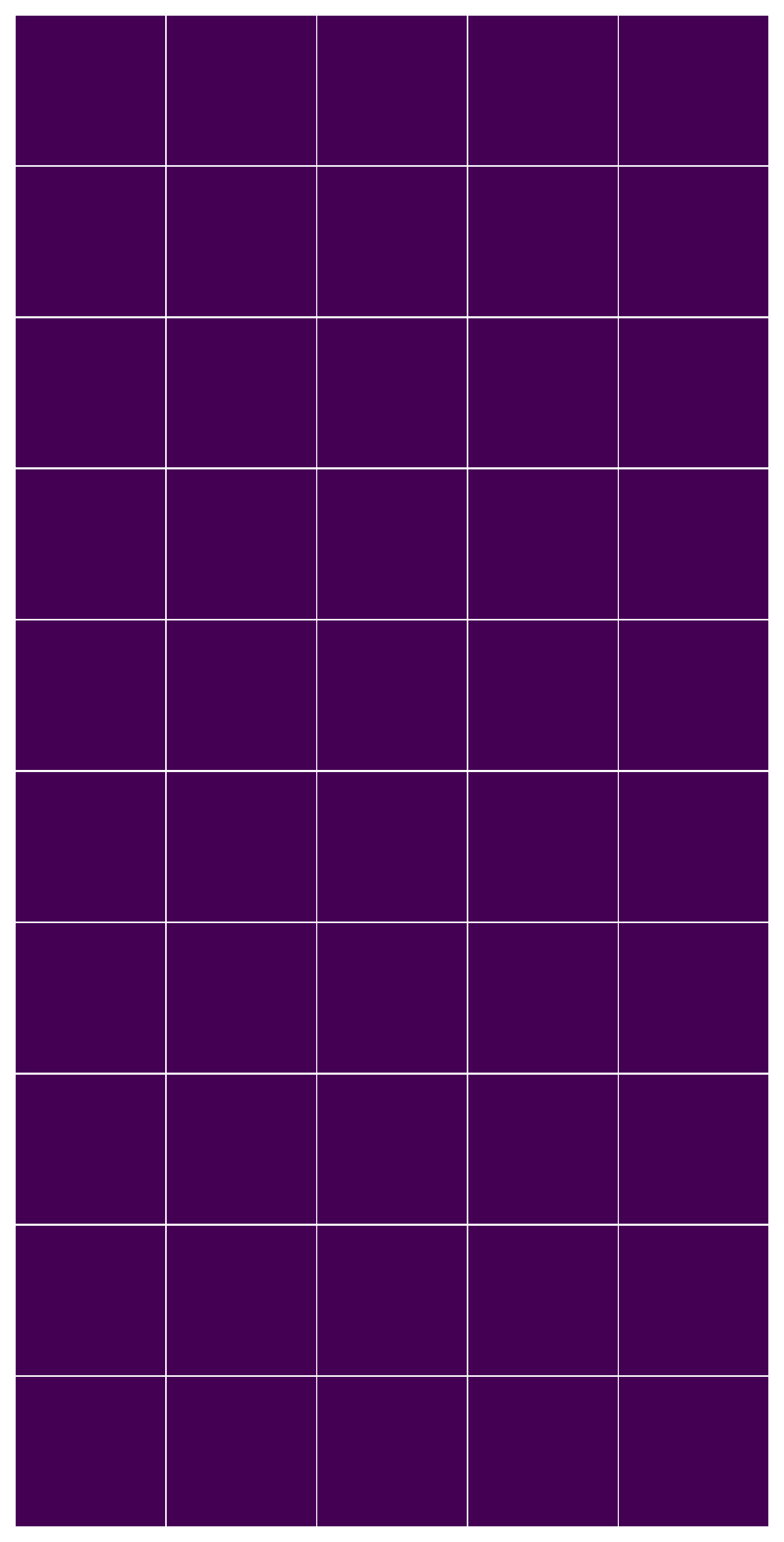}
        \includegraphics[width=0.16\textwidth]{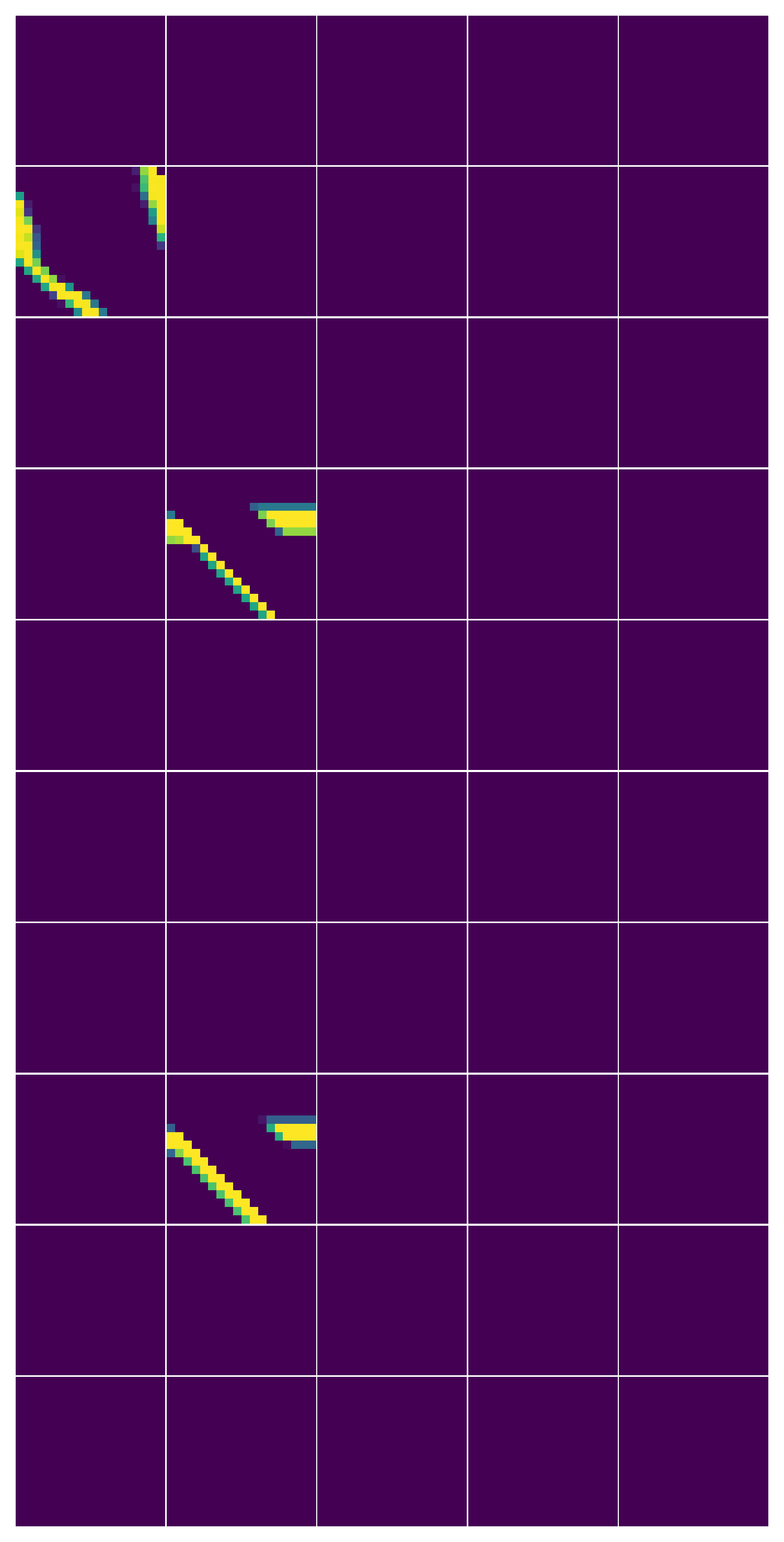}
        \includegraphics[width=0.16\textwidth]{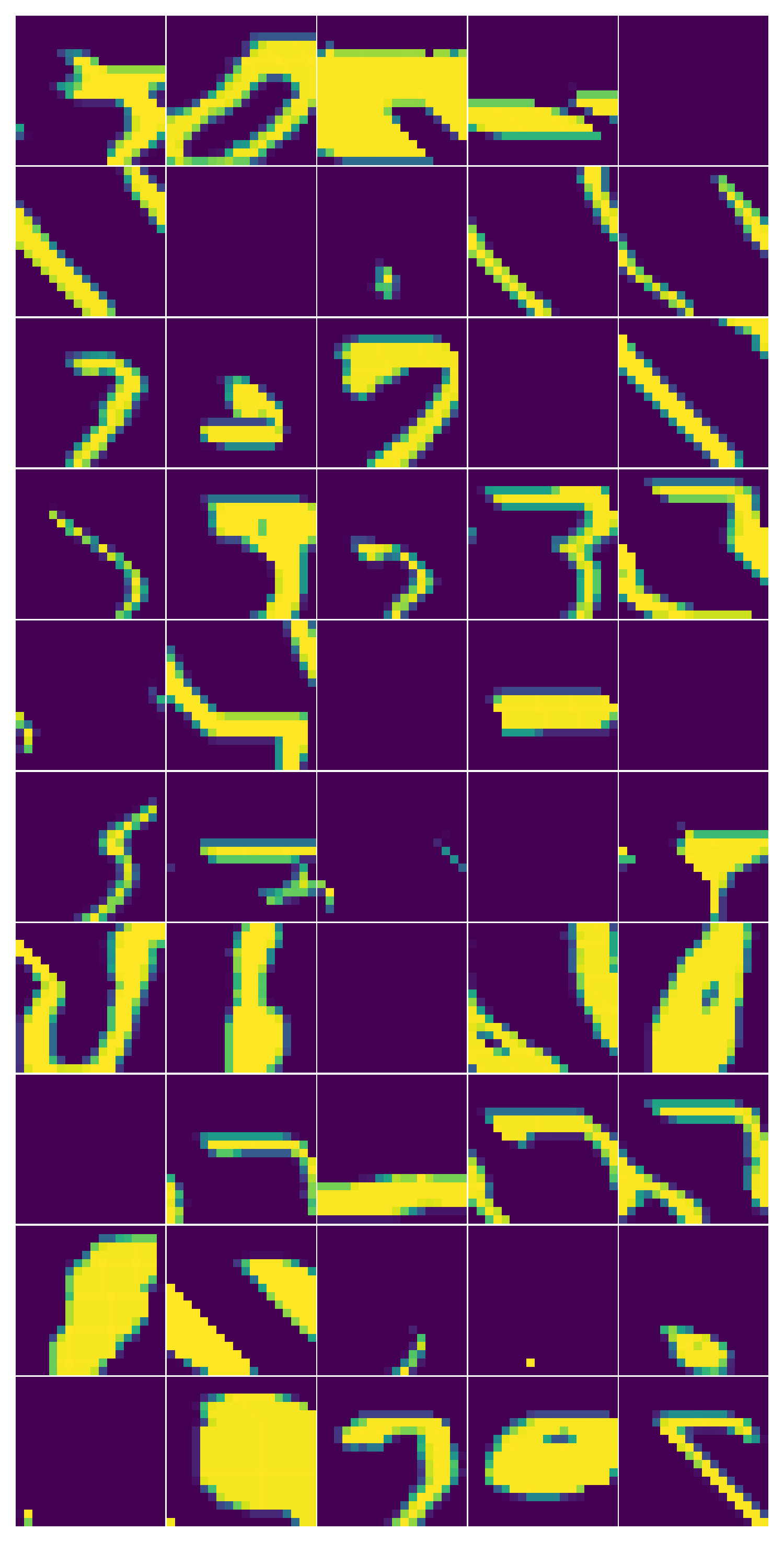}
        \includegraphics[width=0.16\textwidth]{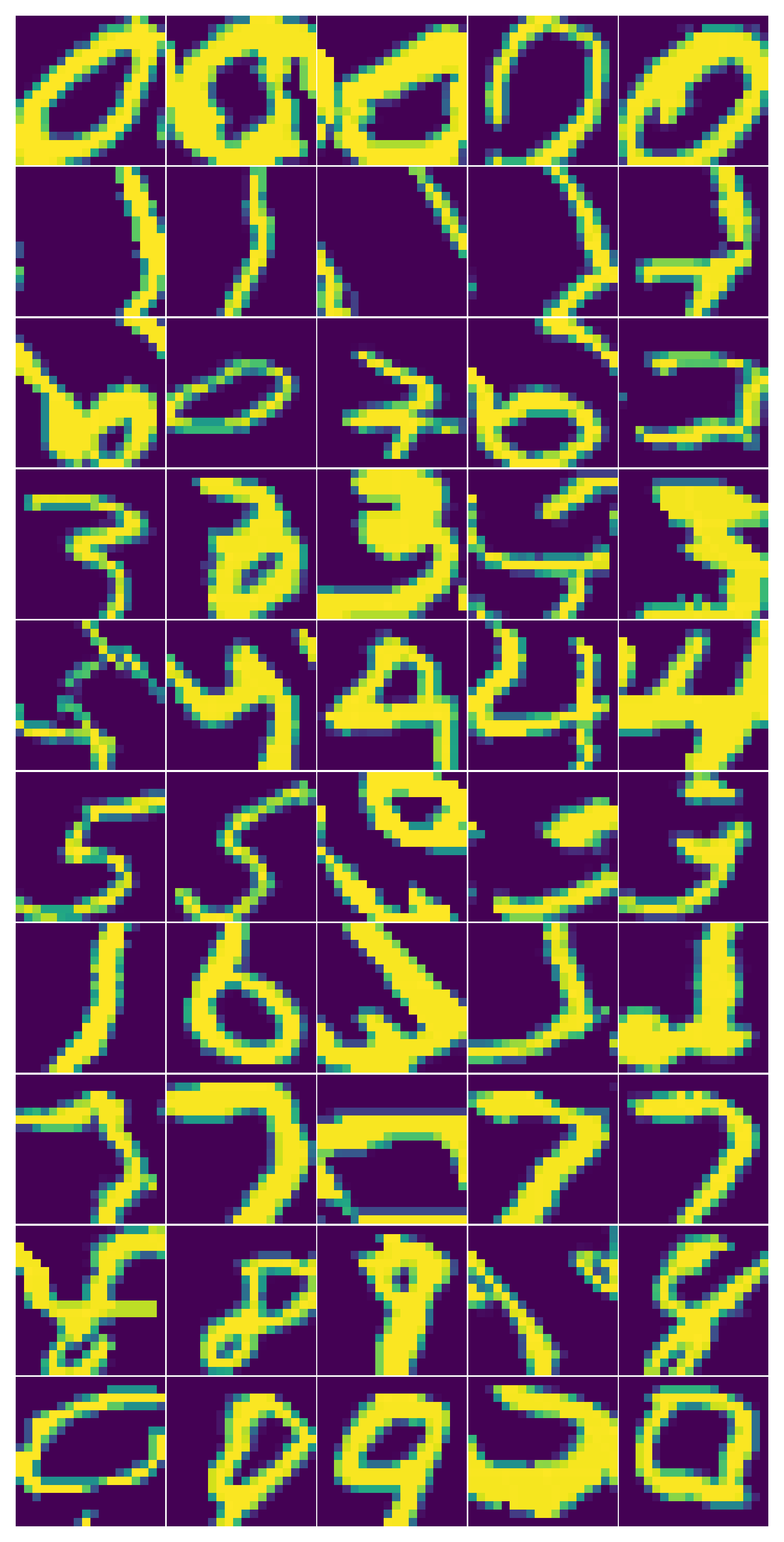}
        \includegraphics[width=0.16\textwidth]{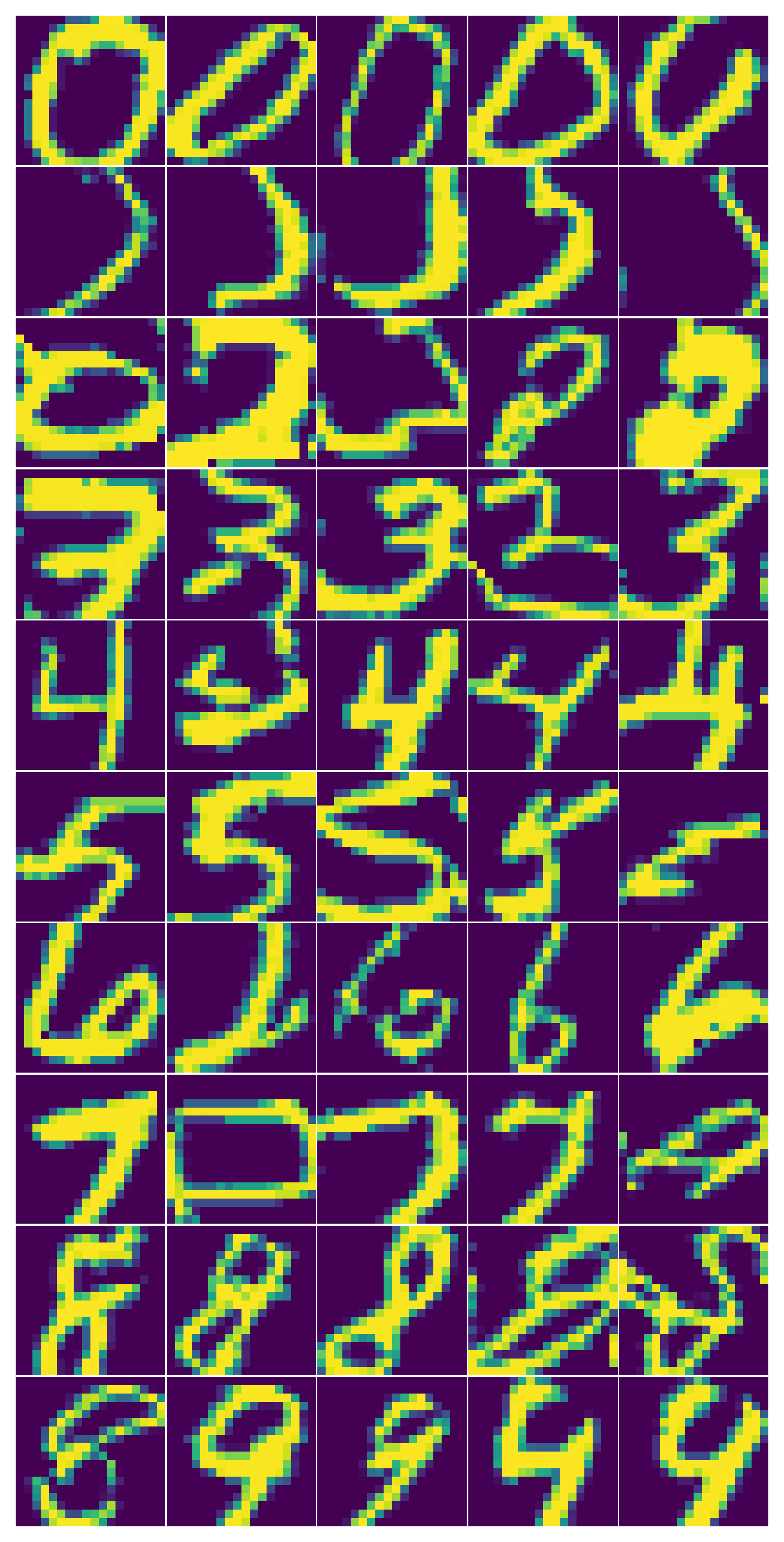}
    \caption{Images generated by \lift{}/\gptt{} given the digit number in the prompt.}
    \label{fig:mnist_gpt3}
    \end{subfigure}
    \begin{subfigure}[b]{\textwidth}
        \includegraphics[width=0.16\textwidth]{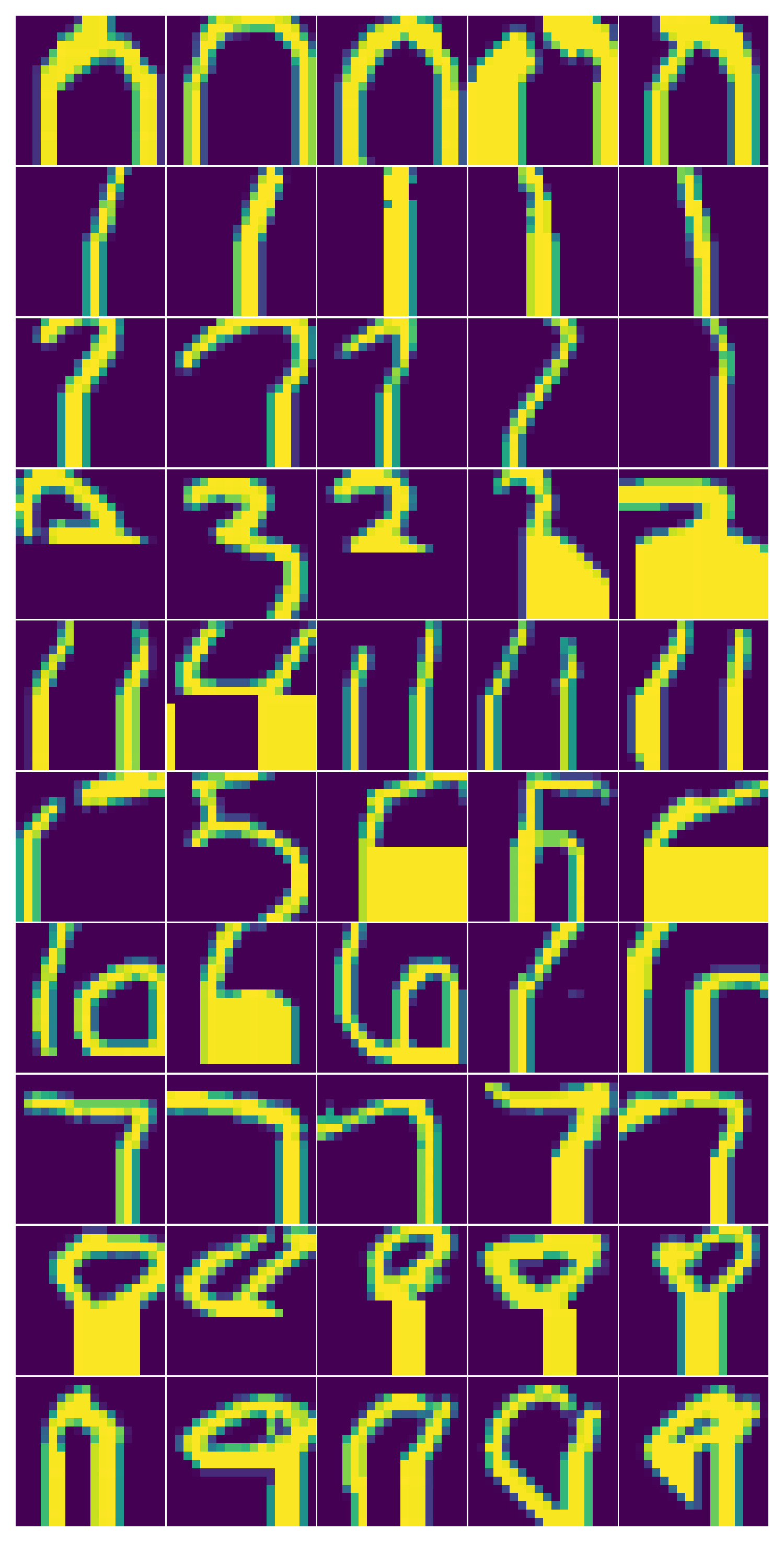}
        \includegraphics[width=0.16\textwidth]{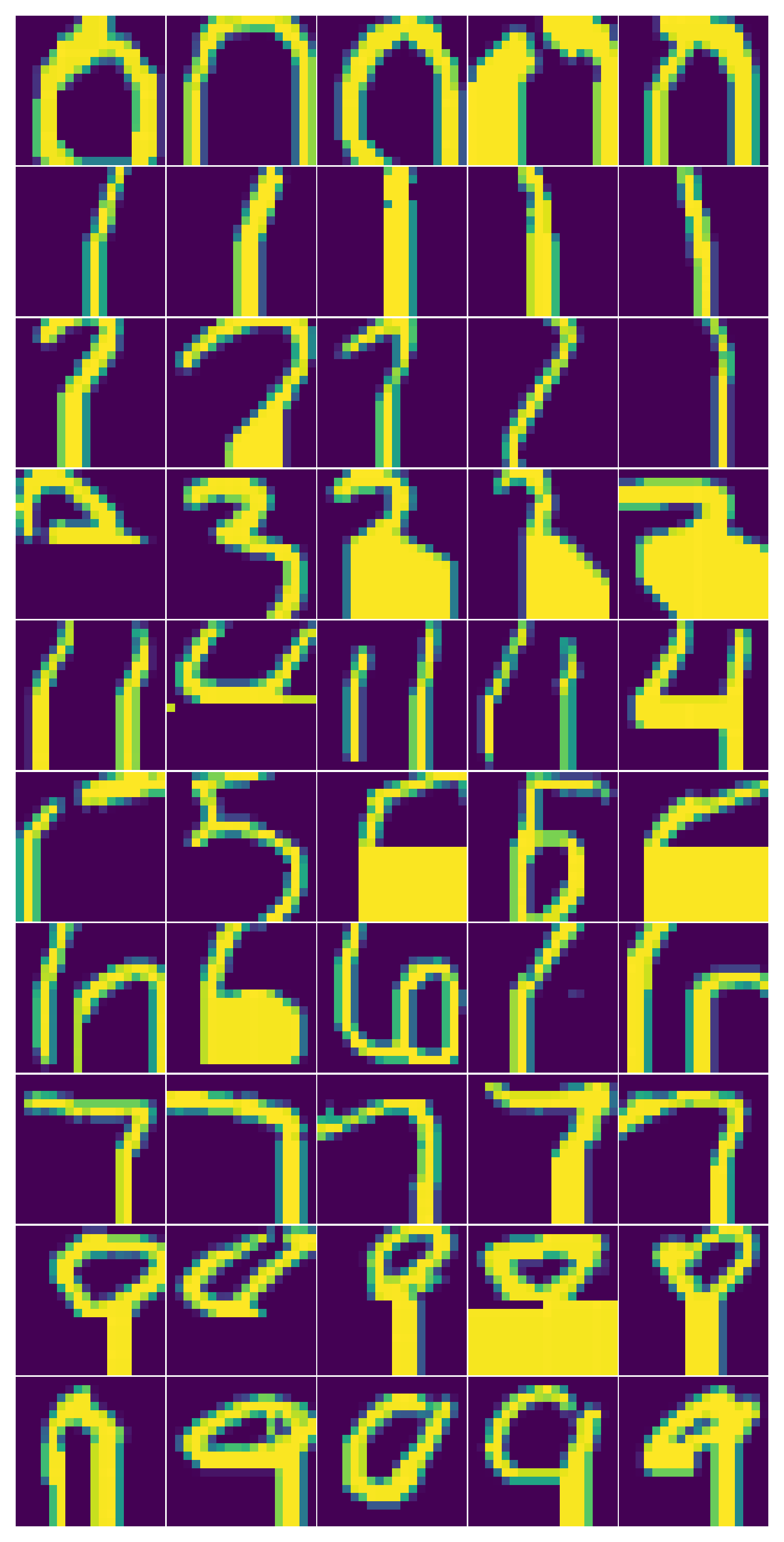}
        \includegraphics[width=0.16\textwidth]{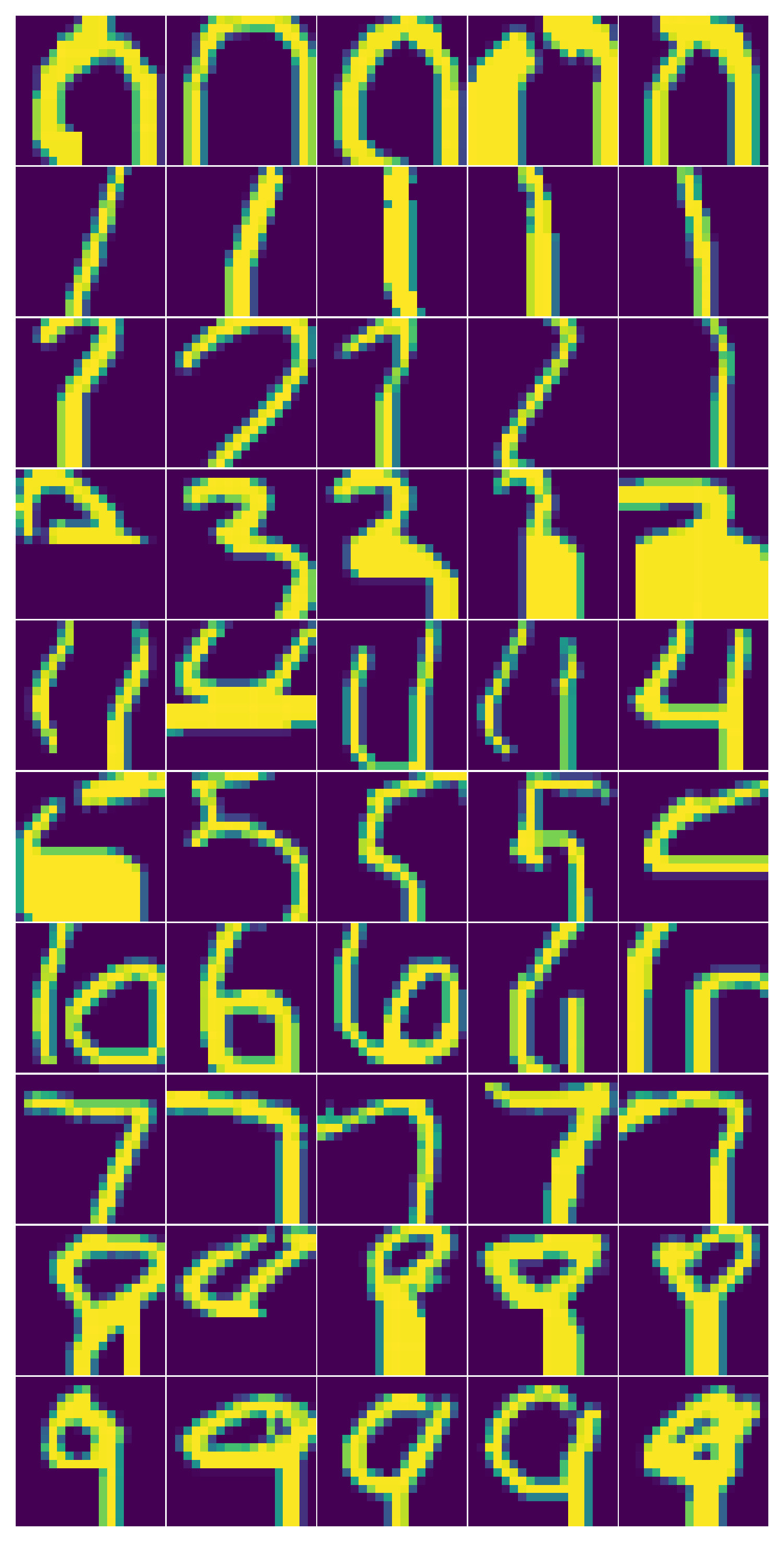}
        \includegraphics[width=0.16\textwidth]{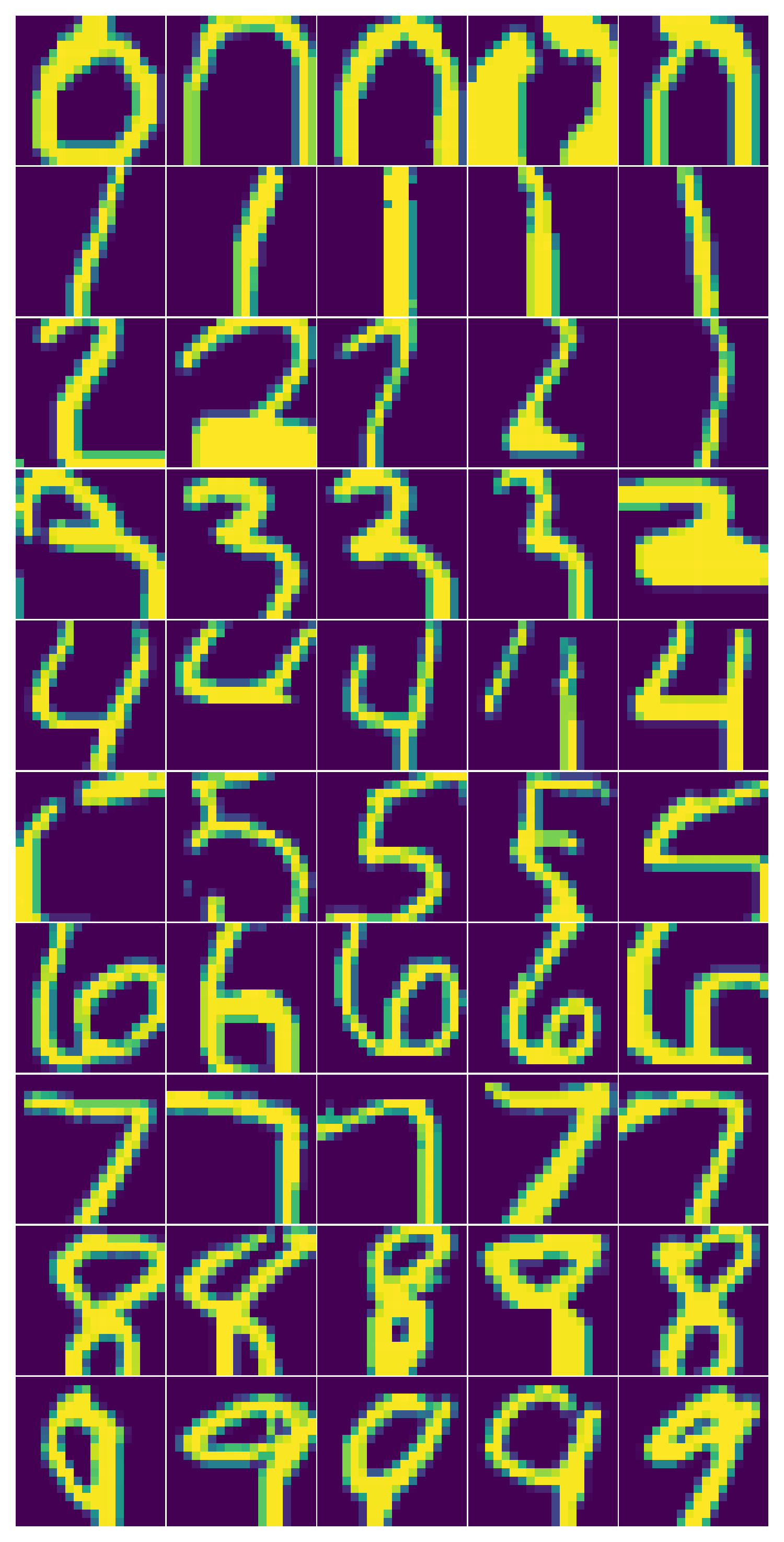}
        \includegraphics[width=0.16\textwidth]{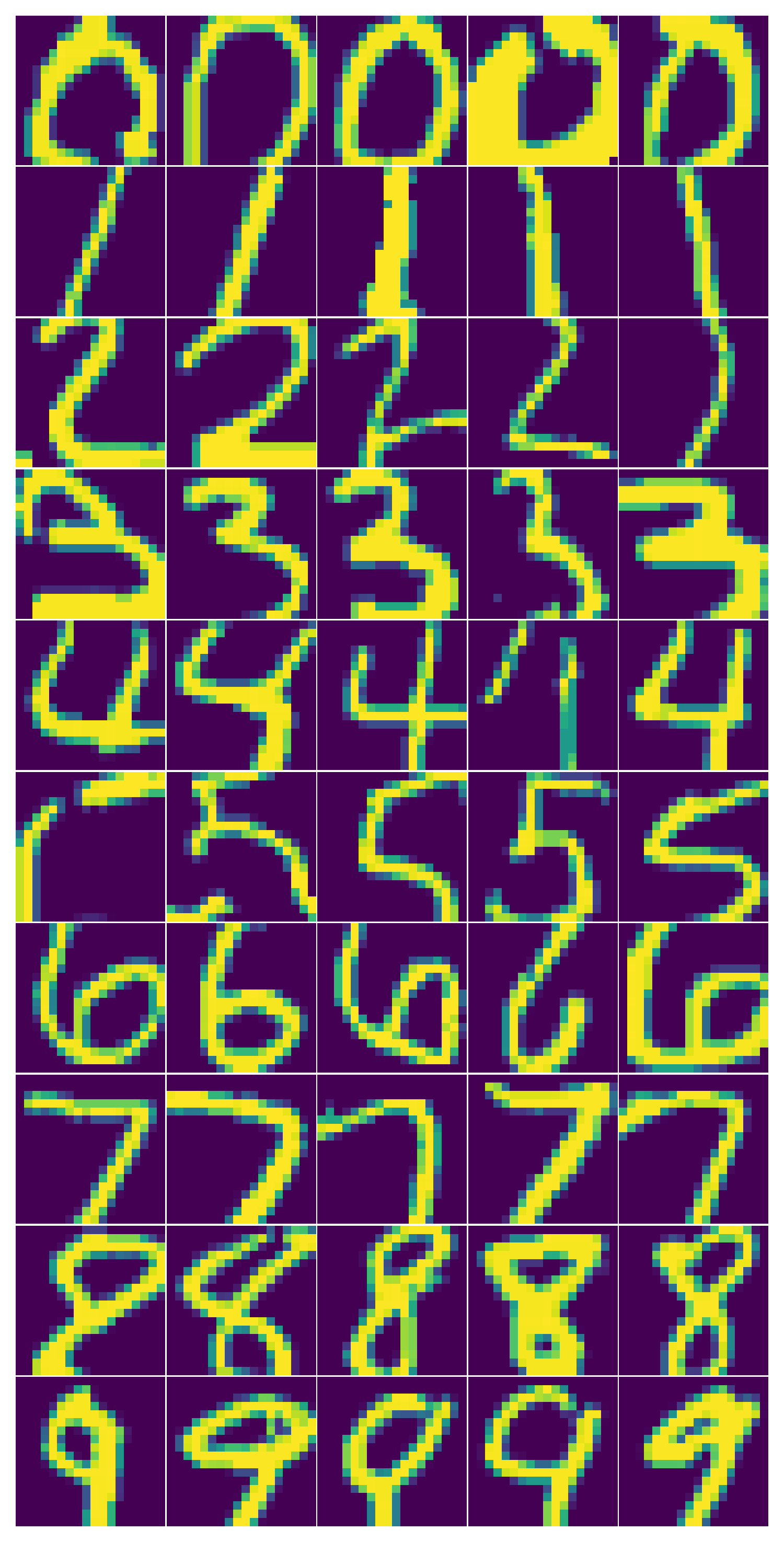}
        \includegraphics[width=0.16\textwidth]{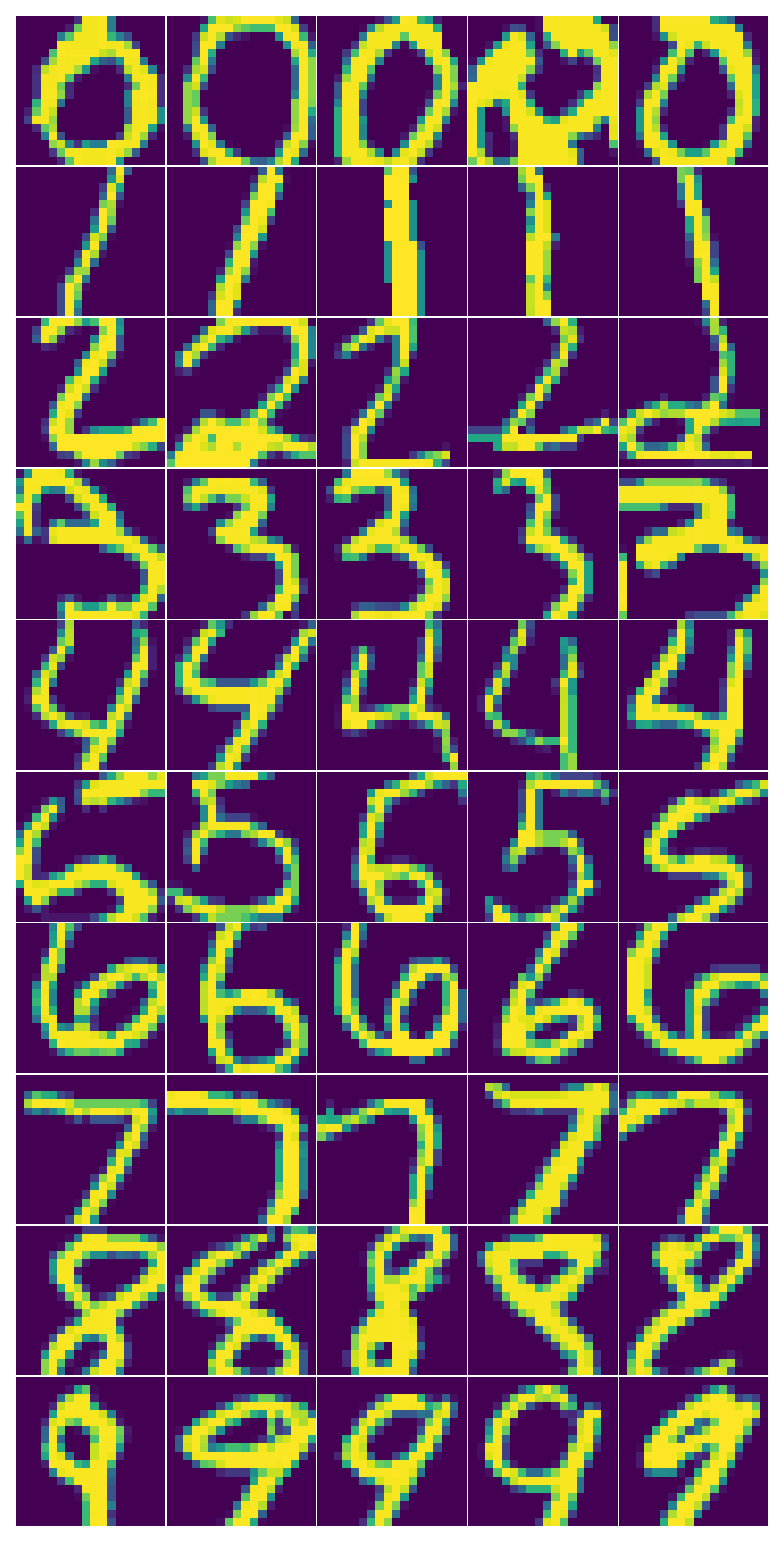}
    \caption{Images generated by \lift{}/\gptj{} given the digit number and top half pixels of the pictures with the corresponding digit number in the prompt.}
    \label{fig:mnist_gptj_half}
    \end{subfigure}
    \begin{subfigure}[d]{\textwidth}
        \includegraphics[width=0.16\textwidth]{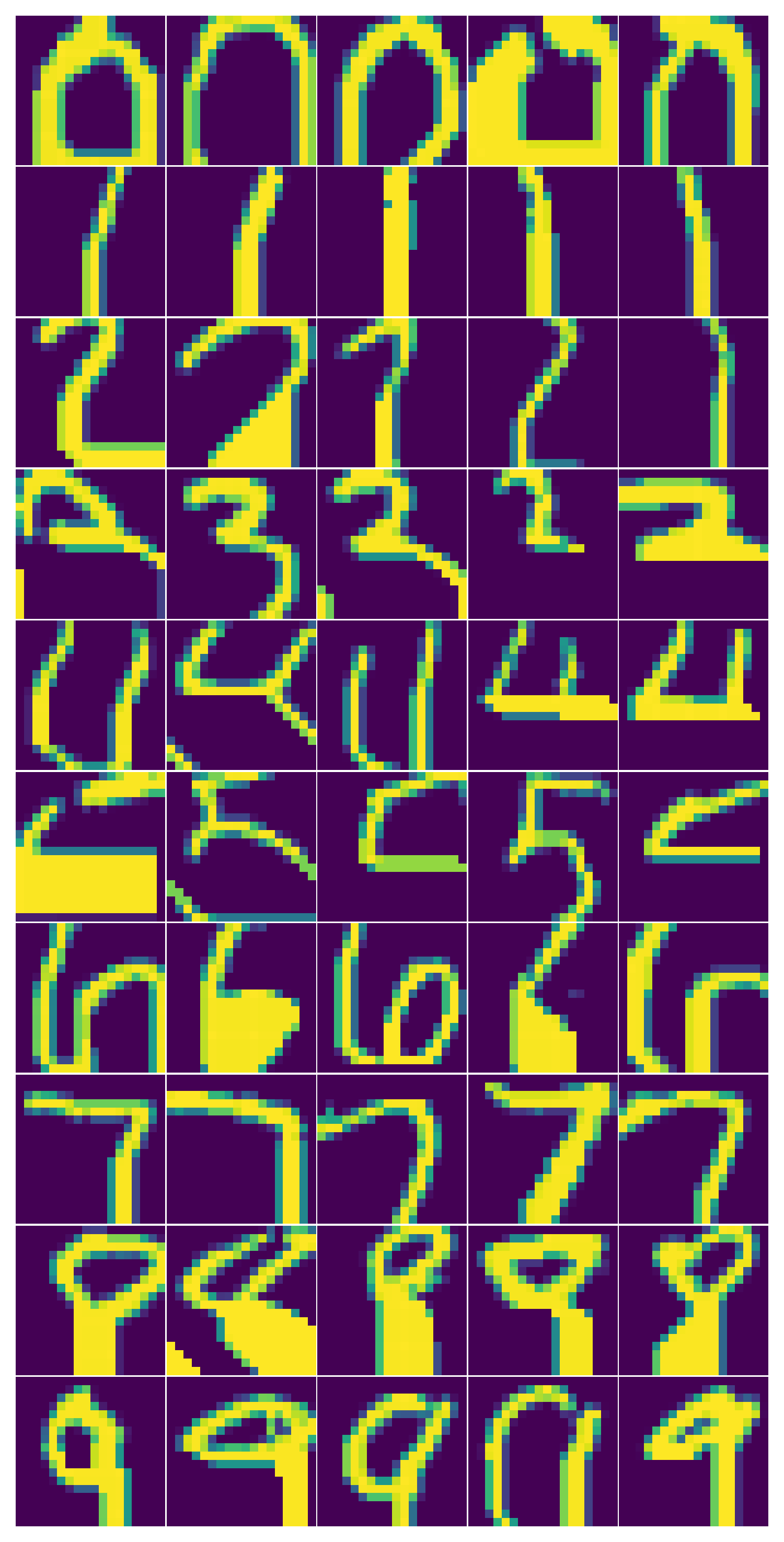}
        \includegraphics[width=0.16\textwidth]{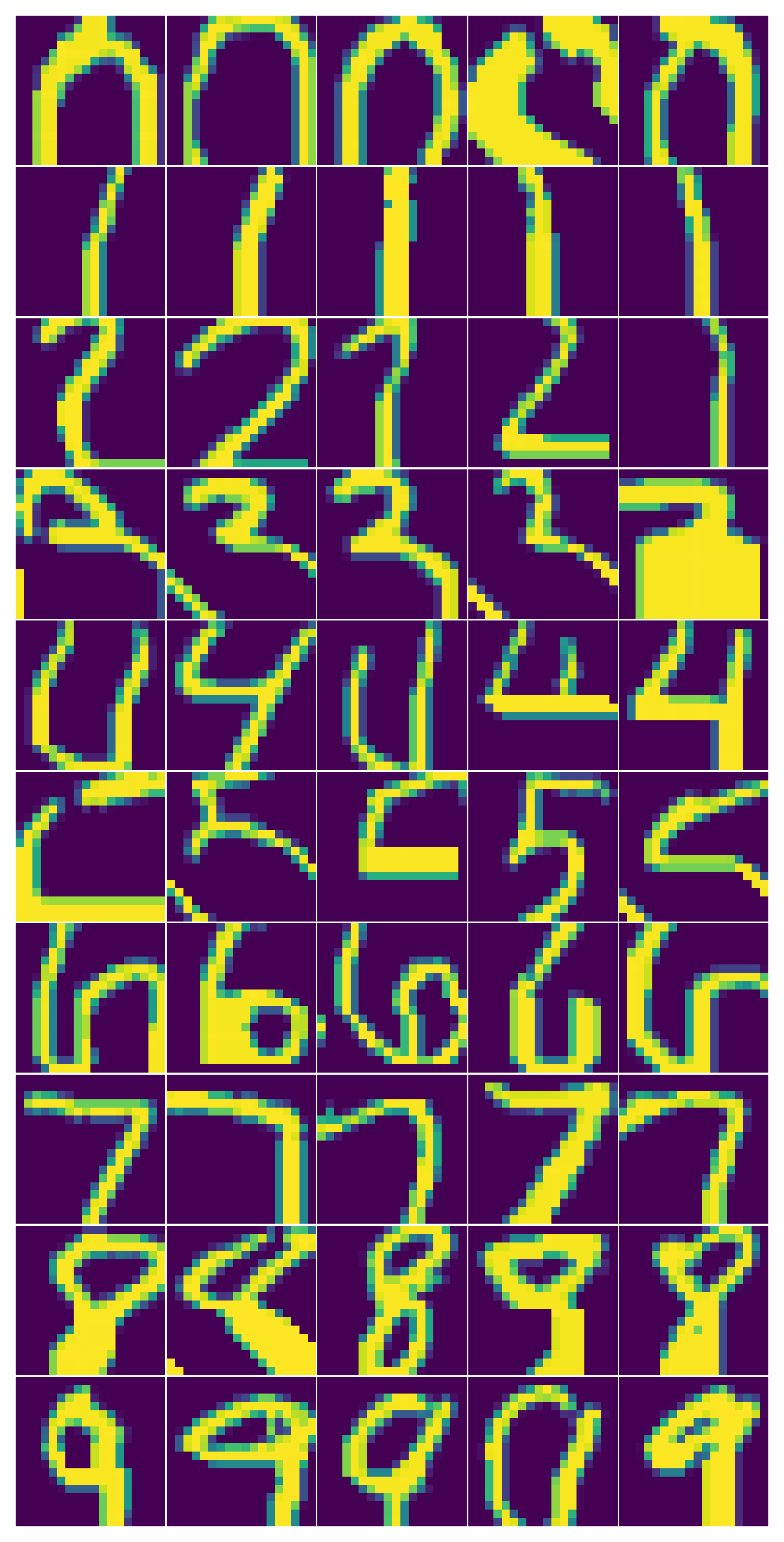}
        \includegraphics[width=0.16\textwidth]{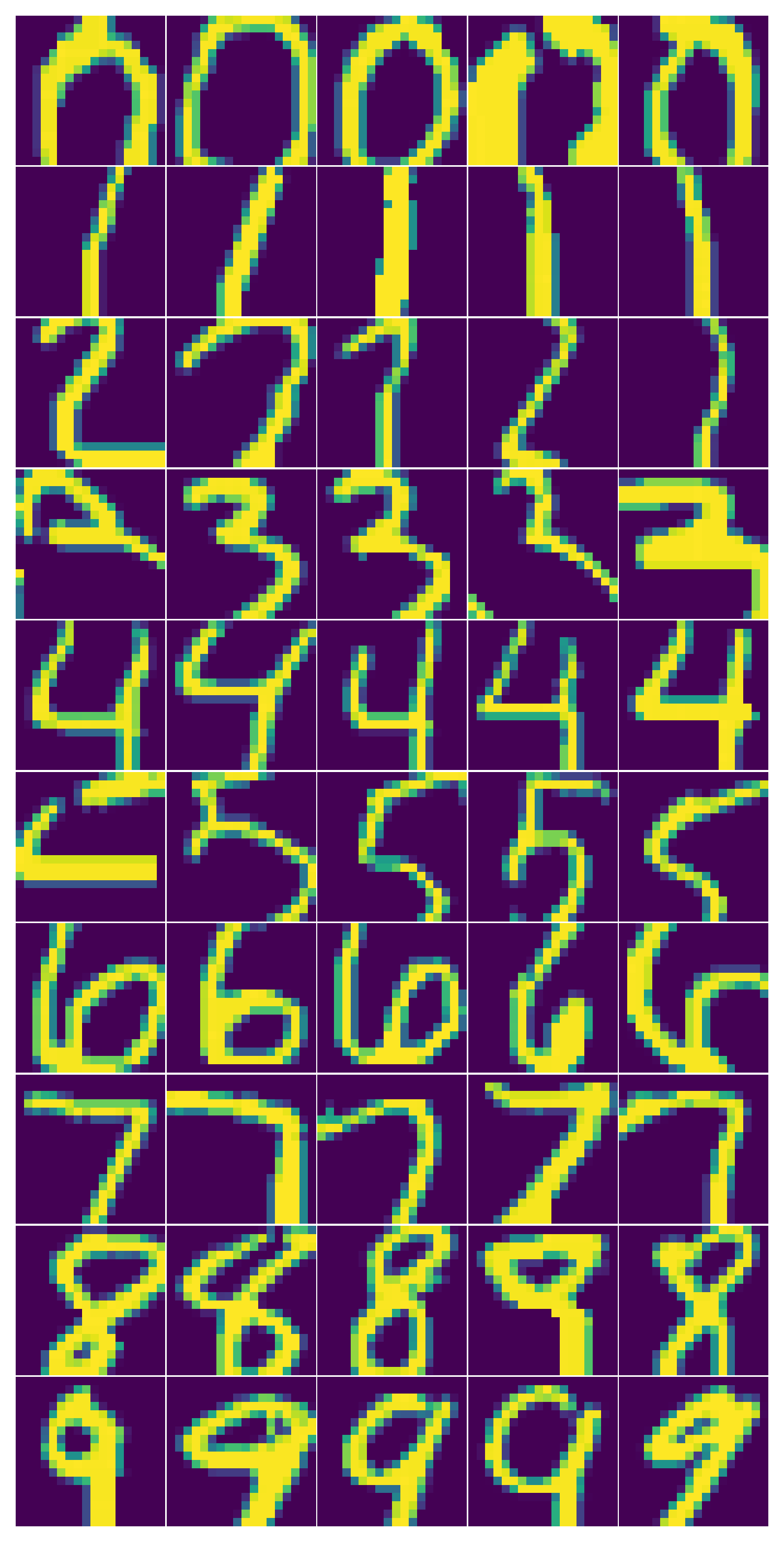}
        \includegraphics[width=0.16\textwidth]{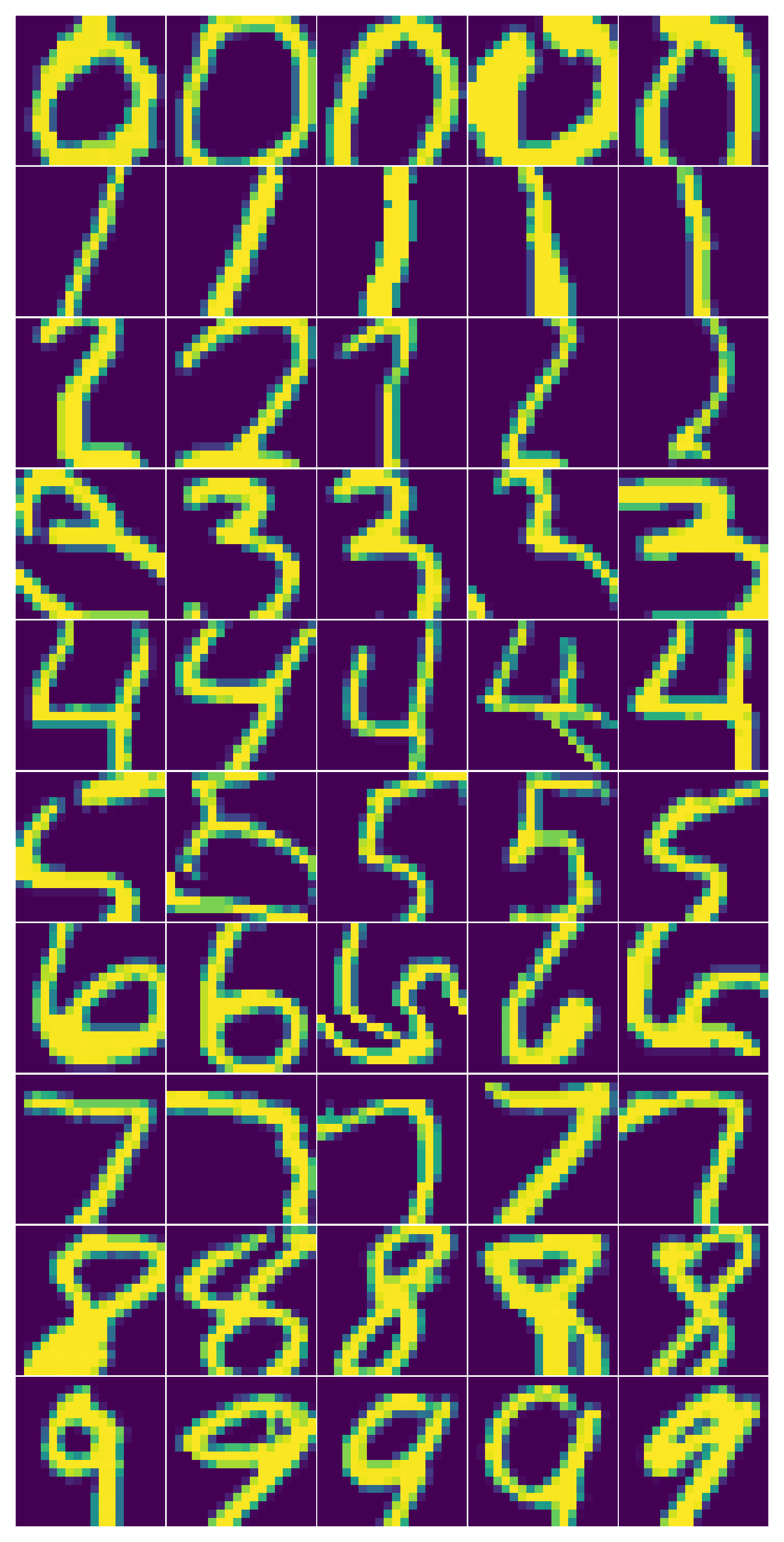}
        \includegraphics[width=0.16\textwidth]{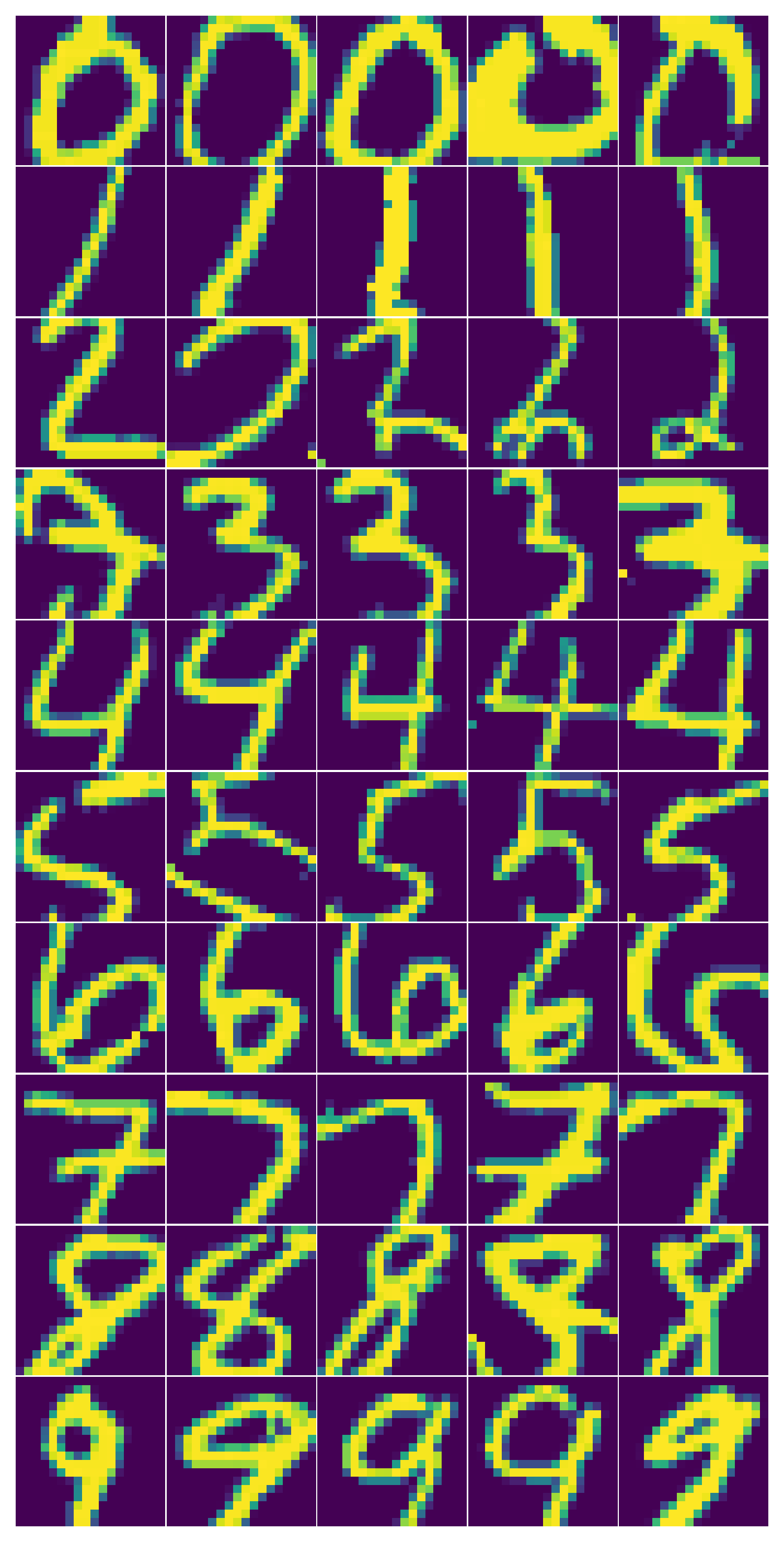}
        \includegraphics[width=0.16\textwidth]{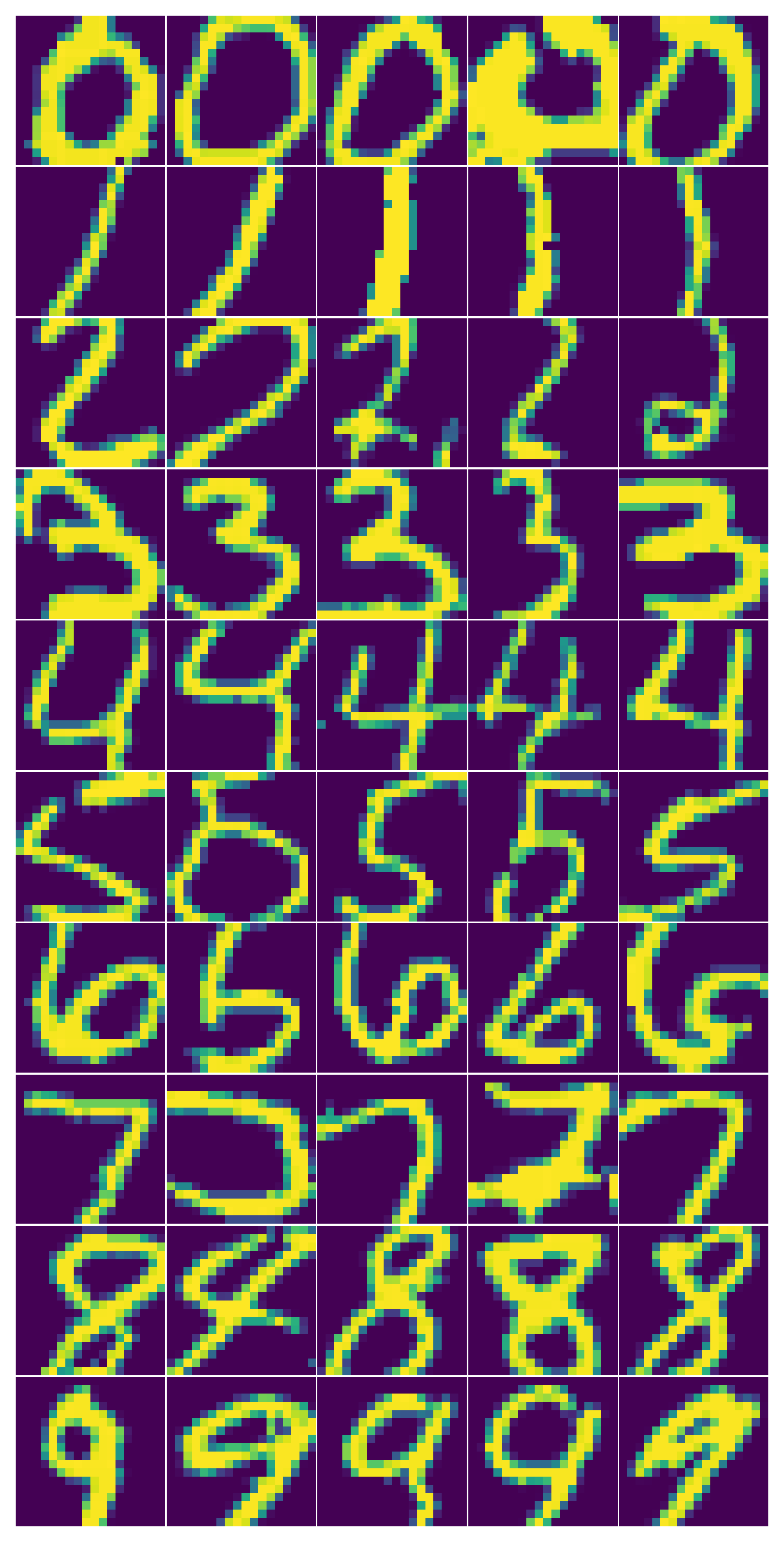}
    \caption{Images generated by \lift{}/\gptt{} given the digit number and top half pixels of the images with the corresponding digit number in the prompt.}
    \label{fig:mnist_gpt3_half}
    \end{subfigure}
    \vspace{-2mm}
\caption{
\textbf{Output of \lift{} as generative models.}
We apply \lift{} to generate new MNIST images. 
Each figure contains six subfigures, where each subfigure visualizes the output of \lift{} when different temperatures $\in [0, 0.3, 0.5, 0.7, 0.9, 1]$ of LMs are chosen. 
We generate five images for each digit by using \lift{} to make the prediction five times.
Task (i): when only the digit number is given, we observe that the \lift{} can generate reasonable images under high temperatures.
Task (ii): when both digit number and top half pixels are given, \lift{} can generate images of comparably high quality under different temperatures. 
}
    \label{fig:mnist_gpt_models}
\end{figure}

\paragraph{Results.}
Fig.~\ref{fig:mnist_gpt_models} visualizes the output of \lift{} in generating the images of different digits under different temperatures. 
We observe that \lift{} is able to generate reasonable images when both the digit number and top half pixels are given, as shown in Fig.~\ref{fig:mnist_gptj_half} and Fig.~\ref{fig:mnist_gpt3_half}.
In contrast, Fig.~\ref{fig:mnist_gptj} and Fig.~\ref{fig:mnist_gpt3} show that \lift{} is able to work only under high temperatures (\eg, \texttt{Temp = 0.9, 1}).
This might be because, under low temperatures, the model is likely to generate output associated with the highest probability. In contrast, a higher temperature gives chances to generate outputs associated with slightly lower probability, which introduces more variety and creativity. 
It has been observed in compositional text generation tasks such as completing stories, high temperature leads to better and more creative performance~\citep{roemmele2018automated}.

\begin{table}[h]
\vspace{4mm}
\caption{
\textbf{Efficacy of \lift{} as generative models.}
Perplexity ($\downarrow$) is a metric for measuring the probability of the sample produced by the model on a dataset. 
We report the average perplexity of \lift{} trained for generating MNIST images.
Note that the difference between the average test perplexity and average training perplexity is small, implying the good generalizability of \lift{} as generative models. 
}
\vspace{2mm}
    \centering
        \begin{tabular}{c|cc}
            \toprule[.2mm]
            & \textbf{\lift{}/\gptj{}} & \textbf{\lift{}/\gptt{}} \\
            \midrule
            Avg Training Perplexity ($\downarrow$) & 3.56 $\pm$ 1.42 & 3.58 $\pm$ 1.46 \\
            Avg Test Perplexity ($\downarrow$) & 3.57 $\pm$ 1.44 & 3.62 $\pm$ 1.51 \\
            \bottomrule[.2mm]
        \end{tabular}
    \vspace{1mm}
\label{tab:perplexity}
\end{table}

\subsection{Results for Improving Techniques of \lift{} (Section~\ref{sec:lift_improve})}
\label{app:lift_improve}

\subsubsection{Two-Stage Intermediate Fine-Tuning for \lift{}}\label{app:two_stage}
We provide an additional evaluation of the two-stage intermediate fine-tuning with \lift{}\gptj{} on four more datasets.
For any given dataset, we first generate two pretext tasks with simple synthetic Gaussian samples (discussed in \ref{app:setup_datatset}).
We fine-tune the \gpt{} with pretext tasks for a few (2--3) epochs, then fine-tune the newly fine-tuned \gpt{} with the target (given) dataset.
Here, due to the black-box API of \gptt{}, we currently can neither keep the order of samples unchanged (pretext, target) during the fine-tuning stage nor  fine-tune the model twice.
Hence, we only use \gptj{} in this experiment.

\begin{figure}[!htbp]
    \centering
    \includegraphics[width=0.9\textwidth]{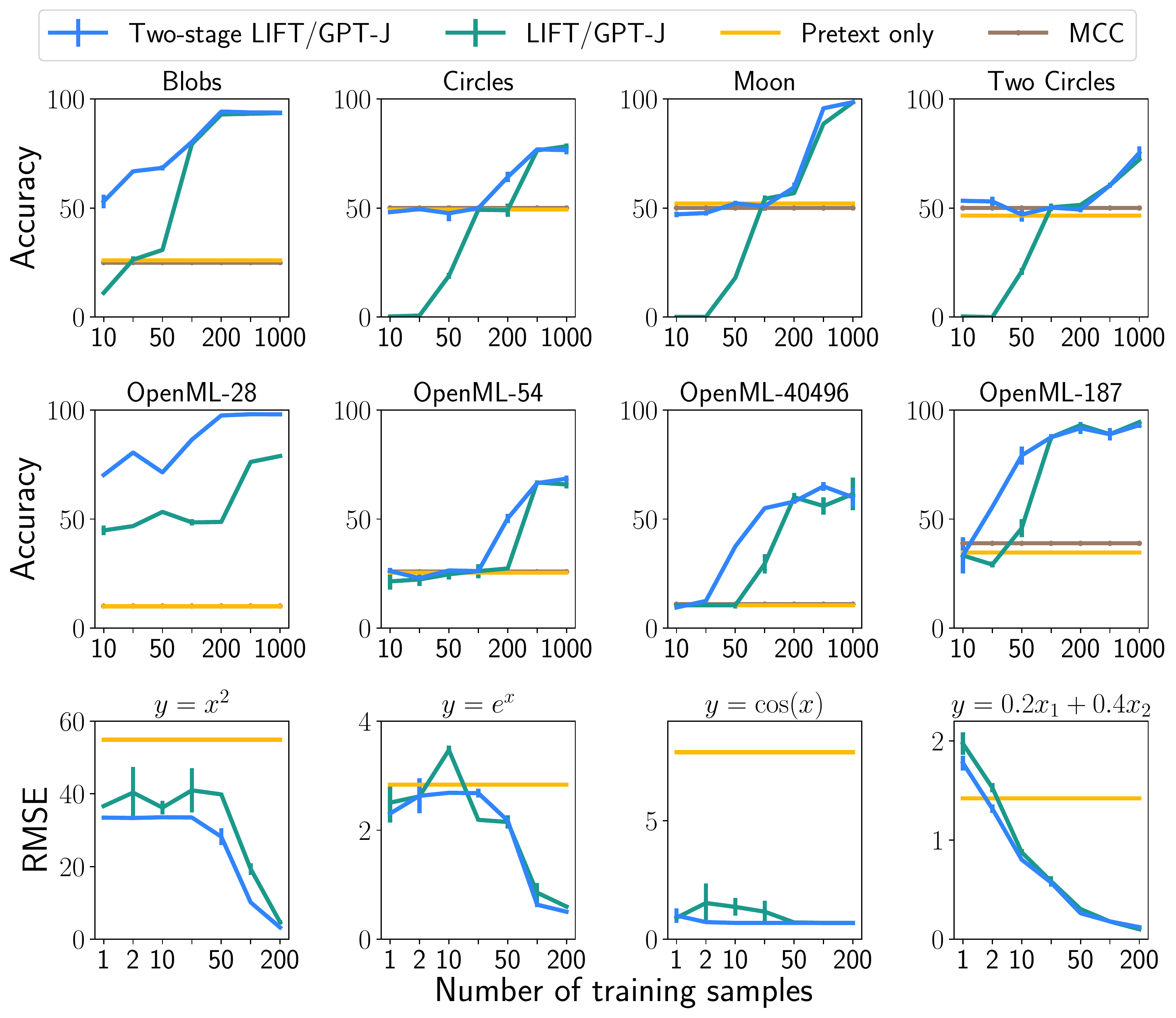}
    \vspace{-2mm}
    \caption{
    \textbf{Two-stage fine-tuning with \lift{}/\gptj{}}.
    We apply \lift{} in two consecutive stages: first on synthetic pretext data, then on the target data.
    We evaluate on classification tasks (with accuracy) and a regression task (RMSE error).
    The two-stage fine-tuning \lift{}(blue) outperforms the original fine-tuning \lift{} (green) when the number of training samples is small, across all tasks.
    }
    \label{fig:app_pretext}
\end{figure}

Fig.~\ref{fig:app_pretext} presents results from eight datasets, including a regression task, three OpenML tasks, and other synthetic tasks.
We see that two-stage intermediate \lift{} helps to improve the original fine-tuning, especially when the number of training samples is small.
Its effect is more clearly shown in synthetic classification datasets (\texttt{Blobs}, \texttt{Circles}, \texttt{Moon}, and \texttt{Two Circles}).
We also observe that besides the number of features and number of classes, the pretexts do not need to represent any other characteristics of the target dataset, such as the linear/non-linear correlation or the relevance of features.
This makes it simpler to generate the pretexts.

\newpage

\section{Additional Experiments and Findings (NOT Discussed in the Main Paper)}
\label{app:additional_findings}
Here, we provide results of additional experiments that have not been discussed in the main manuscript.
To be more specific,  we study 
the effect of replacing the input or output layers (\ref{sec:fpt}), effect of large LMs (\ref{sec:different_gpt}), quantitative classification evaluation on neural-net-based synthetic datasets (\ref{app:boundary_complexity}), and the ability of \lift{} performing the ridge regression (\ref{app:ridge}).
We also provide the visualization of \lift{}'s training curve in Section~\ref{app:training_curves}.

\subsection{What Is the Effect of Replacing the Input or Output Layers?}\label{sec:fpt}
In this experiment, we assess the performance of transformer fine-tuned with replaced input$\slash$output layer, following the methods used in Frozen Pretrained Transformers (FPT)~\citep{lu2021pretrained}.
We consider vanilla FPT and its two variants.
Specifically, for vanilla FPT, we reinitialized a trainable input layer and a trainable output layer, with frozen pretrained GPT-J transformer architectures in the middle. 
As in \cite{lu2021pretrained}, the input dimension equals to the number of features and the output dimension equals to the number of classes, which varies depending on the tasks.
Table~\ref{tab:fpt} compares the result of our method \lift{}/\gptj{}, FPT and  
the two variants of FPT: (i) FPT (Output Only) which only replaces the output layer, and (ii) FPT (Input Only), which only replaces the input layer.
We observe that both FPT and FPT (Output Only) perform slightly better than \lift{} on almost all tested cases, while FPT (Input Only) performs the worst.
Our justification for this observation is that training an output layer is similar to training a linear classifier, which might be easier than training an input layer as an encoder.

\begin{table}[!htbp]
   \caption{
   \textbf{Accuracies$(\uparrow)$ of \lift{}, Frozen Pretrained Transformer (FPT)~\citep{lu2021pretrained} and its two variants — FPT (Output Only) and FPT (Input Only).} 
   We employ \gptj{} in this experiment.
   We observe that replacing the output layer can slightly improve the performance of \lift{}, while only replacing the input layer performs the worst. 
   }
  \centering
  \small
    \setlength{\tabcolsep}{3pt}
    \begin{tabular}{l|c|ccc}
    \toprule[.4mm]
    Dataset (ID) & \textbf{\lift{}/\gptj{}} &  \textbf{FPT/\gptj{}} &  \textbf{FPT (Output Only)/\gptj{}} &  \textbf{FPT (Input Only)/\gptj{}} \\ \midrule
    Blobs (2) & 96.17$\pm$0.59&96.75$\pm$0.00&96.67$\pm$0.12&96.75$\pm$0.00 \\
Two Circles (6)&75.92$\pm$1.65&74.33$\pm$0.31&76.33$\pm$2.49&69.83$\pm$1.31\\
Iris (61)&96.67$\pm$0.00&96.67$\pm$0.00&97.78$\pm$1.57&81.11$\pm$3.14\\
Customers (1511)&85.23$\pm$1.61&87.88$\pm$0.54&88.26$\pm$0.54&86.74$\pm$1.42\\
Wine (187)&93.52$\pm$1.31&100.00$\pm$0.00&99.07$\pm$1.31&92.59$\pm$3.46\\
LED (40496)&65.33$\pm$0.47&73.00$\pm$2.94&71.67$\pm$1.25&68.67$\pm$1.89\\
    \bottomrule[.4mm]
    \end{tabular}
\vspace{1mm}
  \label{tab:fpt}
\end{table}

\subsection{Does \lift{} Benefit from Larger LMs?}
\label{sec:different_gpt}
\begin{table}[!htb]
  \caption{
  \textbf{The effects of larger LMs under different classification settings}.
  Recall that our previous results on \gptt{} are based on the smallest model \texttt{Ada}. 
  Here we use larger \gptt{} versions (\texttt{Babbage, Curie, Davinci}) as the pretrained LMs in our framework and evaluate the classification accuracy ($\uparrow$) of them in three settings: classification \textit{without} feature name, classification \textit{with} feature name and in-context classification. 
  For the setting of classification \textit{with} feature names, we incorporate names of features (columns) into the input prompts (see more details in Sec.~\ref{sec:feature_names}). 
  For in-context learning, the OpenML dataset ID and number of prompts are written together at each column, \textit{e.g.}, \texttt{TAE} (48)/50 means that we run experiments on the OpenML dataset \texttt{TAE} having ID 48, by using 50 input prompts.
  For the first two settings when \lift{} is applied, larger \lift{}/\gptt{} models (\texttt{Babbage, Curie, Davinci}) perform better than the smaller models \lift{}/\gptt{}-\texttt{Ada} and \lift{}/\gptj{}, but the performance gains are not always consistent and significant with model sizes. 
  For the in-context classification (\lift{} is not used), we observe more consistent improvement by using larger models.}
  \centering
\begin{adjustbox}{width=\textwidth,center}
\begin{tabular}{c|c|cccccccccccccccccccccccc}
    \toprule[.5mm]
    \multicolumn{2}{c|}{Tasks} & \multicolumn{15}{c|}{\textbf{\lift{} Classification W/O Feat. Names}}                                                                   & \multicolumn{9}{c}{\textbf{\lift{} Classfication W/ Feat. Names}} \bigstrut\\
    \midrule
    \multicolumn{2}{c|}{Dataset (ID)} & \multicolumn{3}{c}{Customers (1511)} & \multicolumn{3}{c}{Texture (1493)} & \multicolumn{3}{c}{Margin (1491)} & \multicolumn{3}{c}{TAE (48)} & \multicolumn{3}{c|}{Vehicle (54)} & \multicolumn{3}{c}{TAE (48)} & \multicolumn{3}{c}{CMC (23)} & \multicolumn{3}{c}{Vehicle (54)} \bigstrut\\
    \midrule
    \multicolumn{2}{c|}{\textbf{\lift{}/GPT-J}} & \multicolumn{3}{c}{93.97$\pm$1.00} & \multicolumn{3}{c}{50.32$\pm$2.18} & \multicolumn{3}{c}{50.23$\pm$1.33} & \multicolumn{3}{c}{61.29$\pm$6.97} & \multicolumn{3}{c|}{64.31$\pm$2.37} & \multicolumn{3}{c}{67.74$\pm$11.48} & \multicolumn{3}{c}{48.36$\pm$0.97} & \multicolumn{3}{c}{69.02$\pm$3.67}\\
    \midrule
    \multirow{4}{*}{\textbf{\lift{}/\gptt{}}}& \textbf{Ada}   & \multicolumn{3}{c}{95.39$\pm$0.67} & \multicolumn{3}{c}{67.50$\pm$1.42} & \multicolumn{3}{c}{59.37$\pm$0.92} & \multicolumn{3}{c}{65.59$\pm$6.63} & \multicolumn{3}{c|}{70.20$\pm$2.73} & \multicolumn{3}{c}{67.74$\pm$2.63} & \multicolumn{3}{c}{57.48$\pm$1.14} & \multicolumn{3}{c}{72.16$\pm$2.00} \\
    & \textbf{Babbage} & \multicolumn{3}{c}{96.81$\pm$0.07} & \multicolumn{3}{c}{62.19$\pm$1.80} & \multicolumn{3}{c}{67.50$\pm$3.87} & \multicolumn{3}{c}{61.29$\pm6.97$} & \multicolumn{3}{c|}{72.06$\pm$3.82} & \multicolumn{3}{c}{64.52$\pm$6.97} & \multicolumn{3}{c}{57.06$\pm$2.15} & \multicolumn{3}{c}{70.00$\pm$1.44} \\
   & \textbf{Curie} & \multicolumn{3}{c}{95.21$\pm$0.06} & \multicolumn{3}{c}{62.50$\pm$0.97} & \multicolumn{3}{c}{61.88$\pm$1.48} & \multicolumn{3}{c}{66.67$\pm$6.09} & \multicolumn{3}{c|}{74.27$\pm$0.73} & \multicolumn{3}{c}{65.59$\pm$4.02} & \multicolumn{3}{c}{55.42$\pm$0.84} & \multicolumn{3}{c}{70.66$\pm$2.28} \\
    & \textbf{Davinci} & \multicolumn{3}{c}{96.81$\pm$0.41} & \multicolumn{3}{c}{57.19$\pm$0.70} & \multicolumn{3}{c}{58.13$\pm$2.50} & \multicolumn{3}{c}{64.52$\pm$9.50} & \multicolumn{3}{c|}{71.47$\pm$0.88} & \multicolumn{3}{c}{65.59$\pm$6.63} & \multicolumn{3}{c}{56.31$\pm$0.04} & \multicolumn{3}{c}{68.16$\pm$1.69}\\
    \bottomrule[.5mm]
    \toprule[.5mm]
    \multicolumn{2}{c|}{Tasks} & \multicolumn{24}{c}{\textbf{In-context Classifction }} \bigstrut\\
    \midrule
    \multicolumn{2}{c|}{Dataset (ID) / \#Prompts} & \multicolumn{4}{c}{TAE (48)/50}    &  & \multicolumn{3}{c}{Breast (13)/35}    & \multicolumn{4}{c}{LED (40496)/32} & \multicolumn{4}{c}{ Customers (1511)/28}  &  & \multicolumn{3}{c}{Vehicle (54)/42}    & \multicolumn{4}{c}{Hamster (893)/13} \bigstrut\\
    \midrule 
    \multicolumn{2}{c|}{\textbf{\gptj{}}} &\multicolumn{4}{c}{34.33$\pm$1.47} & & \multicolumn{3}{c}{56.90$\pm$19.51} & \multicolumn{4}{c}{10.00$\pm$0.82} & \multicolumn{4}{c}{56.06$\pm$17.14} &  &\multicolumn{3}{c}{25.49$\pm$0.55}  &\multicolumn{4}{c}{48.89$\pm$3.14} \\
    \midrule
    \multirow{4}{*}{\textbf{\gptt{}}}& \textbf{Ada}   & \multicolumn{4}{c}{37.64$\pm$4.02} &  & \multicolumn{3}{c}{62.07$\pm$1.41} & \multicolumn{4}{c}{8.00$\pm$1.63} & \multicolumn{4}{c}{60.61$\pm$1.42} &  & \multicolumn{3}{c}{28.82$\pm$2.10} & \multicolumn{4}{c}{57.78$\pm$6.29} \\
    & \textbf{Babbage} & \multicolumn{4}{c}{47.31$\pm$3.04} &  & \multicolumn{3}{c}{71.26$\pm$0.81} & \multicolumn{4}{c}{11.00$\pm$0.00} & \multicolumn{4}{c}{53.79$\pm$12.07} &  & \multicolumn{3}{c}{24.32$\pm$0.56} & \multicolumn{4}{c}{53.33$\pm$5.44} \\
    & \textbf{Curie} & \multicolumn{4}{c}{32.26$\pm$0.00} &  & \multicolumn{3}{c}{70.69$\pm$0.00} & \multicolumn{4}{c}{20.67$\pm$4.78} & \multicolumn{4}{c}{67.80$\pm$0.53} &  & \multicolumn{3}{c}{26.28$\pm$2.22} & \multicolumn{4}{c}{53.33$\pm$0.00} \\
    & \textbf{Davinci} & \multicolumn{4}{c}{49.46$\pm$4.02} &  & \multicolumn{3}{c}{67.82$\pm$4.06} & \multicolumn{4}{c}{20.67$\pm$6.60} & \multicolumn{4}{c}{68.94$\pm$0.54} &  & \multicolumn{3}{c}{26.28$\pm$2.22} & \multicolumn{4}{c}{55.55$\pm$3.14}\\
    \bottomrule[.5mm]
    \end{tabular}
    \end{adjustbox}
\vspace{1mm}
  \label{tab:comp_classif_full}
\end{table}

\begin{table}[!htbp]
\caption{\textbf{Comparison of \lift{} on different LMs across regression tasks.} 
The regression performance is measured by RAE ($\downarrow$).
In general, \lift{}/\gptt{} with \texttt{Davinci} model performs the best, but the gaps to other models are not always significant.}
\begin{center}
\small{
\begin{tabular}{l|c|ccccccccccc} 
\toprule
\multirow{2}{*}{\diagbox{Function}{Method}} & \multirow{2}{*}{\textbf{\lift{}/\gptj{}}} & \multicolumn{4}{c}{\textbf{\lift{}/\gptt{}}} \\ [.1in]
&& Ada & Babbage &Curie & Davinci \\  \midrule
linear & 0.08$\pm$0.01 & 0.06$\pm$0.01 & 0.06$\pm$0.00 & 0.06$\pm$0.01 & 0.06$\pm$0.00 \\
quadratic & 0.11$\pm$0.00 & 0.13$\pm$0.00 & 0.11$\pm$0.02 & 0.10$\pm$0.01 & 0.09$\pm$0.00 \\
exponential & 0.11$\pm$0.02 & 0.09$\pm$0.00 & 0.09$\pm$0.01 & 0.08$\pm$0.00 & 0.08$\pm$0.00 \\
cosine & 0.38$\pm$0.08 & 0.44$\pm$0.10 & 0.41$\pm$0.06 & 0.38$\pm$0.01 & 0.38$\pm$0.05 \\
L1-norm & 0.10$\pm$0.00 & 0.09$\pm$0.01 & 0.10$\pm$0.01 & 0.08$\pm$0.01 & 0.09$\pm$0.01 \\
piecewise & 0.15$\pm$0.01 & 0.17$\pm$0.05 & 0.15$\pm$0.02 & 0.15$\pm$0.01 & 0.14$\pm$0.01 \\
\bottomrule
\end{tabular} }
\end{center}
\label{table:clf_model_comparison}
\end{table}

In this experiment, we apply \lift{} to different pretrained LMs to verify whether \lift{} benefits more from larger LMs.
Together with previously used \gptj{} and \gptt{} (the version named \texttt{Ada}), we consider three bigger versions of \gptt{}, namely \texttt{Baggage}, \texttt{Curie}, and \texttt{Davinci} (in the ascending order of the number of parameters).
We compare all models on several classification tasks in Table~\ref{tab:comp_classif_full} and regression tasks in Table~\ref{table:clf_model_comparison}.
Overall, we find that the performance gain of using larger LMs is not consistently significant for \lift{}.
Although larger LMs outperform smaller LMs in many cases, the improvements are relatively small.

\paragraph{Verifying the capability of large LMs, when \lift{} is not used.} 
We first verify if larger LMs are more helpful for the evaluated downstream tasks. 
We  evaluate LMs in the in-context classification when no fine-tuning (\lift{}) is involved.
Table~\ref{tab:comp_classif_full} shows consistent improvements in classification performance when the size of LMs increases across all the tasks.
Thus, larger LMs, with larger embedded knowledge, are more useful for these downstream tasks.

\paragraph{When \lift{} is used.} 
Both Table~\ref{tab:comp_classif_full} and Table~\ref{table:clf_model_comparison} show that using larger LMs may positively affect \lift{} in several tasks and settings compared to the smaller LMs.
However, the performance gains from replacing the smaller LMs with larger LMs are not consistent across the settings.
For instance, in the classification settings without feature names, \texttt{Davinci} performs better than \gptj{}  on four datasets and worse than on one dataset.
For the setting with feature names, \gptj{} performs better than \texttt{Davinci} on two out of three tasks.
Furthermore, the performance gains of large LMs over the smaller models are not relatively significant.
We note that \lift{} always outperforms the in-context learning using the same pretrained LMs in most cases.
The regression results shown in Table~\ref{table:clf_model_comparison} further confirm that the improvement from utilizing larger LMs is relatively small.

\subsection{Quantitative Classification Evaluations on Neural-Net-Based Synthetic Datasets}\label{app:boundary_complexity}

\begin{figure}[H]
    \centering
    
    \includegraphics[width=0.85\textwidth]{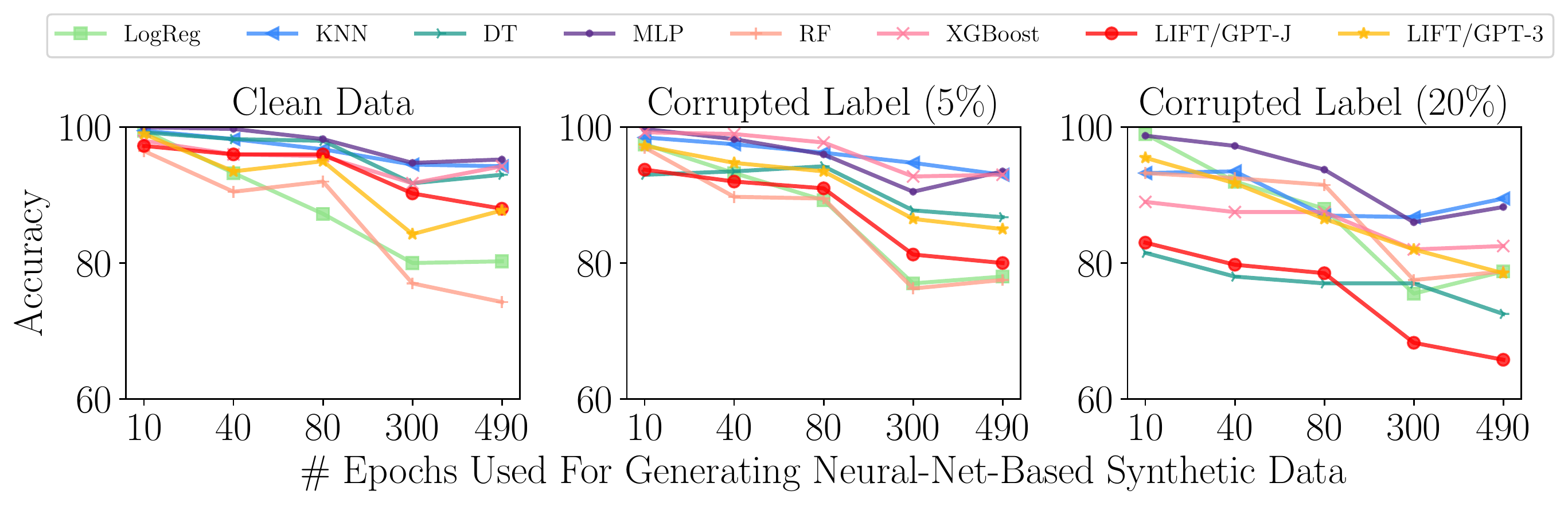}
    \vspace{-2mm}
    \caption{
    \textbf{How accuracy ($\uparrow$) changes as the target classification problem becomes more complex, i.e., the ground-truth decision boundary becomes more complex. }
    The x-axis shows the number of epochs we used to train the neural network on the \texttt{Rolls} dataset. Note that the network becomes more complex as the number of epochs increases. Thus the classification problem also gets challenging.
    We measure the performances on three cases: (left) clean data, (middle) label-corrupted data with corruption probability $5\%$ and (right)  $20\%$.}
    \label{fig:dec_diff}
\end{figure}

In Sec.~\ref{sec:inductive_bias} and Appendix~\ref{app:inductive_bias}, we assess how well \lift{}/\gpt{}s adapt to different shapes of decision boundaries on neural-net-based synthetic datasets.
We now provide the test accuracies for all models.
\textit{For binary-class datasets,}
Fig.~\ref{fig:dec_diff} shows the accuracies on three binary-class datasets when the  difficulty of classification tasks varies by using different network checkpoints at different epochs.
As the difficulty increases or the level of corruption increases, all methods tend to decrease classification accuracy.
We provide three settings of training data: clean data, 5\% label corruption, and 20\% label corruption.
We observe that \lift{}/\gptt{} outperforms logistic regression and decision tree, especially when the label corruption is up to 20\%. However, \lift{}/\gptj{} is performing worse than other baselines in the data corruption scenarios. 
\textit{For 3-class and 5-class datasets,} both \lift{}/\gptj{} and \lift{}/\gptt{} achieve approximately $90\%$, while the best baselines (MLP and XGBoost) obtains approximately $92\%$ and $91\%$ for the 3-class and 5-class data, respectively.

\subsection{Can \lift{} Perform Ridge Regression via Data Augmentation?}\label{app:ridge}

\begin{table}[h]
    \caption{
    \textbf{Performance of \lift{} on Ridge regression.} 
    We measure the RAE ($\downarrow$) of \lift{} corresponding to Linear Regression (LR) and Ridge Regression. 
    The RAEs indicate that \lift{} does not perform well on the Ridge regression problem. 
    }
    \centering
    \scriptsize
    \begin{tabular}{ll|cccc}
    \toprule[.3mm]
    
   \multirow{2}{*}{$p$}  & \multirow{2}{*}{$\lambda$} & \multicolumn{2}{c}{\textbf{\lift{}/\gptj{}}}  & \multicolumn{2}{c}{\textbf{\lift{}/\gptt{}}}  \\ 
   & &LR &Ridge &LR &Ridge \\ \midrule
1   & 0      & 0.000 $\pm$ 0.000        & 0.000 $\pm$ 0.000         &  0.915$\pm$0.000         & 0.000$\pm$0.000         \\
1   & 10     & 0.000 $\pm$ 0.000        & 0.016 $\pm$ 0.000         &  0.915$\pm$0.000         & 0.016$\pm$0.000         \\
1   & 50     & 0.000 $\pm$ 0.000        & 0.403 $\pm$ 0.000         &  0.915$\pm$0.000         & 0.402$\pm$0.000         \\
1   & 100    & 0.000 $\pm$ 0.000        & 1.691 $\pm$ 0.000         &  0.915$\pm$0.000         & 1.690$\pm$0.000         \\
1   & 1000   & 0.000 $\pm$ 0.000        & 170.406 $\pm$ 0.000       &  0.915$\pm$0.000         & 170.612$\pm$0.000       \\
\midrule
10  & 0      & 0.532 $\pm$ 0.000        & 0.532 $\pm$ 0.000         &  0.915$\pm$0.000         & 0.521$\pm$0.000         \\
10  & 10     & 0.374 $\pm$ 0.000        & 0.369 $\pm$ 0.000         &  0.915$\pm$0.000         & 0.504$\pm$0.000         \\
10  & 50     & 0.417 $\pm$ 0.000        & 0.523 $\pm$ 0.000         &  0.915$\pm$0.000         & 0.563$\pm$0.000         \\
10  & 100    & 0.365 $\pm$ 0.000        & 1.307 $\pm$ 0.000         &  0.915$\pm$0.000         & 1.539$\pm$0.000         \\
10  & 1000   & 0.414 $\pm$ 0.000        & 114.042 $\pm$ 0.000       &  0.915$\pm$0.000         & 111.357$\pm$0.000       \\
\midrule
50  & 0      & 0.688 $\pm$ 0.000        & 0.688 $\pm$ 0.000         &  0.915$\pm$0.000         & 1.064$\pm$0.000         \\
50  & 10     & 0.628 $\pm$ 0.000        & 0.635 $\pm$ 0.000         &  0.915$\pm$0.000         & 0.909$\pm$0.000         \\
50  & 50     & 0.553 $\pm$ 0.000        & 0.732 $\pm$ 0.000         &  0.915$\pm$0.000         & 1.296$\pm$0.000         \\
50  & 100    & 0.774 $\pm$ 0.000        & 1.857 $\pm$ 0.000         &  0.915$\pm$0.000         & 2.311$\pm$0.000         \\
50  & 1000   & 0.970 $\pm$ 0.000        & 118.241 $\pm$ 0.000       &  0.915$\pm$0.000         & 133.122$\pm$0.000       \\
         \bottomrule[.3mm]
    \end{tabular}
    \vspace{1mm}
    \label{tab:ridge}
\end{table}

As shown in Fig.~\ref{fig:reg_2d}, \lift{} can perform linear regression. 
We take one step further and study whether \lift{} can perform Ridge regression. 
Note that this is a non-trivial task as the \lift{} framework does not allow any changes to the loss function.
Consider a standard ridge regression problem solving the optimal $\vw$ with $p$ parameters so that $\|\vy - \mX\vw\|_2^2 + \lambda \|\vw\|_2^2$ is minimized. 
Note that this problem is equivalent to minimizing $\|[\vy^T, 0]^T - [\mX^T, \sqrt{\lambda} \mI]^T\vw\|_2^2$.
Therefore, if we add $p$ additional training samples $\sqrt{\lambda}I$, 
one can perform ridge regression via data augmentation.  
Inspired by this, we study whether one can perform ridge regression via data augmentation within the framework of the \lift{} framework.  
The results of \lift{} on Ridge regression are reported in Table~\ref{tab:ridge}. 

We observe that \lift{} fails to perform Ridge regression.
This is expected, as \lift{} is shown to be robust to outliers (in Sec.~\ref{sec:robustness} and Appendix~\ref{app:robustness}).

\subsection{\lift{}'s Training Curve}\label{app:training_curves}
\begin{figure}[!htbp]
    \centering
    \includegraphics[width=0.8\textwidth]{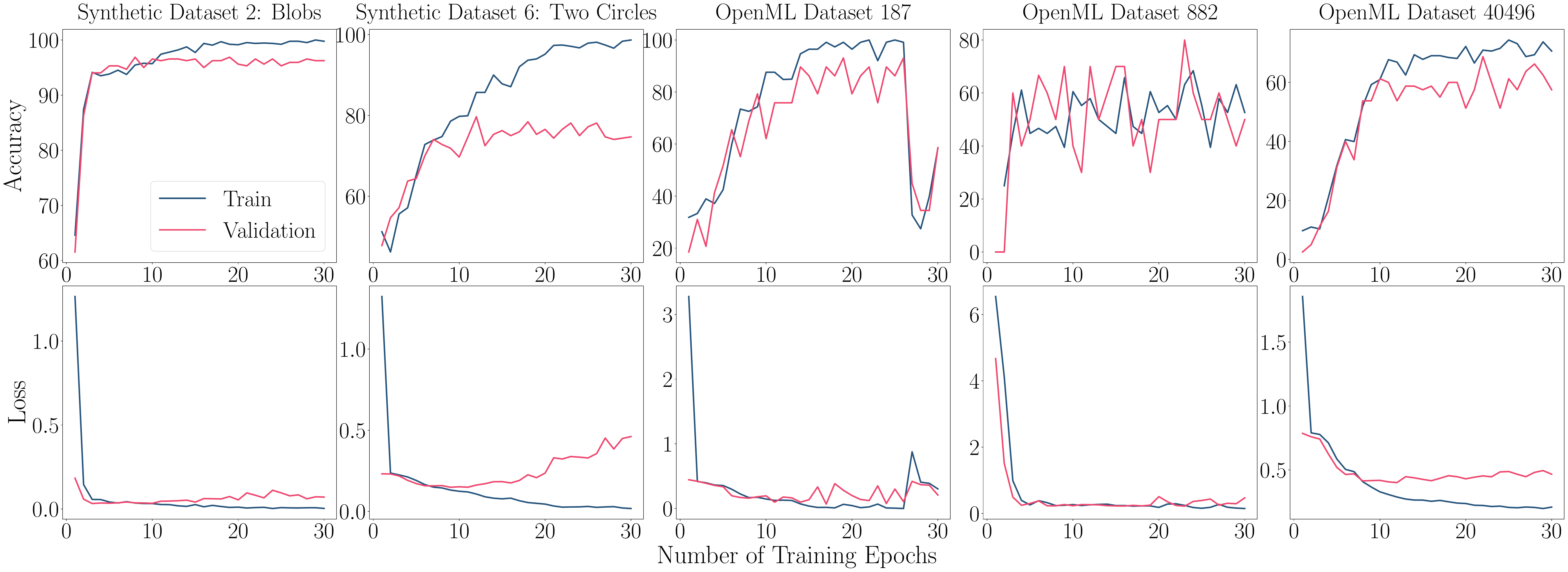}
    \caption{\textbf{Learning curves of \lift{}/\gptj{} on several synthetic/OpenML classification datasets.} We plot the accuracy (top row) and the loss (bottom row) of \lift{} varying the number of training epochs.}
    \label{fig:clf_log}
\end{figure}

\begin{figure}[!htbp]
    \centering
    \includegraphics[width=\textwidth]{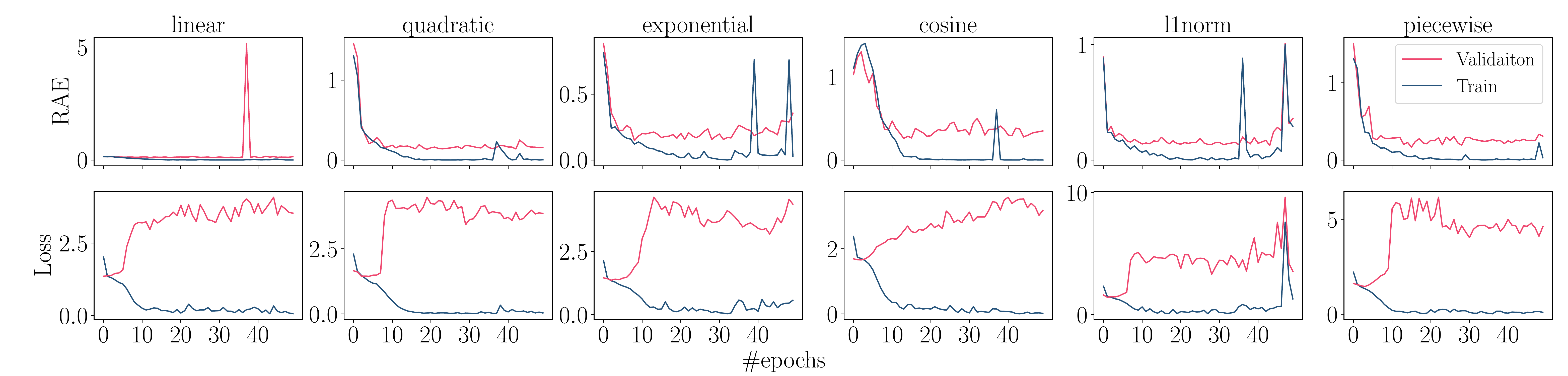}
    \caption{\textbf{Regression RAE and Loss curves of \lift{}/\gptj{} on the synthetic regression datasets.}
    We observe that \lift{}/\gptj{} only requires a few epochs to achieve good performance.}
    \label{fig:reg_log}
\end{figure}

We report the learning curves of \lift{} in terms of LM-loss and accuracies/RAE for several classification and regression tasks.
We observe a decrease in training loss over the tasks and datasets.
We select the best models based on the validation criteria (accuracy for classification and RAE for regression) on the validation sets.
Fig.~\ref{fig:clf_log} visualize the accuracy and loss of \lift{}/\gptj{}  in the training and validation process for classification tasks.
For the regression task, Fig.~\ref{fig:reg_log} shows that the decrease in RAE does not necessarily imply a decrease in loss. 
Furthermore, we observe that \lift{} only requires a few epochs to achieve good performance. 

\section{Additional Discussion}\label{app:discussion}
Continuing from Sec.~\ref{sec:dis_con}, here we elaborate on the difficulty of regression tasks and the broader impact of \lift{}, and discuss the limitation on classification tasks and other open questions.

\subsection{Limitations and Open Questions}

\paragraph{The difficulty of regression tasks.}\label{sec:difficulty_reg}
For regression tasks, 
in addition to poor performance on high-dimensional functions, some interesting phenomena observed in the classification tasks are not consistently observed in the regression tasks.
For example, incorporating feature names in the prompts does not consistently improve \lift{} in the regression tasks (see Sec.~\ref{app:context}).

As previously discussed in Sec.~\ref{sec:performance}, the difficulty of regression tasks on \lift{} may come from the classification loss function used in LMs.
Due to the adoption of the classification loss function, two different predictions will lead to the same loss, even if one of the predictions is closer to the true $y$ value.
As a result, we also observe that a reduction in RAE does not necessarily imply a reduction in LM loss (see Fig.~\ref{fig:reg_log}).
Therefore, we use RAE as the criterion for model selection.
Moreover, how \lift{} understands numerical values may also limit the regression performance of \lift{}. 
Recent works~\citep{zhang2020language,wallace2019nlp,naik2019exploring,ren2020enhancing} have illustrated the difficulty and failures of the existing LMs in understanding the numbers because two numbers  with close values can have very different tokenizations~\citep{wallace2019nlp}. 
Recent attempts~\citep{nogueira2021investigating,ke2020rethinking,he2020deberta,wang2019encoding,huang2020improve} propose new encoding schemes of numbers to improve the LMs' numerical capabilities, probably helping \lift{} in the regression tasks.

A promising method for improving \lift{} on regression is  {\it level encoding}.
The idea of level encoding is to discretize the continuous values of the output $y$ to better utilize the classification loss of LMs.
Assuming that the range of $y$ is known, we can partition this range into a finite number of bins and represent 
all values in the same bin by a unique canonical representation in a way that the number of mismatched bits between the representations of two values is proportional to their absolute difference.
For instance, for all real-value $y \in [0, 3]$, we can define three bins as $\{[0, 1), [1, 2), [2, 3]\}$ with the canonical representations being $00, 01, 11$.
With these bins, 0.3 and 0.7 are represented as $00$, and $1.5$ and $1.1$ are represented as $01$.
The distance between representations of $0.3$ and $1.1$ is only $1$ bit, which is proportional to their absolute distance of $0.8$.
For the training of \lift{}, we convert all output values in the original training dataset into the level-encoding canonical representation and use them as the target values.
By using the level encoding technique, the loss function of LMs can better capture the distance between the prediction and the true values, thus potentially improving the generalization of \lift{} on regression tasks.
We leave this as one of the interesting directions for our future investigation. 

\paragraph{The limitation of \lift{} on classification tasks.}
We observe that \lift{} does not perform comparably well on classification tasks when the number of classes is large.
For instance, Table~\ref{tab:classification_accuracy} shows that the accuracies of  \lift{}/\gptt{} are lower than RBF-SVM and XGboost on the datasets with 100 classes.
Another limitation is that the dimension of features \lift{} can handle is upper bounded due to the limited context length of LMs.
This limitation may be mitigated by using LMs with a more memory-efficient variant or implementation of transformer models, {\it e.g.}, see \citep{dao2022flashattention}.

\paragraph{Other open questions.}
In addition to previously discussed questions of improving \lift{} for regression and classification tasks, 
our pioneering work on \lift{} is also expected to open up interesting research questions on generalist models. First, can generalist LMs ({\it e.g.}, \gpt{}s) play a leading role in developing universal models that can adapt well to any data? Second, can we apply \lift{} to different generalist models, such as GATO~\citep{reed2022generalist}?

\subsection{Broader Impact}
\label{app:impact}

\lift{} greatly simplifies the machine learning pipeline that requires only the reformatting of training datasets of the target task.
This simplicity helps enable \textit{no-code ML} for the masses, where general users without prior knowledge of the ML frameworks can use \lift{} for their target non-language tasks by properly designing the input/output prompt format.
Therefore, \lift{} can apply to a wide range of applications and areas, such as credit loaning, disease diagnosis, and criminal sentencing. 
This is closely related to the line of automated machine learning research~\citep{feurer-neurips15a,feurer-arxiv20a}, which aims to automate the standard machine learning methods pipeline.

Employing \lift{} without careful justification or understanding will lead to undesired outcomes, such as discrimination. 
Since most existing language models (LMs) are pretrained on a large amount of human-annotated data, \lift{} could exhibit discrimination against different demographic groups (\textit{e.g.}, gender, race, ethnicity) due to the bias existing in the training datasets. 
In other words, \lift{} may prefer certain groups while making decisions in downstream tasks, especially when feature names and different demographic contexts are fed at training and inference time. 
This effect is exacerbated by the use of large pretrained LMs (\textit{i.e.}, \gptj{} and \gptt{}), which have been known to inherently contain bias \citep{sheng-etal-2019-woman}. 
The bias in the pretraining data for these large language models adds an opaque layer to regression and classification tasks beyond bias within the downstream data.
Therefore, adopting \lift{} in tasks that consider demographic information requires more consideration to avoid discrimination.
\tuan{To further remove the bias, users can combine \lift{} with the existing fairness-aware reweighting mechanisms~\citep{agarwal2018reductions,roh2021fairbatch} or data augmentation and parameter-efficient fine-tuning techniques~\citep{gira-etal-2022-debiasing}.}

Finally, we emphasize that more model evaluation steps are required when applying \lift{} instead of using it as a panacea for all applications.
We believe our work can significantly benefit society by providing a simple tool for handling various tasks with proper justification.

\end{document}